\documentclass[final,3p,times,twocolumn]{elsarticle}

\pdfoutput=1

\usepackage{lineno}

\usepackage{acronym}
\usepackage[latin1]{inputenc}
\usepackage{graphicx}
\usepackage{url}
\usepackage{amssymb}
\usepackage{amstext}
\usepackage{amsmath}
\usepackage{amsfonts} 
\usepackage{algorithmic}
\usepackage{algorithm} 
\usepackage{color,soul}
\usepackage{booktabs}
\usepackage{tabularx}
\usepackage{multirow} 
\usepackage{rotating}

\usepackage{adjustbox}
\usepackage{longtable}

\usepackage{pdflscape}
\usepackage{longtable}
\usepackage{multirow}
\usepackage{xcolor}
\usepackage{color}

\usepackage{diagbox}
\usepackage{makecell}
\usepackage{calc}

\newcommand\Tstrut{\rule{0pt}{2.6ex}} 
\newcommand\Bstrut{\rule[-0.9ex]{0pt}{0pt}} 
\newcommand{\TBstrut}{\Tstrut\Bstrut} 

\usepackage{pgfplots}
\pgfplotsset{compat=1.17}
\usepgfplotslibrary{statistics}
\usepgfplotslibrary{units}
\usepgfplotslibrary{groupplots}
\usepgfplotslibrary{dateplot}
\usepackage{tikz}
\usetikzlibrary{plotmarks}
\usetikzlibrary{positioning}

\usepackage{csvsimple}
\usepackage{pgfplotstable}

\usepackage{subcaption}

\usepackage{nicematrix}

\newcolumntype{d}[1]{D{.}{.}{#1}}

\usepackage{siunitx}

\renewcommand*\citep[1]{\cite{#1}}

\renewcommand*\citet[1]{\cite{#1}}

\newcommand{\bb}{\color{black}}

\usepackage{caption}
\newcommand\gbh{4.5cm}
\newcommand*\gvsa{\vspace*{-4.5mm}}
\newcommand*\gvsb{\vspace*{0mm}}

\pgfplotsset
{
	select coords between index/.style 2 args=
	{
    x filter/.code=
    {
        \ifnum\coordindex<#1\fi
        \ifnum\coordindex>#2\fi
    }
	},
}

\pgfplotsset{
 box plot width/.initial=1em,
 box plot/.style={
    /pgfplots/.cd,
    black,
    only marks,
    mark=-,
    mark size=\pgfkeysvalueof{/pgfplots/box plot width},
    /pgfplots/error bars/.cd,
    y dir=plus,
    y explicit,
 },
 box plot box/.style={
    /pgfplots/error bars/draw error bar/.code 2 args={%
        \draw  ##1 -- ++(\pgfkeysvalueof{/pgfplots/box plot width},0pt) |- ##2 -- ++(-\pgfkeysvalueof{/pgfplots/box plot width},0pt) |- ##1 -- cycle;
    },
    /pgfplots/table/.cd,
    y index=2,
    y error expr={\thisrowno{3}-\thisrowno{2}},
    /pgfplots/box plot
 },
 box plot top whisker/.style={
    /pgfplots/error bars/draw error bar/.code 2 args={%
        \pgfkeysgetvalue{/pgfplots/error bars/error mark}%
        {\pgfplotserrorbarsmark}%
        \pgfkeysgetvalue{/pgfplots/error bars/error mark options}%
        {\pgfplotserrorbarsmarkopts}%
        \path ##1 -- ##2;
    },
    /pgfplots/table/.cd,
    y index=4,
    y error expr={\thisrowno{2}-\thisrowno{4}},
    /pgfplots/box plot
 },
 box plot bottom whisker/.style={
    /pgfplots/error bars/draw error bar/.code 2 args={%
        \pgfkeysgetvalue{/pgfplots/error bars/error mark}%
        {\pgfplotserrorbarsmark}%
        \pgfkeysgetvalue{/pgfplots/error bars/error mark options}%
        {\pgfplotserrorbarsmarkopts}%
        \path ##1 -- ##2;
    },
    /pgfplots/table/.cd,
    y index=5,
    y error expr={\thisrowno{3}-\thisrowno{5}},
    /pgfplots/box plot
 },
 box plot median/.style={
    /pgfplots/box plot
 }
}

\pgfplotsset{
    discard if/.style 2 args={
        x filter/.code={
            \edef\tempa{\thisrow{#1}}
            \edef\tempb{#2}
            \ifx\tempa\tempb
                
            \fi
        }
    },
    discard if not/.style 2 args={
        x filter/.code={
            \edef\tempa{\thisrow{#1}}
            \edef\tempb{#2}
            \ifx\tempa\tempb
            \else
                
            \fi
        }
    }
}

\newcommand*{\ReadOutElement}[4]
{%
    \pgfplotstablegetelem{#2}{[index]#3}\of{#1}%
    \let#4\pgfplotsretval
}

\makeatletter
\def\ps@pprintTitle{%
\let\@oddhead\@empty
 \let\@evenhead\@empty
 \def\@oddfoot{}%
 \let\@evenfoot\@oddfoot}
\makeatother

\begin{document}

\begin{frontmatter}

\title{An Empirical Study on the Joint Impact of Feature Selection and Data Re-sampling on Imbalance Classification}

\author[lab1]{Chongsheng Zhang\fnref{eq}}\ead{chongsheng.zhang@yahoo.com}
\fntext[eq]{Co-first authors, equally contributed to this work}
\author[lab2]{Paolo Soda\fnref{eq}}\ead{p.soda@unicampus.it}
\author[lab1]{Jingjun Bi}\ead{bi.jingjun@outlook.com}
\author[lab1]{Gaojuan Fan\corref{cor}}\ead{fangaojuan@126.com}
\cortext[cor]{Corresponding author}
\author[lab1]{George Almpanidis}\ead{almpanidis@gmail.com}
\author[lab3]{Salvador Garc\'{\i}a}\ead{salvagl@decsai.ugr.es}

\address[lab1]{School of Computer and Information Engineering, Henan University, China}
\address[lab2]{Department of Engineering, University Campus Bio-Medico of Rome, Italy}
\address[lab3]{DaSCI Andalusian Institute. Department of Computer Science and Artificial Intelligence, University of Granada, Spain}

\begin{abstract}
In predictive tasks, real-world datasets often present different degrees of imbalanced (i.e., long-tailed or skewed)  distributions. While the majority (the head or the most frequent) classes have sufficient samples, the minority (the tail or the less frequent or rare) classes can be under-represented by a rather limited number of samples. Data pre-processing has been shown to be very effective in dealing with such problems. On one hand, data re-sampling is a common  approach to tackling class imbalance. On the other hand, dimension reduction, which reduces the feature space, is a conventional technique for reducing noise and inconsistencies in a dataset. However, the possible synergy between feature selection and data re-sampling for high-performance imbalance classification has rarely been  investigated before. To address this issue, we carry out a comprehensive empirical study on the joint influence of feature selection and re-sampling on two-class imbalance classification. Specifically, we study the performance of two opposite pipelines for imbalance classification by applying  feature selection before or after data re-sampling. We conduct a large number of  experiments, with a total of \textit{9225} tests, \bb on \textit{52} publicly available datasets, using \textit{9} feature selection methods, \textit{6} re-sampling approaches for class imbalance learning, and \textit{3} well-known classification algorithms. Experimental results show that there is no constant winner between the two pipelines; thus both of them should be considered to derive the best performing model for imbalance classification.  We find that the performance of an imbalance classification model not only depends on the classifier adopted and the ratio between the number of majority and minority samples, but also depends on the ratio between the number of samples and features. Overall, this study should provide new reference value for researchers and practitioners in imbalance learning. 

\end{abstract}

\end{frontmatter}

\section{Introduction}
\label{sec:introduction}

Class imbalance, a.k.a., class skew or long-tailed distributions, refers to the case where a few classes/categories have a significantly inadequate number of samples in comparison to others \citep{FernandezGGPKH18}.  Imbalanced datasets are present in a large number of domains, such as banking fraud detection, medical diagnosis, text classification, software defect prediction.  Its pervasiveness, and its close relevance to many real-world applications, have attracted substantial research effort in the last two decades \citep{bib:HeLearningImbalanceData2009}.

In the literature, there are many techniques for combating the class imbalance problem, including internal approaches that tailor an algorithm to imbalanced data,  data re-sampling  approaches~\citep{PanZWY20}, cost-sensitive learning, and ensemble learning~\citep{bib:chawla2009data,LazaroHF20}. Interested readers may  refer to overviews on this topic in~\citet{bib:chawla2009data,bib:branco2016survey,bib:krawczyk2016learning,bib:DECOCKBS2018,bib:ZhangHFKBS2019}. 

Feature selection/dimension reduction is often a very effective technique for improving classification performance. It also simplifies the classification model to be built. Yet, inappropriate reduction in features may lead to a loss in the discrimination power and the recognition accuracy. \textit{Watanabe's ugly duckling theorem}~\citep{bib:Watanabe} supports the need for a careful choice of the features, stating that it is possible to make two arbitrary patterns similar by encoding them with a sufficiently large number of redundant features. Overall, feature selection has been an important research direction in machine learning and has attracted a huge amount of research effort~\citep{bib:Saeys2007ReviewFeature}.

However, up to now,  there are only a limited number of preliminary studies that merely explored the feature selection before data re-sampling (FS+DS for short) pipeline for improving the imbalance classification performance; moreover, the number of experiments is rather limited. For instance,  ~\citet{AliBG2005,KhoshgoftaarGS10,KhoshgoftaarGNW14} only used simple random undersampling techniques, whereas~\citet{WasikowskiTKDE10} just considered one oversampling method.

To derive the best imbalance classification model, the community is still in need of a more in-depth and comprehensive empirical study to investigate the joint impact of feature selection and re-sampling on imbalance classification, in particular the alternative feature selection after data re-sampling (DS+FS) pipeline. To address this issue, in this work, we provide a comprehensive empirical study on the joint influence of feature selection and data re-sampling on two-class imbalance classification. We use 52 benchmark datasets with imbalanced distributions,  9 feature selection methods, 6 re-sampling approaches for  class imbalance learning, and 3 representative classification algorithms. 

We reveal through these extensive experiments that the synergy between feature selection and data re-sampling can yield significantly stronger models for imbalance classification. Moreover, we find that there is no constant winner between the two competing pipelines (i.e., FS+DS v.s.  DS+FS); thus both of them deserve attention when building an imbalanced classifier.
In the following sections,  we also provide a  detailed analysis on the influence of the baseline classifiers, the ratio between the number of  majority and minority samples, and on the ratio  between the number of samples and features on the performance of  imbalance classification. 

The key empirical findings we draw from this work are as follows:

\begin{enumerate}

\item  It is worthwhile to try both the feature selection before or after data re-sampling pipelines to achieve the best imbalance classification performance; in particular, the feature selection after the data re-sampling pipeline (DS+FS) pipeline should not be neglected as it was before. 

\item  When using decision tree as the base learner, DS+FS outperforms FS+DS in terms of $\textnormal{G-Mean}$ and  F-score in more cases than not, especially when oversampling techniques are used. 

\item  When using SVM as the base learner, there is no clear winner between the two pipelines when oversampling methods are adopted; however, with undersampling methods, FS+DS predominates over DS+FS in most cases. 

\item For imbalanced datasets with low samples-to-feature ratios (SFR) but high imbalance ratios (IR), the FS+DS pipeline has a comparative advantage over the DS+FS pipeline. But for imbalanced datasets with high SFR but relatively low IR, there is no constant winner between them. 
\end{enumerate}


Besides, we also provide recommendations of the Top-3 specific combinations of the feature selection and data re-sampling methods that are the top performers for every evaluation metric under each  of the two pipelines. Overall, these studies and findings provide new insights and practical guidance for practitioners and researchers when tackling class imbalance problems.  

The rest of this paper is organized as follows: we introduce the background and related work in the next section, then describe the problem to be investigated in this work in Section~\ref{sec:ProblemDefinition}, which also introduces the specific methods for feature selection and class imbalance learning used in this work. 
Section~\ref{sec:ExperimentalFramework} presents the experimental framework, and Section~\ref{sec:Results} reports and analyzes the experimental results. Section~\ref{section:considANDrecomm} gives concluding remarks.


\section{Related Work} \label{sec:RelatedWork}
In this section, we will first present the background and directly related work of this paper. We will next introduce the general advances in imbalance learning in recent years, in which deep imbalance learning and GAN-based data augmentation are two cutting-edge directions. 

For general reviews of imbalance learning, interested readers may refer to~\citet{GuoLSMYB17,bib:chawla2009data,bib:DECOCKBS2018,bib:ZhangHFKBS2019,bib:lopez2013insight}, whilst reviews of feature selection methods can be found in  ~\citet{bib:Saeys2007ReviewFeature,bib:Huang2015ReviewFeature}. 

\subsection{Closely Related Work}

In this subsection, we chronologically summarize related work which investigates the joint influence of feature selection and data re-sampling approaches on two-class imbalance classification~\citep{AliBG2005,KhoshgoftaarGS10,KhoshgoftaarGNW14,WasikowskiTKDE10}. 

In the context of protein function prediction from sequence, \citet{AliBG2005} compares the respective performance of Support Vector Machine (SVM) and C4.5 when Random Under-Sampling (RUS) ~\citep{RUSds} is applied after the wrapper feature selection approach.

In \citet{KhoshgoftaarGS10}, the authors use 9 datasets to study the influence of feature selection and undersampling on software quality prediction. Their experimental results show that running feature selection on re-sampled data performs better than selecting the features from the original data. However, this work was developed for a very specific area; still, the authors only tested Random Under-Sampling (RUS) ~\citep{RUSds} and a few filter feature selection methods (i.e.,  $\chi^{2}$, Relief,  Gain Ratio). 
 
In \citet{KhoshgoftaarGNW14}, the authors study and propose an iterative method for feature selection on imbalanced data. It first uses Random Under-Sampling (RUS)  technique to obtain balanced data. It next applies a filter-based feature selection technique on the re-sampled data and ranks all the features according to their predictive capacity. However, this study is limited to the use of {\em RUS} only and only one filter-based feature selection method. As the authors admit, there should be future work considering additional datasets from other domains as well as more re-sampling techniques.

\citet{WasikowskiTKDE10} examines the performance of various feature selection methods and how they combat the class imbalance problem. Besides using \textit{7} filter feature selection methods (e.g., $\chi^{2}$, information gain, Pearson correlation coefficient, signal-to-noise correlation coefficient),  the authors also test the performance of one undersampling technique and one oversampling method (SMOTE~\citep{bib:Chawla2002,0001GHC18}). Their results show that the signal-to-noise correlation coefficient  and the feature assessment by sliding thresholds are great candidates for feature selection on imbalanced data with small sample size. However, the authors only test the feature selection before re-sampling strategy for class imbalance learning. Although they find that this strategy does not yield  improved results, they admitted  this direction needs further investigation, since they only test a few combinations of feature selection before re-sampling methods.

The authors in \cite{imsurvey2020is} study the impact of varying class imbalance ratios (IR for short) on the accuracy of a classifier. Empirical study in \cite{bib:lopez2013insight} show that,  imbalance ratio is not the main determinant in class imbalance learning problem. A higher IR will only further deteriorate the classification accuracy, but other data complexities  also influence the classification performance. 

It is worth noting that the all the above mentioned works  perform feature selection before re-sampling the data (FS+DS). None of them has explored the counterpart strategy, i.e., the feature selection after data selection pipeline (DS+FS). Furthermore, the experiments  carried out in the existing work only  used a very limited number of datasets,  feature selection methods, and data re-sampling approaches. In view of the above limitations, this work extends our previous conference paper in ~\citep{dsfs17}, to provide a comprehensive and more in-depth empirical study and insight into the joint impact of feature selection and data re-sampling on imbalance classification\footnote{The arXiv  preprint version of this work is available at  ~\citep{zhang2021empirical}.}. 

\subsection{Recent advances in imbalance learning}

\textbf{New oversampling techniques for imbalance learning}. MWMOTE \cite{MWMOTE} first identifies the borderline (hard-to-learn) minority class samples located near the majority samples, next generates synthetic samples inside the minority clusters after applying clustering on such borderline minority samples. In \cite{zhu2018geometric}, the authors propose an intuitive geometric space partition  (GSE) based method for imbalance classification. GSE constantly uses a new hyper-plane classifier to cut the current geometric data space into two partitions (halves), then removes the partition (half) which only contains majority samples. The basic idea of GSE is very similar to AdaBoost; the main difference lies in that GSE does not weight the samples but continuously eliminates the partitions with purely majority samples as a whole. In \cite{liu2019model}, the authors present MBS (model-based synthetic sampling), which is an oversampling technique for imbalance classification.  By assuming the existence of a relationship between features, for the minority instances, they iteratively leave one feature out and train a regression model on the values of the rest features to predict the value of the reserved feature. The tuples/records consisting of the predicted value for the reserved feature as well as the randomly generated values for the rest features will be the final synthetic samples to be used in imbalance learning.  In \cite{imputation}, the authors propose imputation-based oversampling method for imbalance learning. It first induces some random missing values (for some features) over the minority class samples, then estimates the missing values by means of missing data imputation techniques. The overall idea of this work is similar to \cite{liu2019model}. Finally, in \cite{Li2021438}, the authors modify the basic SMOTE algorithm by using the random difference between a selected base sample and one of its natural neighbors to generate synthetic samples.

\textbf{Deep imbalance learning}. \cite{imrect2019} addresses the problem of imbalanced data for the multi-label classification problem in deep learning and introduces incremental minority class discrimination learning by formulating a class rectification loss regularization, which imposes an additional batch-wise class balancing on top of the cross-entropy loss to rectify model learning bias due to the over-representation of the majority classes. In \cite{oksuz2020imbalance}, the authors provide a very comprehensive review of the imbalance problems in  deep learning based object detection and categorize them into four main types: class imbalance, scale imbalance, spatial imbalance and objective imbalance.

\textbf{GAN-based data augmentation for imbalance learning}. In recent years, Generative Adversarial Networks (GANs) \cite{gan} have been introduced into imbalance learning. \cite{gamo} proposes GAMO, which is an end-to-end deep oversampling model for imbalance classification. It effectively integrates the sample generation and classifier training together. When generating minority class instances, GAMO uses a class-specific instance generation unit to reduce the computation complexity. However, the performance of such a strategy may suffer from the lack of minority class information. To address this issue, the authors in \cite{bagan} design a balancing generative adversarial network (BAGAN), which exploits all classes to estimate the class distributions in the latent space of the AutoEncoders. 

GL-GAN \cite{glgan} utilizes GAN based data augmentation for two-class imbalance classification. It first adopts AutoEncoders to embed the input samples into a new latent space in which the inter-class distances of the samples are maximized and the intra-class distances are minimized.  Next, it uses SMOTE to generate minority samples in the new embedding space, then straightforwardly leverages GAN to generate realistic data. 

For reliability assessment of transmission gears, the authors in \cite{li2018cgan} adopt a CGAN based model to generate instances with a distribution similar to the original data and a heuristic approach to label the generated instances through exploring the nearest distance between generated instances and original class centers. 

In \cite{jing2021multiset}, the authors propose using GAN to generate a subset of majority samples that follow the same distribution as the original dataset. Next, they construct a few balanced subsets, each of which containing all the minority samples and an equal-size subset of GAN generated majority samples. Finally, upon each balanced subset, they use deep metric learning to build a two-class classification model. These models are then integrated to handle multi-class imbalanced classifications.

\section{Problem Definition} \label{sec:ProblemDefinition}


This work aims to study how the classification performance is affected by the joint use of feature selection and data re-sampling methods for imbalance learning. To this end, we compare the performance obtained by feature selection before the data re-sampling pipeline with the opposite feature selection after data re-sampling pipeline. These two pipelines are hereafter referred to as \textit{FS+DS} and \textit{DS+FS} for short. 

The main research questions that we will investigate in this work are as follows:

\begin{enumerate}

\item Which pipeline, FS+DS or DS+FS, can achieve the best performance for two-class imbalance classification, and under which conditions? 

\item How does the classification performance vary when using DS+FS and FS+DS on various datasets with different degrees of skewness and different number of samples and features?

\item  Although there is no constant winner between the general FS+DS and DS+FS pipelines, for a given classifier, which specific combinations of feature selection and data re-sampling methods have higher probability to become the top performers, thus should be recommended with priority?

\end{enumerate}

The first two issues are ``macro-level'' research questions,  which investigate and assess the overall comparative advantages of the two pipelines in two-class imbalance classification. The last issue is a ``micro-level'' research question; it aims to provide practical suggestions and guidelines for choosing the specific combinations of feature selection and data re-sampling methods, which have a higher probability of achieving the best imbalance classification performance, given datasets with different characteristics, and using different classifiers.

Before studying the three problems raised above, we first introduce the feature selection and data re-sampling methods for class imbalance learning that are considered in this work.

\subsection{Feature selection methods} \label{subsec:FeatureSelection}

In the literature, there exists a large variety of feature selection approaches, which can be categorized into filter, wrapper, and embedded methods~\citep{bib:Saeys2007ReviewFeature,bib:SurveyFeature2014Chandrashekar}. 
In this work, we will use 9 major feature selection approaches, which are presented below.

\subsubsection{Filter methods}
Filter methods assess the relevance of features in the data. 
They usually compute a feature relevance score, and features with low scores are removed.
The main advantages of filter methods are: they are independent of the classification algorithms and can easily scale to very high-dimensional datasets. Moreover, they are computationally efficient. But they also have drawbacks: they ignore the interaction with the classifier, as the search in the feature subset space is separated from the hypotheses space; moreover, most techniques are univariate.


The specific filter methods considered in this work are information gain, $\chi^{2}$ statistic,  Fisher feature selection, Gini Index and Student's T-test. The first method, referred to as {\em Info} for short,  evaluates the importance of an individual feature by measuring its corresponding information gain with respect to the class. $\chi^{2}$ statistic ({\em ChiS} for short) evaluates the weight of an attribute by computing its chi-squared statistic value with respect to the class. Fisher feature selection ({\em Fish}) computes a score derived from the distances between data points in different classes, divided by the distances between data points of the same class. Gini Index ({\em Gini}) makes use of a statistical phenomenon called the Lorentz curve to quantify wealth/income inequality situations. Student's T-test ({\em T-test}) feature selection commonly uses Welch's T-test metric to measure a feature's capability in distinguishing two classes. Interested readers may refer to \citet{li2016feature} for a more detailed  descriptions of these filter methods.

\subsubsection{Wrapper methods}
Wrapper methods embed the model hypotheses within the feature subset search. 
After defining an expand-and-search procedure for all the possible feature subsets,  various candidate subsets are generated and evaluated using a specific classification algorithm, making this approach specifically tailored to a given classification algorithm. 
Its apparent high computational overhead motivated the design of non-complete search methods~\citep{bib:Pudil1994}, which can be divided into deterministic and randomized search algorithms. The advantages of wrapper approaches lie in the interaction between feature subset search and model selection and the ability to take  feature dependencies into account. Their main problems are the higher risk of overfitting than filter approaches, and the relatively high computation cost.

The wrapper methods that we consider are Forward feature selection ({\em FWD})~\citep{FWDds},  Linear forward selection ({\em Linear})~\citep{Lineards}, and Ranked Search ({\em RS})~\citep{RSds}.
{\em FWD} feature selection first ranks each feature  on the basis of  its relevance to the class label, then sequentially adds features to an empty candidate set until the addition of new features can not further improve the prediction performance~\citep{FWDds}. Linear forward selection ({\em Linear}) improves the runtime FWD performance by limiting the attributes in the subsequent forward selection process to the top-ranked attributes~\citep{Lineards}. 
 In Rank Search ({\em RS}), subsets of increasing size (the best attribute plus the next best attribute) are evaluated and the best attribute set is derived~\citep{RSds}. 

\subsubsection{Embedded methods}
Embedded methods incorporate feature selection as part of the process of building a specific model. Their objective is to produce a predictive model, in which the features remained in the model are a byproduct of the modelling process. Similar to wrappers, embedded approaches are specially tailored to a given learning algorithm.  The main advantage of such methods is that they include the interaction with the classification model, but their major limitation is the computational load, which is less than wrapper methods though.

In our work, we considered the Sparse Multinomial Logistic Regression with Bayesian regularisation ({\em SBMLR}), which realizes sparse feature selection by using a Laplace prior ~\citep{SBMLRds}.

\subsection{Re-sampling approaches for class imbalance learning} \label{subsec:ImbalanceLearning}

In class imbalance learning, re-sampling (data selection) approaches can be divided into undersampling and oversampling methods. Both types of approaches resize the training set to make the class distribution more balanced so as to match the size of the other class. We focus on two-class imbalance classification in this work. Let  $N^{+}$ and $N^{-}$  denote the set of samples in the minority and majority classes, respectively. Undersampling methods sample a subset $\overline{N^{-}}$ from $N^{-}$, with $|\overline{N^{-}}|<|N^{-}|$, such that $|\overline{N^{-}}| \approx |N^{+}|$. On the contrary, oversampling approaches generate a set $\overline{N^{+}}$, with $|\overline{N^{+}}| \approx |N^{-}|$.  $\overline{N^{+}}$ is composed of all samples in $N^{+}$ and the generated instances. Both undersampling and oversampling have  their own drawbacks. The former may remove potentially useful data, while the latter may increase the likelihood of overfitting due to sample random replication~\citep{bib:Chawla2002,bib:Kubat1997}.

\subsubsection{Undersampling}
The undersampling methods considered in this work are Random Under-Sampling (RUS)~\citep{RUSds},  Condensed Nearest Neighbor decision rule (CNN)~\citep{cnnds} and One-Sided Selection (OSS)~\citep{bib:Kubat1997,bib:Batista2000}.

Random Under-Sampling ({\em RUS})  balances class distribution through the random elimination of majority class examples, until the desired class ratio between the minority and majority classes is reached~\citep{RUSds}. 

Condensed Nearest Neighbor decision rule ({\em CNN})   successively discards instances (samples) that can be correctly classified using a prediction model built upon the current subset of instances~\citep{cnnds}.

One-sided Selection ({\em OSS}) detects and removes samples that are less reliable according to certain heuristics~\citep{bib:Kubat1997,bib:Batista2000}. Samples are divided into four groups: (i) samples suffering from class-label noise, (ii) borderline examples, which are close to the boundary between negative and positive regions, (iii)  redundant samples, (iv) safe samples that are worth being kept for classification. Borderline and noisy cases are detected by Tomek links \citep{bib:Tomek}, whereas redundant cases are defined as  those not in a consistent subset\footnote{A  subset $C$ of the training set $S$ is said to be a consistent subset when  the Nearest Neighbour  rule using $C$ correctly classifies samples in $S$.} of the training set.
OSS creates a training set composed of safe samples from the majority class and all the samples in the minority one.

\subsubsection{Oversampling}
The oversamplimg methods considered in this work are {\em SMOTE}~\citep{bib:Chawla2002}, {\em SPIDER}~\citep{SPIDERds} and {\em ADASYN}~\citep{ADASYNds}. A comprehensive review on the extensions/variants of SMOTE can be found in ~\citet{0001GHC18}.

{\em SMOTE} stands for Synthetic Minority Over-Sampling Technique~\citep{bib:Chawla2002}, which is  an oversampling approach  for generating synthetic samples in the feature space along the line segments joining pairs of  minority class nearest neighbors~\citep{bib:Chawla2002}. Depending on the number of required samples to be generated, neighbors from the $k$ nearest neighbors are randomly chosen. The synthetic samples are generated by: (i) computing the difference between the feature vector under consideration and its nearest neighbour, (ii) multiplying the difference by a random number in $[0,1]$, (iii) adding this quantity to the feature vector under consideration. 

{\em SPIDER}~\citep{SPIDERds} is the acronym of Selective Preprocessing of Imbalanced Data. It oversamples instances of the minority class that are misclassified,  meanwhile filtering difficult examples from the majority classes~\citep{SPIDERds}. 

{\em ADASYN}~\citep{ADASYNds} is the abbreviation of ADAptive SYNthetic sampling. It is an improvement upon SMOTE, and the main idea is to use weighted distribution for different minority class examples according to their level of difficulty in learning. In ADASYN, more synthetic samples are generated for minority class examples that are harder to learn than those that are easier to learn. 


\section{Experimental Setup} \label{sec:ExperimentalFramework}

In this section, we first provide details about the datasets chosen for the experiments (subsection~\ref{sec:Dataset}), next describe the algorithms selected for this study and their configuration parameters (subsection~\ref{subsec:LearningParadigms}). Then in  subsections~\ref{subsec:EvaluationMetrics} and~\ref{subsec:StatisticalTests}, we present the evaluation metrics for imbalance classification and the statistical test measures to be adopted, respectively.

\subsection{Datasets} \label{sec:Dataset}
In our experiments, we use 52 datasets belonging to real-world problems which are publicly available in the UCI and KEEL repositories~\citep{bib:UCI,bib:KEEL}.  ~\tablename~\ref{tab:summaryDataset} provides a summary of these datasets. It also reports the imbalance ratio (IR) and the samples-to-features ratio  (SFR): the former is derived by $\frac{|N^{+}|}{|N^{-}|}$, while the latter is defined as  $\frac{|N^{+}|+|N^{-}|}{m}$, where $m$ is the original dimensionality of the dataset.

We use 5-fold stratified cross validation, considering the situation that re-sampling of the subsets introduces randomness. 

These \textit{52} datasets, together with the use of  \textit{3} different classification algorithms, \textit{9} feature selection methods, \textit{6} re-sampling methods and  with different orders in the pipelines, resulting in a very large number of experiments\footnote{We note that a small proportion of the experiments did not complete, primarily due to the reasons of out-of-memory crashes or overly long running time. The actual number of experiments is 9225. Notably, it took more than two months for these 9225 experiments to accomplish, even with a powerful server.}.

\subsection{Baseline classification algorithms} \label{subsec:LearningParadigms}
We adopt three well-established classification algorithms belonging to different paradigms~\citep{bib:ZhangESWA17}:  C4.5 algorithm, which is a well-known decision tree based classification algorithm~\cite{bib:qC45}, and SVM ~\citep{bib:libsvm} as a kernel machine, and Multi-Layer Perceptron (MLP), which is an established artificial neural network for classification. We set the classifier parameters to the default values used in the Weka library~\citep{bib:Weka} for the C4.5 and the MLP, while a Gaussian Radial Basis Function is adopted for SVM using the LIBSVM implementation~\citep{bib:libsvm}. 

Although we acknowledge that the tuning of the parameters in the algorithm may lead to better results, the No-Free-Lunch Theorems for Optimization tell us that all configurations perform equally well when averaged over all possible experiments~\citep{bib:NFL1997}. Therefore, we adopt the default parameter set, so that the only way a method can outperform the other(s) relies on its fitness to the dataset~\citep{bib:NFL4Dummies}. Our hypothesis is that the method which wins on average on all the experiments, would also win if a better setting is specified. Furthermore, in a framework where the classifier is not tuned, the top-performing methods tend to be the most robust, which is also a desirable characteristic~\citep{bib:fernandez2013}.

\subsection{Performance evaluation metrics for imbalance learning} \label{subsec:EvaluationMetrics}

Confusion matrix is commonly used to evaluate the performance of a classifier. According to the notation shown in  \tablename~\ref{tab:confusion-matrix}, the classification accuracy,  defined as $Acc = (TP + TN)/(|N^{+}| + |N^{-}|)$,  is the traditional evaluation measure for classification algorithms.  However, this metric tends to favour the majority class.

\begin{table}
  \centering
    \caption{Confusion matrix of a binary problem.}\label{tab:confusion-matrix}
 \resizebox{1.0\columnwidth}{!}{
  \begin{tabular}{l|cc} \toprule
	
	{} & Actual positive & Actual negative \\ \hline  
	
	Hypothesised positive & True Positive ($TP$) &  False Positive ($FP$) \TBstrut\\
	Hypothesised negative & False Negative ($FN$) &  True Negative ($TN$) \\ 
	\hline
	{\em Sum} & $|N^{+}|$ & $|N^{-}|$ \TBstrut\\
	 \bottomrule

  \end{tabular}}
\end{table}

It is often necessary to measure the per-class prediction accuracy/error. From \tablename~\ref{tab:confusion-matrix}, we compute two independent metrics that separately reflects the performance on the two classes, namely the  {\em true positive rate} or {\em recall} ($TPR = \frac{TP}{|N^{+}|}$) and the  {\em true negative rate} ($ TNR = \frac{TN}{|N^{-}|}$). They are independent of  prior probabilities and are  robust when class distributions are different in the training and testing sets. 

Upon $TPR$ and $TNR$, we can further compute the geometric mean of accuracies, F1 score and the index of balanced accuracy, which are three performance metrics widely used in the literature for imbalance learning. 
 
The geometric mean  of accuracies is defined as ${G\text{-}Mean} = \sqrt{TPR \cdot TNR}$.
It increases when the accuracy of each class increases.
It is also a non-linear measure since a change in one of the two parameters has a different effect on $G\text{-}Mean$, depending on its magnitude; for instance, the smaller the $TPR$ value, the larger the $G\text{-}Mean$ variation.
It is worth noting that  $G\text{-}Mean$ is closely related to the distance to the perfect classification in the ROC space~\citep{bib:Barandela-PAA}; therefore, we do not report the values of the area under the ROC curve.

F1 score, also known as $F$-measure, is defined as  the harmonic mean of precision ($p=\frac{TP}{TP+FP}$) and recall, i.e.  $ \mbox{F1}= \frac{2}{1/p + 1/TPR}$.

The index of balanced accuracy (IBA) was  introduced in~\citet{bib:garcia2009IBA} and it  is  computed as $IBA = (1 +\alpha \cdot (TPR - TNR)) \cdot TPR \cdot TNR$, with $\alpha$ set to 0.5 as suggested by the authors (note that if $\alpha= 0$, IBA turns into the square of $G\text{-}Mean$). 
 
The $IBA$ metric quantifies a certain trade-off between an unbiased measure of the overall accuracy and the index of how balanced the two class accuracies are. It intends to favour classifiers with better results on the minority class, which is the most important criteria in imbalance learning.

\subsection{Statistical tests} \label{subsec:StatisticalTests}

The extensive experimental tests result in such a huge amount of classification results that separate comparison analysis is practically infeasible. 
Instead, we need to use statistical analysis to find significant differences among the results obtained by the different approaches or pipelines. Moreover, according to the recommendations presented in~\citet{bib:demvsar2006statistical,bib:derrac2011practical}, we use non-parametric tests since the assumptions that guarantee the reliability of parametric tests (i.e., independence, normality and homoscedasticity) may not be satisfied, making the statistical analysis lose credibility. 

\textbf{Pairwise comparisons}. We will apply the Wilcoxon signed-rank test~\citep{bib:sheskin2003handbook} which performs pairwise comparisons between the two pipelines and computes the differences between their performance scores. The two pipelines are ranked according to their absolute values in ascending order, and average ranks are assigned in cases of ties. Let $R^+$  be the sum of ranks for the datasets on which  FS+DS outperforms DS+FS, and  $R^{-}$ be the sum of ranks for the opposite case (i.e., DS+FS outperforms FS+DS).  If $ min(R^{+}, R^{-})$ is less than or equal to the value of the distribution of Wilcoxon for $N_{d}$ degrees of freedom\footnote{$N_{d}$ stands for the number of datasets.}, the null hypothesis of equality of means is rejected at the specified level of significance. We set $\alpha = 0.1$ not only based on the suggestions reported in~\citet{bib:galar2011overview}, but also due to the large number of datasets and methods used in this work. To check whether two classification approaches are significantly different and the degree of their difference, we also compute the $p$-value associated with each comparison, which represents the lowest level of significance of a hypothesis that results in a rejection. 

To summarize the results, we also compute the quantity $\mu^{+}$ and $\mu^{-}$. The former denotes the number of times where   $R^{+}$  outperforms $R^{-}$ at the given significance level, while $\mu^{-}$  represents the opposite. These two values will be further normalized with respect to the number of comparisons. It provides a synthetic index representing the overall predominance of an approach with respect to the other.

\textbf{Heuristic measures for Top-K specific combinations selections}. Given a specified dataset and evaluation metric, it is often necessary and appealing to have the recommendations on the Top-K specific combinations that are most promising in yielding the best imbalance classification performance, which can substantially reduce the workload in finding the best solution to imbalanced data classification. To this end, we provide two heuristic measures for recommending the Top-K specific combinations for imbalance classification. 

In the experiments, there are 52 datasets, and it is common that some specific combinations (between the FS and DS methods) can only rank Top-K on some of the datasets. Our heuristic measure is to count the number of datasets (out of the 52 total datasets) that each specific combination can achieve the Top-K performance, in terms of a given evaluation metric (such as TPR). This heuristic measure is hereafter referred to as ``Rank-Sum''. The heuristic idea behind Rank-Sum is that the greater the  Rank-Sum value for a specific combination, the more frequently (likely) that this combination can become the Top-K performer on a new dataset. In our experiments, we set K to 3, to limit the number of choices.

Besides, we also introduce the ``Group-SUM'' measure to check the overall performance of all the Top-K combinations (as a group). When using Group-SUM, we try each of the Top-K combinations in the group,  then count the unique number of datasets where one of these Top-K combinations can become the Top-K method. We note that, in some cases, the Top-K combinations can be complementary, that is, only one of them can rank the Top-K methods on a dataset, while in other cases, they can be ``redundant'', e.g., several or all of them rank as the Top-K methods on a dataset. Therefore, if Group-SUM is close to the total sum of the Rank-Sum values of all the Top-K methods in the group, it implies these Top-K methods are highly complementary, which is more desirable. 

Finally, Rank-Sum and Group-SUM are the two heuristic measures we use for recommending the Top-K specific combinations.

\section{Results} \label{sec:Results}

Table~\ref{tab:tabella_risultati_C45},  and Tables~\ref{tab:tabella_risultati_SVM} and \ref{tab:tabella_risultati_MLP} in the appendix, report the Wilcoxon test results for pairwise comparisons between FS+DS and DS+FS when different classification algorithms are adopted. 
Table~\ref{tab:ShafferAccuracyIR_IBA_c45}, and Tables~\ref{tab:ShafferAccuracyIR_IBA_lib}, \ref{tab:ShafferAccuracyIR_IBA_mlp} and \ref{tab:ShafferAccuracyIR_IBA_mlp2} in the appendix, present the Iman-Davenport Test and Shaffer's post-hoc test results for the multiple comparisons among the specific combinations in both pipelines. 


\subsection{Pairwise comparisons using Wilxcon Test}
\subsubsection{Statistical comparison results between the two pipelines using Wilxcon Test}
\begin{figure*}[ht]
\centering
\resizebox{1.0\textwidth}{!}{
\includegraphics[]{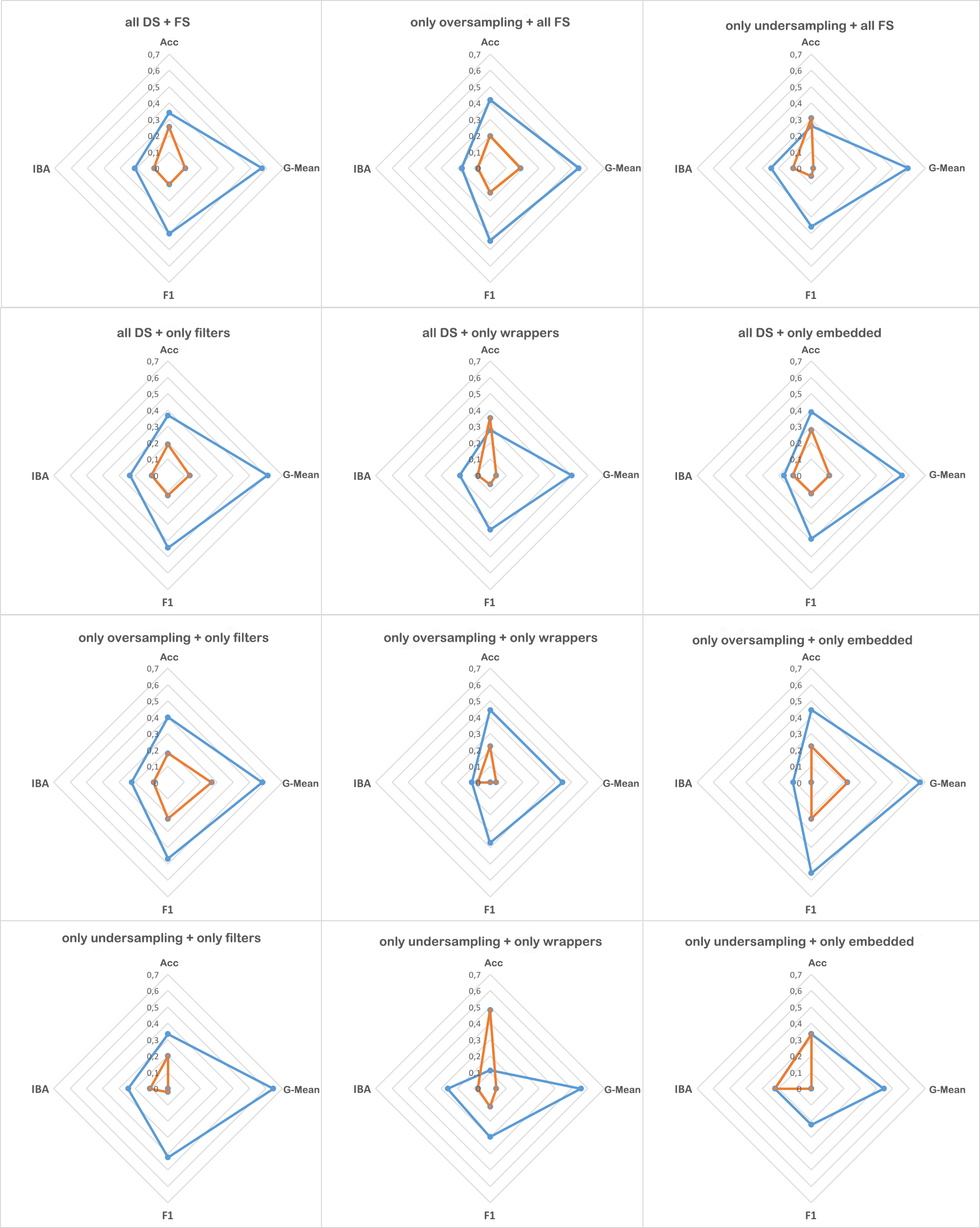}
}
\caption{Spider plot with values of  $\mu^{+}$ (in blue) and $\mu^{-}$ (in orange) for all the performance metrics; $\mu^{+}$ and $\mu^{-}$ correspond to  FS+DS and DS+FS, respectively.}
\label{fig:Script2Radar}
\end{figure*}

\figurename~\ref{fig:Script2Radar} summarizes via  spider plots the number of times where the ranks of the Wilcoxon signed-rank test of the classifiers trained with FS+DS  are larger than those obtained by DS+FS with the same classifier. In these charts, the blue and the orange lines correspond to FS+DS and  DS+FS, respectively. The plot/diagram in the upper left of the figure shows the radar plot attained considering all the DS and FS methods. The second plot in the first row reports what happens when only combining the oversampling approaches with all the feature selection methods. Similarly, the third plot in the same row reports what happens when we combine undersampling approaches with all the feature selection methods.

The second row of \figurename~\ref{fig:Script2Radar} shows the spider plots obtained when taking into account all the DS methods combined with different feature selection methods. Finally, the next six charts report the spider plots computed from the Cartesian product between the sets of DS and FS methods. For instance, the subfigure named as ``only oversampling + only filters'', which is shown as  the first plot in the third row, summarizes the number of times where the ranks of the Wilcoxon  signed-rank test of a classifier, trained on data preprocessed by an oversampling method followed by a filter feature selection method (DS+FS), are larger than those obtained by the same classifier but with the FS+DS pipeline (in specific, the filter feature selection before oversampling method).

Visual inspections of~\figurename~\ref{fig:Script2Radar}  show that FS+DS provides better  results than DS+FS in most of the cases,  irrespective of the performance metrics. In the case of $G\text{-}Mean$, $F1$ and $IBA$,  preprocessing the training set with the FS+DS pipeline provides better performance than the DS+FS counterpart, since the blue lines in the spider plots always run more external than the orange ones. But the opposite situation occurs in terms of  accuracy in three plots, which is expected, since it is known that many re-sampling approaches improve the recognition performance on the minority class but negatively affect the accuracy of the majority to some extent~\citep{bib:soda2011multiPR}.

Interestingly, the second and third plots in the first row also show that oversampling methods mitigate the decrease in accuracy with respect to undersampling, irrespective of the feature selection methods used.  Furthermore, the third and fourth rows show that FS+DS outperforms DS+FS by a more significant extent when using oversampling rather than undersampling.

\textbf{Summary}. Overall, there is no constant winner between the two pipelines. Thus, one should exhaustively test all the possible combinations of DS and FS, to derive the best imbalance classification model for a specific data/application. 

\onecolumn

\begin{table}[]
\centering
\caption{Wilcoxon test for the comparison between FS+DS ($R^{+}$) and DS+FS ($R^{-}$), when the C4.5 classifier is used.}
\label{tab:tabella_risultati_C45}
\resizebox{1\textwidth}{!}{
\begin{tabular}{llllllllllllll}
\toprule 
\toprule
\multirow{2}{*}{FS} & \multirow{2}{*}{DS} & \multicolumn{3}{l}{Accuracy (Acc)} & \multicolumn{3}{l}{G-Mean (g)} & \multicolumn{3}{l}{F1} & \multicolumn{3}{l}{IBA} \\ \cline{3-14} 
                    &                     & R+      & R-       & $p$-value     & R+      & R-     & $p$-value   & R+   & R-  & $p$-value & R+    & R-  & $p$-value \\ 
                   \hline

Info & RUS & 812 & 463 & 9.2e-02 & 578 & 748 & 4.3e-01 & 705 & 621 & 6.9e-01 & 457 & 869 & 5.3e-02 \\
ChiS & RUS & 619 & 707 & 6.8e-01 & 636 & 690 & 8.0e-01 & 595 & 731 & 5.2e-01 & 525 & 801 & 2.0e-01 \\
Fish & RUS & 804.5 & 470.5 & 1.1e-01 & 922 & 456 & 3.4e-02 & 960 & 418 & 1.4e-02 & 657 & 721 & 7.7e-01 \\
Gini & RUS & 729.5 & 596.5 & 5.3e-01 & 572 & 754 & 3.9e-01 & 557 & 769 & 3.2e-01 & 590 & 736 & 4.9e-01 \\
SBMLR & RUS & 602 & 479 & 5.0e-01 & 202 & 833 & 3.7e-04 & 222 & 813 & 8.5e-04 & 630 & 405 & 2.0e-01 \\
Ttest & RUS & 475 & 386 & 5.6e-01 & 345 & 645 & 8.0e-02 & 404 & 586 & 2.9e-01 & 453 & 493 & 8.1e-01 \\
RS & RUS & 599.5 & 675.5 & 7.1e-01 & 523 & 803 & 1.9e-01 & 517 & 809 & 1.7e-01 & 552 & 774 & 3.0e-01 \\
Linear & RUS & 648 & 678 & 8.9e-01 & 635 & 743 & 6.2e-01 & 643 & 735 & 6.8e-01 & 560 & 818 & 2.4e-01 \\
FWD & RUS & 842 & 484 & 9.3e-02 & 625 & 701 & 7.2e-01 & 776 & 499 & 1.8e-01 & 506 & 769 & 2.0e-01 \\
Info & CNN & 676 & 549 & 5.3e-01 & 334 & 891 & 5.6e-03 & 367 & 858 & 1.5e-02 & 705 & 520 & 3.6e-01 \\
ChiS & CNN & 650 & 526 & 5.2e-01 & 356 & 869 & 1.1e-02 & 384 & 841 & 2.3e-02 & 741 & 484 & 2.0e-01 \\
Fish & CNN & 682 & 543 & 4.9e-01 & 294 & 931 & 1.5e-03 & 343 & 882 & 7.3e-03 & 769 & 456 & 1.2e-01 \\
Gini & CNN & 625 & 600 & 9.0e-01 & 270 & 955 & 6.6e-04 & 278 & 947 & 8.8e-04 & 660 & 565 & 6.4e-01 \\
SBMLR & CNN & 89 & 992 & 8.1e-07 & 659 & 244 & 9.5e-03 & 840 & 241 & 1.1e-03 & 631 & 359 & 1.1e-01 \\
Ttest & CNN & 493 & 732 & 2.3e-01 & 262 & 819 & 2.3e-03 & 422 & 853 & 3.7e-02 & 683 & 398 & 1.2e-01 \\
RS & CNN & 824 & 352 & 1.5e-02 & 807 & 369 & 2.5e-02 & 854 & 322 & 6.4e-03 & 621 & 555 & 7.4e-01 \\
Linear & CNN & 825.5 & 302.5 & 5.7e-03 & 712 & 464 & 2.0e-01 & 778 & 398 & 5.1e-02 & 647 & 529 & 5.5e-01 \\
FWD & CNN & 253 & 875 & 1.0e-03 & 708 & 468 & 2.2e-01 & 704 & 472 & 2.3e-01 & 641 & 535 & 5.9e-01 \\
Info & OSS & 376 & 365 & 9.4e-01 & 268 & 552 & 5.6e-02 & 323 & 497 & 2.4e-01 & 506 & 314 & 2.0e-01 \\
ChiS & OSS & 501.5 & 444.5 & 7.3e-01 & 278 & 712 & 1.1e-02 & 367 & 623 & 1.4e-01 & 518 & 472 & 7.9e-01 \\
Fish & OSS & 448.5 & 454.5 & 9.7e-01 & 293 & 742 & 1.1e-02 & 361 & 674 & 7.7e-02 & 615 & 420 & 2.7e-01 \\
Gini & OSS & 362 & 673 & 7.9e-02 & 301 & 827 & 5.4e-03 & 327 & 801 & 1.2e-02 & 698 & 430 & 1.6e-01 \\
SBMLR & OSS & 227 & 98 & 8.3e-02 & 112 & 239 & 1.1e-01 & 134 & 217 & 2.9e-01 & 150 & 201 & 5.2e-01 \\
Ttest & OSS & 355.5 & 310.5 & 7.2e-01 & 209 & 421 & 8.3e-02 & 241 & 389 & 2.3e-01 & 383 & 283 & 4.3e-01 \\
RS & OSS & 611 & 565 & 8.1e-01 & 447 & 828 & 6.6e-02 & 501 & 774 & 1.9e-01 & 706 & 569 & 5.1e-01 \\
Linear & OSS & 599 & 577 & 9.1e-01 & 542 & 683 & 4.8e-01 & 561 & 664 & 6.1e-01 & 692 & 533 & 4.3e-01 \\
FWD & OSS & 456.5 & 173.5 & 2.0e-02 & 280 & 461 & 1.9e-01 & 329 & 412 & 5.5e-01 & 451 & 290 & 2.4e-01 \\
Info & SMOTE & 625 & 650 & 9.0e-01 & 669 & 657 & 9.6e-01 & 708 & 618 & 6.7e-01 & 776 & 550 & 2.9e-01 \\
ChiS & SMOTE & 633 & 745 & 6.1e-01 & 703 & 675 & 9.0e-01 & 688 & 690 & 9.9e-01 & 694 & 684 & 9.6e-01 \\
Fish & SMOTE & 842 & 484 & 9.3e-02 & 864 & 462 & 6.0e-02 & 883 & 443 & 3.9e-02 & 658 & 668 & 9.6e-01 \\
Gini & SMOTE & 795.5 & 582.5 & 3.3e-01 & 783 & 543 & 2.6e-01 & 789 & 537 & 2.4e-01 & 649 & 677 & 9.0e-01 \\
SBMLR & SMOTE & 340 & 521 & 2.4e-01 & 358 & 422 & 6.6e-01 & 344 & 397 & 7.0e-01 & 323 & 418 & 4.9e-01 \\
Ttest & SMOTE & 452 & 538 & 6.2e-01 & 635 & 355 & 1.0e-01 & 665 & 325 & 4.7e-02 & 616 & 287 & 4.0e-02 \\
RS & SMOTE & 492 & 886 & 7.3e-02 & 320 & 1058 & 7.8e-04 & 407 & 971 & 1.0e-02 & 623 & 755 & 5.5e-01 \\
Linear & SMOTE & 555 & 720 & 4.3e-01 & 504 & 822 & 1.4e-01 & 585.5 & 740.5 & 4.7e-01 & 637 & 638 & 1.0e+00 \\
FWD & SMOTE & 765.5 & 612.5 & 4.9e-01 & 718 & 608 & 6.1e-01 & 828 & 498 & 1.2e-01 & 735 & 591 & 5.0e-01 \\
Info & SPIDER & 290 & 340 & 6.8e-01 & 420 & 246 & 1.7e-01 & 375.5 & 254.5 & 3.2e-01 & 327 & 339 & 9.2e-01 \\
ChiS & SPIDER & 254.5 & 375.5 & 3.2e-01 & 381 & 249 & 2.8e-01 & 360 & 270 & 4.6e-01 & 232 & 398 & 1.7e-01 \\
Fish & SPIDER & 296.5 & 333.5 & 7.6e-01 & 351 & 315 & 7.8e-01 & 340 & 326 & 9.1e-01 & 296 & 370 & 5.6e-01 \\
Gini & SPIDER & 270 & 291 & 8.5e-01 & 385.5 & 244.5 & 2.5e-01 & 378.5 & 251.5 & 3.0e-01 & 246.5 & 383.5 & 2.6e-01 \\
SBMLR & SPIDER & 77.5 & 93.5 & 7.3e-01 & 59 & 94 & 4.1e-01 & 54 & 99 & 2.9e-01 & 64 & 89 & 5.5e-01 \\
Ttest & SPIDER & 127.5 & 172.5 & 5.2e-01 & 153 & 198 & 5.7e-01 & 132 & 219 & 2.7e-01 & 254 & 97 & 4.6e-02 \\
RS & SPIDER & 450 & 145 & 9.1e-03 & 485 & 218 & 4.4e-02 & 533 & 170 & 6.2e-03 & 340 & 363 & 8.6e-01 \\
Linear & SPIDER & 339 & 222 & 3.0e-01 & 369 & 297 & 5.7e-01 & 414 & 252 & 2.0e-01 & 306 & 360 & 6.7e-01 \\
FWD & SPIDER & 175.5 & 230.5 & 5.3e-01 & 158 & 307 & 1.3e-01 & 170 & 295 & 2.0e-01 & 252 & 213 & 6.9e-01 \\
Info & ADASYN & 210 & 1168 & 1.3e-05 & 1172 & 206 & 1.1e-05 & 913 & 465 & 4.1e-02 & 1048 & 330 & 1.1e-03 \\
ChiS & ADASYN & 285 & 1041 & 4.0e-04 & 1140 & 186 & 7.8e-06 & 904 & 422 & 2.4e-02 & 978 & 348 & 3.2e-03 \\
Fish & ADASYN & 546 & 780 & 2.7e-01 & 1203 & 123 & 4.2e-07 & 1180 & 146 & 1.3e-06 & 860 & 466 & 6.5e-02 \\
Gini & ADASYN & 356 & 1022 & 2.4e-03 & 1305 & 73 & 2.0e-08 & 1212 & 166 & 1.9e-06 & 766 & 612 & 4.8e-01 \\
SBMLR & ADASYN & 187 & 1088 & 1.4e-05 & 834 & 441 & 5.8e-02 & 659 & 616 & 8.4e-01 & 1126 & 149 & 2.4e-06 \\
Ttest & ADASYN & 437 & 739 & 1.2e-01 & 958 & 170 & 3.1e-05 & 921 & 207 & 1.6e-04 & 732 & 396 & 7.5e-02 \\
RS & ADASYN & 266.5 & 1111.5 & 1.2e-04 & 975 & 403 & 9.2e-03 & 565 & 813 & 2.6e-01 & 1090 & 288 & 2.6e-04 \\
Linear & ADASYN & 207.5 & 1118.5 & 2.0e-05 & 949 & 377 & 7.3e-03 & 492 & 834 & 1.1e-01 & 1055 & 271 & 2.4e-04 \\
FWD & ADASYN & 276 & 1102 & 1.7e-04 & 882 & 496 & 7.9e-02 & 783 & 595 & 3.9e-01 & 599 & 779 & 4.1e-01 \\
\bottomrule
\bottomrule
\end{tabular}
}
\end{table}

\twocolumn

\subsubsection{Pairwise comparisons of the two pipelines when using different classifiers}

\textbf{C4.5}. When using  the C4.5 decision tree as the learner,  experimental results in  \tablename~\ref{tab:tabella_risultati_C45} show that, in terms of $G\text{-}Mean$ and $F1$, DS+FS outperforms FS+DS in more cases than not.  
This result can be explained by the intrinsic capacity of a decision tree in choosing the relevant features during training. This trend is more significant when using oversampling rather than undersampling, suggesting that providing more synthetic minority class samples is beneficial to the training of the decision tree(s) for imbalance learning.

\textbf{SVM}. When the base classifier is SVM, as reported in \tablename~\ref{tab:tabella_risultati_SVM} in the appendix, our results show that  FS+DS outperforms DS+FS  when using the undersampling methods in more cases than not, in terms of $Acc$, $G\text{-}Mean$  and  $F1$. But with the oversampling methods, there is no clear winner between the two pipelines. It is important to note that, in terms of the $IBA$ metric, there is no clear dominance relationship between the two pipelines, either. This observation indicates that SVM is less sensitive to oversampling and undersampling. This is because SVM algorithm looks for the support vectors inherently, which are  (possibly) a small set of samples determining the boundary in the feature space induced by the given kernel satisfying an optimization function and certain constraints. Oversampling may introduce synthetic samples that will not alter the set of support vectors, whilst undersampling may remove samples that will not become the support vectors. 

\textbf{MLP}. In case of the MLP classification algorithm, shown in \tablename~\ref{tab:tabella_risultati_MLP} in the appendix, we notice a predominance of FS+DS over DS+FS for all the performance metrics. Contrary to the previous case, the supremacy of FS+DS over DS+FS is more remarkable in undersampling than oversampling. In terms of $G\text{-}Mean$, $F1$, and $IBA$, we see that with undersampling data selection methods,  FS+DS is often better than the DS+FS pipeline in most cases. The topology of the artificial neural network (MLP) structure is fixed, and the elimination of features prior to data sampling process simplifies the number of input neurons, leading to faster convergence in learning. Moreover, a better feature selection method should be able to find a better feature representation (a subset of features) of the data, bringing better convergence in learning. 

\textbf{Short summary}.  For MLP and SVM, with oversampling data re-sampling methods, the dominance between the two pipelines is uncertain; but with undersampling, the FS+DS pipeline often outperforms DS+FS. However, for the decision tree classifier, DS+FS usually outperforms FS+DS, especially with oversampling methods. Therefore, the conclusions concerning the performance of the two pipelines using different classification algorithms (decision tree classifier v.s. MLP or SVM) can be quite different, even contrary.

\subsubsection{Insights into the influence of IR and SFR on the two pipelines}
We also investigate how the results vary with the imbalance ratio (IR) and the samples-to-features ratio (SFR). As mentioned before, IR is the ratio between the cardinality of the majority and the minority class, whereas SFR is the ratio between the number of samples and the number of features. For each of these two factors/ratios, the experimental  datasets were divided in three groups:   the first contains those falling below the $33^{{th}}$ percentile, the second includes those between the $34^{{th}}$ and the $66^{{th}}$ percentile, and the third consists of those falling above the $66^{{th}}$ percentile. The values of the $33^{{th}}$ and the $66^{{th}}$ percentile for IR  are 6.01 and 9.18; but for SFR, the values are 37.3 and 115.09, indicating that SFR may have a significant influence on the performance of the pipelines.

\begin{figure*}[htb]
\centering
\resizebox{1\textwidth}{!}{
\includegraphics{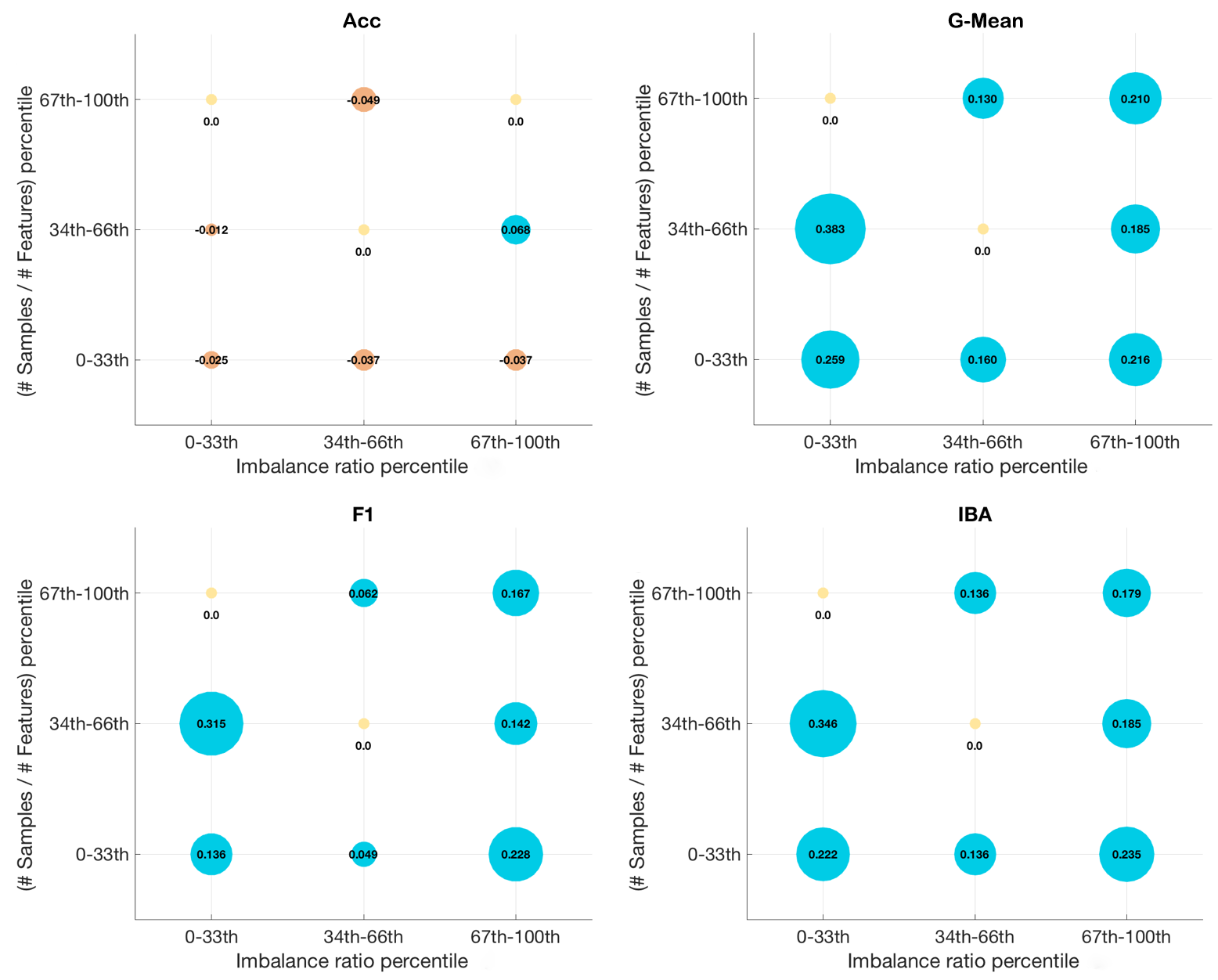}}
\caption{Visual analysis of the predominance of FS+DS w.r.t. DS+FS for different pairs of IR and SFR, for each metric considered.}
\label{fig:Script2_Bolle}
\end{figure*}

\figurename~\ref{fig:Script2_Bolle} offers a visual presentation of the predominance of FS+DS over DS+FS for different pairs of IR and SFR, with different metrics considered.
The predominance is measured in terms of $\mu^{+}$ ($\mu^{-}$) which calculates the number of times where   $R^{+}$  outperforms (underperforms) $R^{-}$ at the given significance level. On these grounds, given one of the three possible values of the  percentile for both IR and SFR, the ball plots in the figure visualize the rate of predominance of FS+DS w.r.t. DS+FS. Positive values, colored in blue, denote that the former outperforms the latter, whereas negative values, colored in orange, have the opposite meaning. The results show that FS+DS outperforms DS+FS in general, in terms of the $G\text{-}Mean$, $F1$, and $IBA$. But in terms of $Acc$, we observe the predominance of  DS+FS over FS+DS. 


While these observations hold on datasets with  high imbalance ratios, for datasets with large value of the SFR but small value of IR (see the upper left part of the plots), we notice that there is no predominance between the two pipelines. Such datasets are characterized by high SFRs and low IRs (the number of training samples is significantly larger than the number of features, while the ratio between the number of samples of majority and minority classes is not very large).  Similar observations also hold for datasets with medium SFR and IR values, as seen from the central bubble in each plot of \figurename~\ref{fig:Script2_Bolle}.

\textbf{Summary}. For datasets with high SFR but relatively low IR ratios, the FS+DS and DS+FS pipelines return low or zero values of $\mu^{+}$ and $\mu^{-}$ and it is hard to tell which pipeline is better. But for datasets with low SFR but high IR ratios, the FS+DS pipeline has a comparative advantage over the DS+FS pipeline. However, once again, there is no constant winner between them.

\subsection{Detailed multiple comparisons among the specific combinations in the two pipelines using Iman-Davenport Test} \label{imantest}

Above, we have investigated the overall performance of the two general pipelines. In this section, we will look for specific combinations between the FS and DS methods to recommend the Top-K methods for a new dataset/application with skewed distributions.  We clarify that by ``pipleline'', we denote the two general frameworks of putting feature selection before and after data re-sampling for imbalance classification, while by ``specific combination'', we denote a specified pair of  FS and DS methods, e.g., the combination of ``Linear+SMOTE'' using the C4.5 classifier. But for each specific combination, we should still take the two general pipelines in two considerations.  For instance, the specific combination of Linear (FS) and SMOTE (DS) results in two specific methods, which are Linear+SMOTE and SMOTE+Linear.

The results reported so far are computed considering all the base classifiers tested and all the data sampling methods and feature selection approaches. In the following, we will focus on the specific results for each pair/combination of the data sampling and feature selection methods: for each pair of combination, we run multiple comparisons by using the Iman-Davenport test and the  Shaffer's post-hoc test. This enables us to figure out which combinations are significantly better/worse among the $n \times n$ comparisons, and it could help researchers and practitioners with their choices of algorithms when facing imbalanced datasets. We use the $IBA$ metric since it is a metric that quantifies a  trade-off between an unbiased measure of overall accuracy and an index of how balanced the per-class accuracies are. 

The results for each classifier are presented in Tables~\ref{tab:ShafferAccuracyIR_IBA_c45}, \ref{tab:ShafferAccuracyIR_IBA_lib}, and \ref{tab:ShafferAccuracyIR_IBA_mlp} and \ref{tab:ShafferAccuracyIR_IBA_mlp2}, which are vertically divided into three percentile groups of imbalance ratios. The tables are also horizontally split into two parts: the left part reports the results with the oversampling methods, whilst the right part shows the results with the undersampling data selection methods. We also take imbalance ratio and the types of feature selection approaches into consideration. 
Each section of the tables contains three columns: the first column reports the specific combination/pair which achieves at least one comparison where its un-adjusted $p$-value  is smaller than the un-adjusted $p$-value given by the Shaffer's procedure, i.e., Shaffer's post-hoc test rejects such hypotheses. The second column reports the average values of the ranks obtained by applying the Iman-Davenport test (the smaller the rank, the better the specific combination/pair in the  $n \times n$ comparisons). Finally, the third column counts for each feature selection type, the number of pairwise comparisons in the Shaffer's post-hoc test where the considered combination/pair has an unadjusted $p$-value smaller than Shaffer's procedure $p$-value: the larger this number, the more times (the larger proportions) the specific combination/pair wins against the others. 

\begin{table*}
\centering
\caption{Results of the Iman-Davenport test and the Shaffer's post-hoc test described in subsection~\ref{subsec:StatisticalTests} when the C4.5 classifier is used as the base learner. The tests are applied to classification results computed in terms of IBA. The first column ``method'' in each section  denotes the name of a specific combination of the FS+DS or DS+FS pipeline,  e.g., ``ChiS+SMOTE''. The  second column ``Rank'' reports the average ranks of the specific combination when using the  Iman-Davenport Test. The smaller the rank, the better the pipeline in the  $n \times n$ comparisons in the  Iman-Davenport Test.  The third column ``$\#p$'' calculates the number of pairwise comparisons in the Shaffer's Post-hoc Test where the corresponding pipeline has an unadjusted $p$-value smaller than Shaffer's procedure $p$-value. The larger this  $\#p$ value, the more times that the specific combination  (e.g., ``ChiS+SMOTE'')  wins against the others.   }
\label{tab:ShafferAccuracyIR_IBA_c45}  
\small
  \begin{tabular}{lcllllll}
  \hline 
  \hline
	\multicolumn{2}{l}{\multirow{2}{*}{\diagbox[width=3.1cm,height=0.8cm]{\raisebox{0.2ex}{FS}}{\raisebox{0.2ex}{DS}}}} & \multicolumn{3}{c}{Oversampling} & \multicolumn{3}{c}{Undersampling}\\
	\cline{3-8}
	\multicolumn{2}{l}{}
	                    & Method      & Rank  & $\#p$   & Method      & Rank  & $\#p$ \\
	\hline
	\multirow{17}{*}{\rotatebox[origin=c]{90}{0-33th IR Percentile}}    & \multirow{11}{*}{Filter} &  ChiS+SMOTE        & 6.2353  & 6  & RUS+Gini        & 6.1765 & 7 \TBstrut\\
	                                          &                          & Info+SMOTE         & 6.5882  & 6  & Info+RUS        & 6.5882 & 6 \\
	                                          &                          & Gini+ADASYN        & 7.4706  & 5  & RUS+Info        & 6.5882 & 6 \\
	                                          &                          & Fish+ADASYN        & 7.9412  & 5  & RUS+ChiS        & 6.7059 & 5 \\
	                                          &                          & ChiS+ADASYN        & 8.2941  & 4  & Fish+RUS        & 6.8824 & 5 \\
	                                          &                          & Info+ADASYN        & 8.4706  & 4  & Gini+RUS        & 7.5    & 2 \\
	                                          &                          & SMOTE+ChiS         & 8.7059  & 3  & ChiS+RUS        & 7.5588 & 1 \\
	                                          &                          & Fish+SMOTE         & 8.7647  & 3  & RUS+Fish        & 8      & 1 \\
	                                          &                          & Gini+SMOTE         & 8.7941  & 3  &                 &        &   \\
	                                          &                          & SMOTE+Info         & 9.2059  & 1  &                 &        &   \\
	                                          &                          & SMOTE+Gini         & 10.2647 & 1  &                 &        &   \\
  \cline{2-8}
                                            & \multirow{6}{*}{Wrapper} & SMOTE+RS   & 3.6765  & 4  & RS+RUS  & 3.5294 & 6 \TBstrut\\
                                            &                          & SMOTE+Linear        & 3.9118  & 4  & RUS+RS  & 3.6176 & 6 \\
                                            &                          & Linear+SMOTE        & 5.1765  & 2  & RUS+Linear       & 3.6176 & 6 \\
                                            &                          & RS+SMOTE   & 5.7059  & 2  & Linear+RUS       & 4.4706 & 5 \\
                                            &                          & Linear+ADASYN       & 6.0294  & 2  &                 &        &   \\
                                            &                          & RS+ADASYN  & 6.1765  & 2  &                 &        &   \\ 
  \midrule
  \midrule
	\multirow{12}{*}{\rotatebox[origin=c]{90}{34th-66th IR Percentile}}     & \multirow{8}{*}{Filter}    & SMOTE+ChiS        & 6.3235  & 6  & ChiS+RUS        & 5.5882 & 5 \TBstrut\\
	                                             &                        & SMOTE+Info        & 6.5588  & 4  & Fish+RUS        & 5.7647 & 5 \\
	                                             &                        & Info+SMOTE        & 7.2353  & 4  & RUS+Gini        & 5.8529 & 5 \\
	                                             &                        & ChiS+SMOTE        & 7.4412  & 3  & RUS+Info        & 6.1765 & 5 \\
	                                             &                        & Info+ADASYN       & 8.5882  & 1  & Gini+RUS        & 6.2941 & 5 \\
	                                             &                        &                   &         &    & RUS+ChiS        & 6.4706 & 5 \\
	                                             &                        &                   &         &    & Info+RUS        & 7      & 5 \\
	                                             &                        &                   &         &    & RUS+Fish        & 7.4118 & 5 \\
	\cline{2-8}
	                                             & \multirow{4}{*}{Wrapper}   & SMOTE+RS  & 2.7941  & 9  & Linear+RUS       & 2.8235 & 8 \TBstrut\\
	                                             &                        & SMOTE+Linear       & 2.7941  & 9  & RUS+Linear       & 3      & 8 \\
	                                             &                        & Linear+SMOTE       & 3.6471  & 5  & RUS+RS  & 3.2353 & 8 \\
	                                             &                        & RS+SMOTE  & 4.4706  & 2  & RS+RUS  & 3.7353 & 8 \\
	\midrule
	\midrule
	\multirow{13}{*}{\rotatebox[origin=c]{90}{67th-100th IR Percentile}}    & \multirow{8}{*}{Filter}    & SMOTE+ChiS        & 6.1944  & 3  & RUS+Info        & 4.75   & 12\TBstrut\\
	                                             &                        & Info+SMOTE        & 6.8611  & 3  & RUS+ChiS        & 4.8333 & 12\\
	                                             &                        & SMOTE+Info        & 7.2222  & 3  & RUS+Gini        & 6.3889 & 7 \\
	                                             &                        & SMOTE+Gini        & 7.5556  & 3  & Info+RUS        & 6.4167 & 7 \\
	                                             &                        & ChiS+SMOTE        & 7.6667  & 3  & Fish+RUS        & 6.6111 & 7 \\
	                                             &                        & Gini+SMOTE        & 8.1389  & 2  & ChiS+RUS        & 7.1667 & 6 \\
	                                             &                        & Fish+SMOTE        & 9.9722  & 2  & RUS+Fish        & 8.3889 & 1 \\
	                                             &                        & Gini+ADASYN       & 10.1111 & 1  & Gini+RUS        & 8.4167 & 1 \\
	\cline{2-8}
	                                             & \multirow{5}{*}{Wrapper}   & Linear+SMOTE       & 2.9444  & 5  & Linear+RUS       & 2.5556 & 6 \TBstrut\\
	                                             &                        & SMOTE+RS  & 3.8333  & 4  & RUS+RS  & 3.4167 & 6 \\
	                                             &                        & SMOTE+Linear       & 3.8889  & 3  & RUS+Linear       & 3.5556 & 6 \\
	                                             &                        & RS+SMOTE  & 4.6667  & 3  & RS+RUS  & 3.8333 & 6 \\
	                                             &                        & Linear+ADASYN      & 7.0833  & 2  &                 &        &   \\  
  \bottomrule
  \bottomrule
  \end{tabular}
\end{table*}

\tablename~\ref{tab:ShafferAccuracyIR_IBA_c45} reveals that, if a dataset has an imbalance ratio lower than 6, and if one needs to use a filter to select the features and perform oversampling, the method that often provides the best performance is first applying  $\chi^{2}$ feature selection, next using the SMOTE oversampling method. Moreover, for both C4.5 and MLP classifiers, as can be observed from \tablename~\ref{tab:ShafferAccuracyIR_IBA_c45}, and \tablename~\ref{tab:ShafferAccuracyIR_IBA_mlp} and \ref{tab:ShafferAccuracyIR_IBA_mlp2} in the appendix, SMOTE and RUS seem to be two ideal oversampling and undersampling methods, respectively; but different feature selection methods still need to be tested to find the best combination between DS and FS. We note that, when the columns in each section of the tables are empty, this means that we do not observe any outperforming method (e.g., the wrapper and oversampling combination for datasets having IR percentile between 34 and 66 using the MLP). In this case, our results suggest that researchers should extensively compare all the possible combinations between the FS and DS methods through exhaustive experiments to determine which specific combination is the best for imbalance classification.

Overall, such specific inspections and statistical analysis should give practitioners and researchers new reference value when tackling class imbalance.

\subsection{Recommending the Top-K specific methods for imbalance classification  using Rank-Sum}

It is usually tempting for researchers and practitioners to have practical suggestions and guidelines about the specific top performing combinations for imbalance classification on their particular data. In this context, we make a heuristic but intuitive analysis on the top-3 best performing combinations between the FS and DS methods, using the heuristic measures we introduced for Top-K specific combination selection in subsection \ref{subsec:StatisticalTests}.

For each specific combination between the FS and DS methods (e.g.,  ``Linear+SMOTE'' with the  C4.5 classifier), and under every different evaluation metric (e.g., $G\text{-}Mean$),  we calculate its $Rank\text{-}Sum$, where we count the number of datasets where this combination ranks among the top-3 best performers. Then, for this metric, we select the group of the top-3 specific  combinations in terms of $Rank\text{-}Sum$ (i.e. ``Linear+SMOTE'', ``RS+SMOTE'', and ``Linear+RUS''), then calculate their \textit{$Group\text{-}Sum$}. Note that, to facilitate the comparison of top performers under both pipelines, we still put the feature selection methods before the data re-sampling methods in the DS+FS pipeline. 

\begin{figure*}[htb]
\label{fig:sum_c45_fsds}
\resizebox{1.0\textwidth}{!}
{
\begin{tikzpicture}
\begin{scope}[local bounding box=gbox]
	\begin{axis}[
			title  = \textbf{Acc},
			xbar stacked,
			height = 40mm,
			width=0.3\textwidth,
			enlarge x limits  = 0.02,
			enlarge y limits={abs=0.2},
			bar width=2mm,
	    axis y line*=none,
	    axis x line*=none,
	    y axis line style = { opacity = 0 },
	    x axis line style = { opacity = 0 },
	    tickwidth = 0pt,
	    xtick=\empty,
	    ytick=data,
	    yticklabels={$Group\text{-}Sum$, RS+OSS, RS+SPIDER, Linear+SPIDER},
	    xmin=0,
	    xmax=52,
	    nodes near coords,
	    nodes near coords style={text=black, at ={(\pgfplotspointmeta,\pgfplotspointy)},anchor=west},
	    visualization depends on=y \as \pgfplotspointy,
	    every axis plot/.append style={fill},
	]
	\addplot[fill=red] coordinates {(32,0) (0,1) (0,2) (0,3)};
	\addplot[fill=blue!30!white] coordinates {(0,0) (16,1) (0,2) (0,3)};
	\addplot[fill=blue!30!white] coordinates {(0,0) (0,1) (25,2) (0,3)};
	\addplot[fill=blue!30!white] coordinates {(0,0) (0,1) (0,2) (26,3)};
\end{axis}  
\end{scope}

\begin{scope}[xshift=6cm,local bounding box=gbox]
	\begin{axis}[
			title  = \textbf{TPR},
			xbar stacked,
			height = 40mm,
			width=0.3\textwidth,
			enlarge x limits  = 0.02,
			enlarge y limits={abs=0.2},
			bar width=2mm,
	    axis y line*=none,
	    axis x line*=none,
	    y axis line style = { opacity = 0 },
	    x axis line style = { opacity = 0 },
	    tickwidth = 0pt,
	    xtick=\empty,
	    ytick=data,
	    yticklabels={$Group\text{-}Sum$, Linear+RUS, Gini+ADASYN, SBMLR+ADASYN},
	    xmin=0,
	    xmax=52,
	    nodes near coords,
	    nodes near coords style={text=black, at ={(\pgfplotspointmeta,\pgfplotspointy)},anchor=west},
	    visualization depends on=y \as \pgfplotspointy,
	    every axis plot/.append style={fill},
	]
	\addplot[fill=red] coordinates {(32,0) (0,1) (0,2) (0,3)};
	\addplot[fill=blue!30!white] coordinates {(0,0) (12,1) (0,2) (0,3)};
	\addplot[fill=blue!30!white] coordinates {(0,0) (0,1) (13,2) (0,3)};
	\addplot[fill=blue!30!white] coordinates {(0,0) (0,1) (0,2) (16,3)};
\end{axis}  
\end{scope}

\begin{scope}[xshift=12cm,local bounding box=gbox]
	\begin{axis}[
			title  = \textbf{IBA},
			xbar stacked,
			height = 40mm,
			width=0.3\textwidth,
			enlarge x limits  = 0.02,
			enlarge y limits={abs=0.2},
			bar width=2mm,
	    axis y line*=none,
	    axis x line*=none,
	    y axis line style = { opacity = 0 },
	    x axis line style = { opacity = 0 },
	    tickwidth = 0pt,
	    xtick=\empty,
	    ytick=data,
	    yticklabels={$Group\text{-}Sum$, RS+SMOTE, Linear+RUS, Linear+SMOTE},
	    xmin=0,
	    xmax=52,
	    nodes near coords,
	    nodes near coords style={text=black, at ={(\pgfplotspointmeta,\pgfplotspointy)},anchor=west},
	    visualization depends on=y \as \pgfplotspointy,
	    every axis plot/.append style={fill},
	]
	\addplot[fill=red] coordinates {(28,0) (0,1) (0,2) (0,3)};
	\addplot[fill=blue!30!white] coordinates {(0,0) (15,1) (0,2) (0,3)};
	\addplot[fill=blue!30!white] coordinates {(0,0) (0,1) (15,2) (0,3)};
	\addplot[fill=blue!30!white] coordinates {(0,0) (0,1) (0,2) (18,3)};
\end{axis}  
\end{scope}

\begin{scope}[xshift=18cm,local bounding box=gbox]
	\begin{axis}[
			title  = \textbf{G-Mean},
			xbar stacked,
			height = 40mm,
			width=0.3\textwidth,
			enlarge x limits  = 0.02,
			enlarge y limits={abs=0.2},
			bar width=2mm,
	    axis y line*=none,
	    axis x line*=none,
	    y axis line style = { opacity = 0 },
	    x axis line style = { opacity = 0 },
	    tickwidth = 0pt,
	    xtick=\empty,
	    ytick=data,
	    yticklabels={$Group\text{-}Sum$, Linear+RUS, RS+SMOTE, Linear+SMOTE},
	    xmin=0,
	    xmax=52,
	    nodes near coords,
	    nodes near coords style={text=black, at ={(\pgfplotspointmeta,\pgfplotspointy)},anchor=west},
	    visualization depends on=y \as \pgfplotspointy,
	    every axis plot/.append style={fill},
	]
	\addplot[fill=red] coordinates {(28,0) (0,1) (0,2) (0,3)};
	\addplot[fill=blue!30!white] coordinates {(0,0) (13,1) (0,2) (0,3)};
	\addplot[fill=blue!30!white] coordinates {(0,0) (0,1) (17,2) (0,3)};
	\addplot[fill=blue!30!white] coordinates {(0,0) (0,1) (0,2) (21,3)};
\end{axis}  
\end{scope}

\begin{scope}[xshift=24cm,local bounding box=gbox]
	\begin{axis}[
			title  = \textbf{F1},
			xbar stacked,
			height = 40mm,
			width=0.3\textwidth,
			enlarge x limits  = 0.02,
			enlarge y limits={abs=0.2},
			bar width=2mm,
	    axis y line*=none,
	    axis x line*=none,
	    y axis line style = { opacity = 0 },
	    x axis line style = { opacity = 0 },
	    tickwidth = 0pt,
	    xtick=\empty,
	    ytick=data,
	    yticklabels={$Group\text{-}Sum$, Linear+SPIDER, RS+SMOTE, Linear+SMOTE},
	    xmin=0,
	    xmax=52,
	    nodes near coords,
	    nodes near coords style={text=black, at ={(\pgfplotspointmeta,\pgfplotspointy)},anchor=west},
	    visualization depends on=y \as \pgfplotspointy,
	    every axis plot/.append style={fill},
	]
	\addplot[fill=red] coordinates {(34,0) (0,1) (0,2) (0,3)};
	\addplot[fill=blue!30!white] coordinates {(0,0) (13,1) (0,2) (0,3)};
	\addplot[fill=blue!30!white] coordinates {(0,0) (0,1) (15,2) (0,3)};
	\addplot[fill=blue!30!white] coordinates {(0,0) (0,1) (0,2) (23,3)};
\end{axis}  
\end{scope}

\end{tikzpicture}
}
\caption{Rank-Sum results for the Top-3 specific combinations under the FS+DS pipeline, using the C4.5 classifier}
\label{gf:1}
\end{figure*}
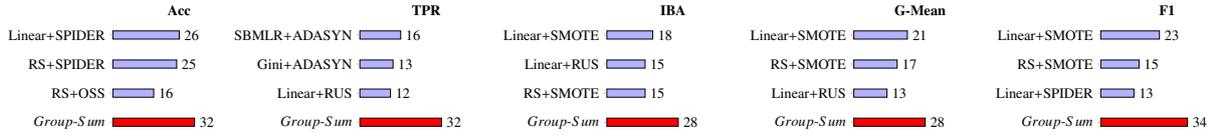

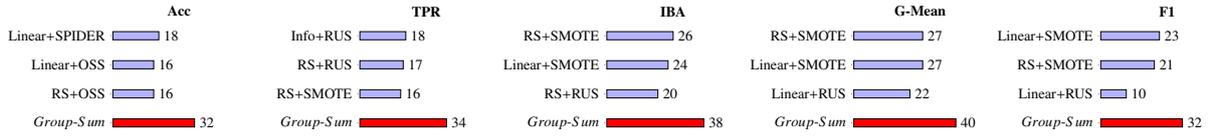
\begin{figure*}[htb]
\label{fig:sum_c45_dsfs}
\resizebox{1.0\textwidth}{!}
{
\begin{tikzpicture}
\begin{scope}[local bounding box=gbox]
	\begin{axis}[
			title  = \textbf{Acc},
			xbar stacked,
			height = 40mm,
			width=0.3\textwidth,
			enlarge x limits  = 0.02,
			enlarge y limits={abs=0.2},
			bar width=2mm,
	    axis y line*=none,
	    axis x line*=none,
	    y axis line style = { opacity = 0 },
	    x axis line style = { opacity = 0 },
	    tickwidth = 0pt,
	    xtick=\empty,
	    ytick=data,
	    yticklabels={$Group\text{-}Sum$, RS+OSS, Linear+OSS, Linear+SPIDER},
	    xmin=0,
	    xmax=52,
	    nodes near coords,
	    nodes near coords style={text=black, at ={(\pgfplotspointmeta,\pgfplotspointy)},anchor=west},
	    visualization depends on=y \as \pgfplotspointy,
	    every axis plot/.append style={fill},
	]
	\addplot[fill=red] coordinates {(32,0) (0,1) (0,2) (0,3)};
	\addplot[fill=blue!30!white] coordinates {(0,0) (16,1) (0,2) (0,3)};
	\addplot[fill=blue!30!white] coordinates {(0,0) (0,1) (16,2) (0,3)};
	\addplot[fill=blue!30!white] coordinates {(0,0) (0,1) (0,2) (18,3)};
\end{axis}  
\end{scope}

\begin{scope}[xshift=6cm,local bounding box=gbox]
	\begin{axis}[
			title  = \textbf{TPR},
			xbar stacked,
			height = 40mm,
			width=0.3\textwidth,
			enlarge x limits  = 0.02,
			enlarge y limits={abs=0.2},
			bar width=2mm,
	    axis y line*=none,
	    axis x line*=none,
	    y axis line style = { opacity = 0 },
	    x axis line style = { opacity = 0 },
	    tickwidth = 0pt,
	    xtick=\empty,
	    ytick=data,
	    yticklabels={$Group\text{-}Sum$, RS+SMOTE, RS+RUS, Info+RUS},
	    xmin=0,
	    xmax=52,
	    nodes near coords,
	    nodes near coords style={text=black, at ={(\pgfplotspointmeta,\pgfplotspointy)},anchor=west},
	    visualization depends on=y \as \pgfplotspointy,
	    every axis plot/.append style={fill},
	]
	\addplot[fill=red] coordinates {(34,0) (0,1) (0,2) (0,3)};
	\addplot[fill=blue!30!white] coordinates {(0,0) (16,1) (0,2) (0,3)};
	\addplot[fill=blue!30!white] coordinates {(0,0) (0,1) (17,2) (0,3)};
	\addplot[fill=blue!30!white] coordinates {(0,0) (0,1) (0,2) (18,3)};
\end{axis}  
\end{scope}

\begin{scope}[xshift=12cm,local bounding box=gbox]
	\begin{axis}[
			title  = \textbf{IBA},
			xbar stacked,
			height = 40mm,
			width=0.3\textwidth,
			enlarge x limits  = 0.02,
			enlarge y limits={abs=0.2},
			bar width=2mm,
	    axis y line*=none,
	    axis x line*=none,
	    y axis line style = { opacity = 0 },
	    x axis line style = { opacity = 0 },
	    tickwidth = 0pt,
	    xtick=\empty,
	    ytick=data,
	    yticklabels={$Group\text{-}Sum$, RS+RUS, Linear+SMOTE, RS+SMOTE},
	    xmin=0,
	    xmax=52,
	    nodes near coords,
	    nodes near coords style={text=black, at ={(\pgfplotspointmeta,\pgfplotspointy)},anchor=west},
	    visualization depends on=y \as \pgfplotspointy,
	    every axis plot/.append style={fill},
	]
	\addplot[fill=red] coordinates {(38,0) (0,1) (0,2) (0,3)};
	\addplot[fill=blue!30!white] coordinates {(0,0) (20,1) (0,2) (0,3)};
	\addplot[fill=blue!30!white] coordinates {(0,0) (0,1) (24,2) (0,3)};
	\addplot[fill=blue!30!white] coordinates {(0,0) (0,1) (0,2) (26,3)};
\end{axis}  
\end{scope}

\begin{scope}[xshift=18cm,local bounding box=gbox]
	\begin{axis}[
			title  = \textbf{G-Mean},
			xbar stacked,
			height = 40mm,
			width=0.3\textwidth,
			enlarge x limits  = 0.02,
			enlarge y limits={abs=0.2},
			bar width=2mm,
	    axis y line*=none,
	    axis x line*=none,
	    y axis line style = { opacity = 0 },
	    x axis line style = { opacity = 0 },
	    tickwidth = 0pt,
	    xtick=\empty,
	    ytick=data,
	    yticklabels={$Group\text{-}Sum$, Linear+RUS, Linear+SMOTE, RS+SMOTE},
	    xmin=0,
	    xmax=52,
	    nodes near coords,
	    nodes near coords style={text=black, at ={(\pgfplotspointmeta,\pgfplotspointy)},anchor=west},
	    visualization depends on=y \as \pgfplotspointy,
	    every axis plot/.append style={fill},
	]
	\addplot[fill=red] coordinates {(40,0) (0,1) (0,2) (0,3)};
	\addplot[fill=blue!30!white] coordinates {(0,0) (22,1) (0,2) (0,3)};
	\addplot[fill=blue!30!white] coordinates {(0,0) (0,1) (27,2) (0,3)};
	\addplot[fill=blue!30!white] coordinates {(0,0) (0,1) (0,2) (27,3)};
\end{axis}  
\end{scope}

\begin{scope}[xshift=24cm,local bounding box=gbox]
	\begin{axis}[
			title  = \textbf{F1},
			xbar stacked,
			height = 40mm,
			width=0.3\textwidth,
			enlarge x limits  = 0.02,
			enlarge y limits={abs=0.2},
			bar width=2mm,
	    axis y line*=none,
	    axis x line*=none,
	    y axis line style = { opacity = 0 },
	    x axis line style = { opacity = 0 },
	    tickwidth = 0pt,
	    xtick=\empty,
	    ytick=data,
	    yticklabels={$Group\text{-}Sum$, Linear+RUS, RS+SMOTE, Linear+SMOTE},
	    xmin=0,
	    xmax=52,
	    nodes near coords,
	    nodes near coords style={text=black, at ={(\pgfplotspointmeta,\pgfplotspointy)},anchor=west},
	    visualization depends on=y \as \pgfplotspointy,
	    every axis plot/.append style={fill},
	]
	\addplot[fill=red] coordinates {(32,0) (0,1) (0,2) (0,3)};
	\addplot[fill=blue!30!white] coordinates {(0,0) (10,1) (0,2) (0,3)};
	\addplot[fill=blue!30!white] coordinates {(0,0) (0,1) (21,2) (0,3)};
	\addplot[fill=blue!30!white] coordinates {(0,0) (0,1) (0,2) (23,3)};
\end{axis}  
\end{scope}

\end{tikzpicture}
}
\caption{Rank-Sum results for the Top-3 specific combinations under the DS+FS pipeline, using the C4.5 classifier}
\label{gf:2}
\end{figure*}

\textbf{C4.5}. Figure \ref{gf:1} presents the top-3 specific combinations between the DS and FS methods that achieve the highest ``Rank-Sum'' values, using the C4.5 classification algorithm and under the FS+DS pipeline. For the IBA,  $G\text{-}Mean$ and F1 measures, ``Linear+SMOTE'' (in which the feature selection method is Linear, and the data re-sampling method is SMOTE) and ``RS+SMOTE'' are constantly among the top-3 combinations that yield the best $Rank\text{-}Sum$ values. Interestingly, this pattern still holds under the DS+FS pipeline using the same classifier, as depicted in Figure \ref{gf:2}. Therefore, they can be recommended to the practitioners to try with priority. In addition, we see that Linear+SPIDER is constantly the best performer in terms of $Acc$, under both the FS+DS and DS+FS pipelines. But in terms of TPR, the top-3 specific combinations under the two pipelines are totally different. We also notice that, the top-3 performers rank among the best performers only on half or less of the 52 datasets. Even if all the top-3  methods are tried, their $Group\text{-}Sum$ value indicates that, in union, they rank among the  top-3 in less than 75\% of the total datasets. 


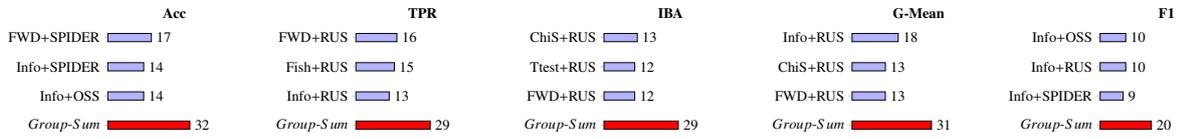
\begin{figure*}[h!]
\label{fig:sum_svm_fsds}
\resizebox{1.0\textwidth}{!}
{
\begin{tikzpicture}
\begin{scope}[local bounding box=gbox]
	\begin{axis}[
			title  = \textbf{Acc},
			xbar stacked,
			height = 40mm,
			width=0.3\textwidth,
			enlarge x limits  = 0.02,
			enlarge y limits={abs=0.2},
			bar width=2mm,
	    axis y line*=none,
	    axis x line*=none,
	    y axis line style = { opacity = 0 },
	    x axis line style = { opacity = 0 },
	    tickwidth = 0pt,
	    xtick=\empty,
	    ytick=data,
	    yticklabels={$Group\text{-}Sum$, Info+OSS, Info+SPIDER, FWD+SPIDER},
	    xmin=0,
	    xmax=52,
	    nodes near coords,
	    nodes near coords style={text=black, at ={(\pgfplotspointmeta,\pgfplotspointy)},anchor=west},
	    visualization depends on=y \as \pgfplotspointy,
	    every axis plot/.append style={fill},
	]
	\addplot[fill=red] coordinates {(32,0) (0,1) (0,2) (0,3)};
	\addplot[fill=blue!30!white] coordinates {(0,0) (14,1) (0,2) (0,3)};
	\addplot[fill=blue!30!white] coordinates {(0,0) (0,1) (14,2) (0,3)};
	\addplot[fill=blue!30!white] coordinates {(0,0) (0,1) (0,2) (17,3)};
\end{axis}  
\end{scope}

\begin{scope}[xshift=6cm,local bounding box=gbox]
	\begin{axis}[
			title  = \textbf{TPR},
			xbar stacked,
			height = 40mm,
			width=0.3\textwidth,
			enlarge x limits  = 0.02,
			enlarge y limits={abs=0.2},
			bar width=2mm,
	    axis y line*=none,
	    axis x line*=none,
	    y axis line style = { opacity = 0 },
	    x axis line style = { opacity = 0 },
	    tickwidth = 0pt,
	    xtick=\empty,
	    ytick=data,
	    yticklabels={$Group\text{-}Sum$, Info+RUS, Fish+RUS, FWD+RUS},
	    xmin=0,
	    xmax=52,
	    nodes near coords,
	    nodes near coords style={text=black, at ={(\pgfplotspointmeta,\pgfplotspointy)},anchor=west},
	    visualization depends on=y \as \pgfplotspointy,
	    every axis plot/.append style={fill},
	]
	\addplot[fill=red] coordinates {(29,0) (0,1) (0,2) (0,3)};
	\addplot[fill=blue!30!white] coordinates {(0,0) (13,1) (0,2) (0,3)};
	\addplot[fill=blue!30!white] coordinates {(0,0) (0,1) (15,2) (0,3)};
	\addplot[fill=blue!30!white] coordinates {(0,0) (0,1) (0,2) (16,3)};
\end{axis}  
\end{scope}

\begin{scope}[xshift=12cm,local bounding box=gbox]
	\begin{axis}[
			title  = \textbf{IBA},
			xbar stacked,
			height = 40mm,
			width=0.3\textwidth,
			enlarge x limits  = 0.02,
			enlarge y limits={abs=0.2},
			bar width=2mm,
	    axis y line*=none,
	    axis x line*=none,
	    y axis line style = { opacity = 0 },
	    x axis line style = { opacity = 0 },
	    tickwidth = 0pt,
	    xtick=\empty,
	    ytick=data,
	    yticklabels={$Group\text{-}Sum$, FWD+RUS, Ttest+RUS, ChiS+RUS},
	    xmin=0,
	    xmax=52,
	    nodes near coords,
	    nodes near coords style={text=black, at ={(\pgfplotspointmeta,\pgfplotspointy)},anchor=west},
	    visualization depends on=y \as \pgfplotspointy,
	    every axis plot/.append style={fill},
	]
	\addplot[fill=red] coordinates {(29,0) (0,1) (0,2) (0,3)};
	\addplot[fill=blue!30!white] coordinates {(0,0) (12,1) (0,2) (0,3)};
	\addplot[fill=blue!30!white] coordinates {(0,0) (0,1) (12,2) (0,3)};
	\addplot[fill=blue!30!white] coordinates {(0,0) (0,1) (0,2) (13,3)};
\end{axis}  
\end{scope}

\begin{scope}[xshift=18cm,local bounding box=gbox]
	\begin{axis}[
			title  = \textbf{G-Mean},
			xbar stacked,
			height = 40mm,
			width=0.3\textwidth,
			enlarge x limits  = 0.02,
			enlarge y limits={abs=0.2},
			bar width=2mm,
	    axis y line*=none,
	    axis x line*=none,
	    y axis line style = { opacity = 0 },
	    x axis line style = { opacity = 0 },
	    tickwidth = 0pt,
	    xtick=\empty,
	    ytick=data,
	    yticklabels={$Group\text{-}Sum$, FWD+RUS, ChiS+RUS, Info+RUS},
	    xmin=0,
	    xmax=52,
	    nodes near coords,
	    nodes near coords style={text=black, at ={(\pgfplotspointmeta,\pgfplotspointy)},anchor=west},
	    visualization depends on=y \as \pgfplotspointy,
	    every axis plot/.append style={fill},
	]
	\addplot[fill=red] coordinates {(31,0) (0,1) (0,2) (0,3)};
	\addplot[fill=blue!30!white] coordinates {(0,0) (13,1) (0,2) (0,3)};
	\addplot[fill=blue!30!white] coordinates {(0,0) (0,1) (13,2) (0,3)};
	\addplot[fill=blue!30!white] coordinates {(0,0) (0,1) (0,2) (18,3)};
\end{axis}  
\end{scope}

\begin{scope}[xshift=24cm,local bounding box=gbox]
	\begin{axis}[
			title  = \textbf{F1},
			xbar stacked,
			height = 40mm,
			width=0.3\textwidth,
			enlarge x limits  = 0.02,
			enlarge y limits={abs=0.2},
			bar width=2mm,
	    axis y line*=none,
	    axis x line*=none,
	    y axis line style = { opacity = 0 },
	    x axis line style = { opacity = 0 },
	    tickwidth = 0pt,
	    xtick=\empty,
	    ytick=data,
	    yticklabels={$Group\text{-}Sum$, Info+SPIDER, Info+RUS, Info+OSS},
	    xmin=0,
	    xmax=52,
	    nodes near coords,
	    nodes near coords style={text=black, at ={(\pgfplotspointmeta,\pgfplotspointy)},anchor=west},
	    visualization depends on=y \as \pgfplotspointy,
	    every axis plot/.append style={fill},
	]
	\addplot[fill=red] coordinates {(20,0) (0,1) (0,2) (0,3)};
	\addplot[fill=blue!30!white] coordinates {(0,0) (9,1) (0,2) (0,3)};
	\addplot[fill=blue!30!white] coordinates {(0,0) (0,1) (10,2) (0,3)};
	\addplot[fill=blue!30!white] coordinates {(0,0) (0,1) (0,2) (10,3)};
\end{axis}  
\end{scope}

\end{tikzpicture}
}
\caption{Rank-Sum results for the Top-3 specific combinations under the FS+DS pipeline, using the SVM classifier}
\label{gf:5}
\end{figure*}

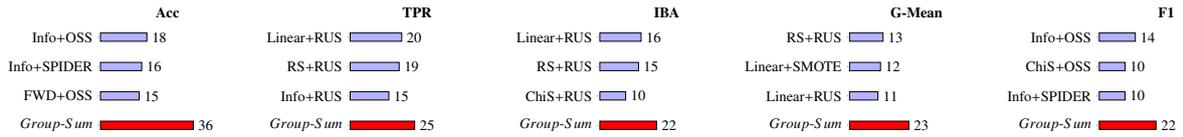
\begin{figure*}[h!]
\label{fig:sum_svm_dsfs}
\resizebox{1.0\textwidth}{!}
{
\begin{tikzpicture}
\begin{scope}[local bounding box=gbox]
	\begin{axis}[
			title  = \textbf{Acc},
			xbar stacked,
			height = 40mm,
			width=0.3\textwidth,
			enlarge x limits  = 0.02,
			enlarge y limits={abs=0.2},
			bar width=2mm,
	    axis y line*=none,
	    axis x line*=none,
	    y axis line style = { opacity = 0 },
	    x axis line style = { opacity = 0 },
	    tickwidth = 0pt,
	    xtick=\empty,
	    ytick=data,
	    yticklabels={$Group\text{-}Sum$, FWD+OSS, Info+SPIDER, Info+OSS},
	    xmin=0,
	    xmax=52,
	    nodes near coords,
	    nodes near coords style={text=black, at ={(\pgfplotspointmeta,\pgfplotspointy)},anchor=west},
	    visualization depends on=y \as \pgfplotspointy,
	    every axis plot/.append style={fill},
	]
	\addplot[fill=red] coordinates {(36,0) (0,1) (0,2) (0,3)};
	\addplot[fill=blue!30!white] coordinates {(0,0) (15,1) (0,2) (0,3)};
	\addplot[fill=blue!30!white] coordinates {(0,0) (0,1) (16,2) (0,3)};
	\addplot[fill=blue!30!white] coordinates {(0,0) (0,1) (0,2) (18,3)};
\end{axis}  
\end{scope}

\begin{scope}[xshift=6cm,local bounding box=gbox]
	\begin{axis}[
			title  = \textbf{TPR},
			xbar stacked,
			height = 40mm,
			width=0.3\textwidth,
			enlarge x limits  = 0.02,
			enlarge y limits={abs=0.2},
			bar width=2mm,
	    axis y line*=none,
	    axis x line*=none,
	    y axis line style = { opacity = 0 },
	    x axis line style = { opacity = 0 },
	    tickwidth = 0pt,
	    xtick=\empty,
	    ytick=data,
	    yticklabels={$Group\text{-}Sum$, Info+RUS, RS+RUS, Linear+RUS},
	    xmin=0,
	    xmax=52,
	    nodes near coords,
	    nodes near coords style={text=black, at ={(\pgfplotspointmeta,\pgfplotspointy)},anchor=west},
	    visualization depends on=y \as \pgfplotspointy,
	    every axis plot/.append style={fill},
	]
	\addplot[fill=red] coordinates {(25,0) (0,1) (0,2) (0,3)};
	\addplot[fill=blue!30!white] coordinates {(0,0) (15,1) (0,2) (0,3)};
	\addplot[fill=blue!30!white] coordinates {(0,0) (0,1) (19,2) (0,3)};
	\addplot[fill=blue!30!white] coordinates {(0,0) (0,1) (0,2) (20,3)};
\end{axis}  
\end{scope}

\begin{scope}[xshift=12cm,local bounding box=gbox]
	\begin{axis}[
			title  = \textbf{IBA},
			xbar stacked,
			height = 40mm,
			width=0.3\textwidth,
			enlarge x limits  = 0.02,
			enlarge y limits={abs=0.2},
			bar width=2mm,
	    axis y line*=none,
	    axis x line*=none,
	    y axis line style = { opacity = 0 },
	    x axis line style = { opacity = 0 },
	    tickwidth = 0pt,
	    xtick=\empty,
	    ytick=data,
	    yticklabels={$Group\text{-}Sum$, ChiS+RUS, RS+RUS, Linear+RUS},
	    xmin=0,
	    xmax=52,
	    nodes near coords,
	    nodes near coords style={text=black, at ={(\pgfplotspointmeta,\pgfplotspointy)},anchor=west},
	    visualization depends on=y \as \pgfplotspointy,
	    every axis plot/.append style={fill},
	]
	\addplot[fill=red] coordinates {(22,0) (0,1) (0,2) (0,3)};
	\addplot[fill=blue!30!white] coordinates {(0,0) (10,1) (0,2) (0,3)};
	\addplot[fill=blue!30!white] coordinates {(0,0) (0,1) (15,2) (0,3)};
	\addplot[fill=blue!30!white] coordinates {(0,0) (0,1) (0,2) (16,3)};
\end{axis}  
\end{scope}

\begin{scope}[xshift=18cm,local bounding box=gbox]
	\begin{axis}[
			title  = \textbf{G-Mean},
			xbar stacked,
			height = 40mm,
			width=0.3\textwidth,
			enlarge x limits  = 0.02,
			enlarge y limits={abs=0.2},
			bar width=2mm,
	    axis y line*=none,
	    axis x line*=none,
	    y axis line style = { opacity = 0 },
	    x axis line style = { opacity = 0 },
	    tickwidth = 0pt,
	    xtick=\empty,
	    ytick=data,
	    yticklabels={$Group\text{-}Sum$, Linear+RUS, Linear+SMOTE, RS+RUS},
	    xmin=0,
	    xmax=52,
	    nodes near coords,
	    nodes near coords style={text=black, at ={(\pgfplotspointmeta,\pgfplotspointy)},anchor=west},
	    visualization depends on=y \as \pgfplotspointy,
	    every axis plot/.append style={fill},
	]
	\addplot[fill=red] coordinates {(23,0) (0,1) (0,2) (0,3)};
	\addplot[fill=blue!30!white] coordinates {(0,0) (11,1) (0,2) (0,3)};
	\addplot[fill=blue!30!white] coordinates {(0,0) (0,1) (12,2) (0,3)};
	\addplot[fill=blue!30!white] coordinates {(0,0) (0,1) (0,2) (13,3)};
\end{axis}  
\end{scope}

\begin{scope}[xshift=24cm,local bounding box=gbox]
	\begin{axis}[
			title  = \textbf{F1},
			xbar stacked,
			height = 40mm,
			width=0.3\textwidth,
			enlarge x limits  = 0.02,
			enlarge y limits={abs=0.2},
			bar width=2mm,
	    axis y line*=none,
	    axis x line*=none,
	    y axis line style = { opacity = 0 },
	    x axis line style = { opacity = 0 },
	    tickwidth = 0pt,
	    xtick=\empty,
	    ytick=data,
	    yticklabels={$Group\text{-}Sum$, Info+SPIDER, ChiS+OSS, Info+OSS},
	    xmin=0,
	    xmax=52,
	    nodes near coords,
	    nodes near coords style={text=black, at ={(\pgfplotspointmeta,\pgfplotspointy)},anchor=west},
	    visualization depends on=y \as \pgfplotspointy,
	    every axis plot/.append style={fill},
	]
	\addplot[fill=red] coordinates {(22,0) (0,1) (0,2) (0,3)};
	\addplot[fill=blue!30!white] coordinates {(0,0) (10,1) (0,2) (0,3)};
	\addplot[fill=blue!30!white] coordinates {(0,0) (0,1) (10,2) (0,3)};
	\addplot[fill=blue!30!white] coordinates {(0,0) (0,1) (0,2) (14,3)};
\end{axis}  
\end{scope}

\end{tikzpicture}
}
\caption{Rank-Sum results for the Top-3 specific combinations under the DS+FS pipeline, using the SVM classifier}
\label{gf:6}
\end{figure*}

\textbf{SVM}. Shown in Figures \ref{gf:5} and \ref{gf:6}, we see that,  with SVM and under the $G\text{-}Mean$ and $F1$ metrics, the relative advantages of the top-3 combinations are not very outstanding. Nevertheless, we can still find that RUS seems to be the best data re-sampling method in the two pipelines under all of  the IBA, $G\text{-}Mean$ and $TPR$ metrics. Overall, it depends on the specific pipeline and evaluation metrics to select the corresponding top performing combinations.

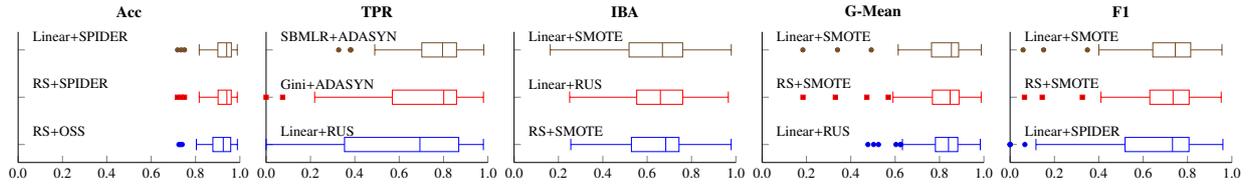
\begin{figure*}[htb]
\label{fig:boxplot_c45_fsds}
\resizebox{1.0\textwidth}{!}
{
\centering
\begin{tikzpicture}
\pgfsetplotmarksize{0.3ex}
\begin{scope}[xshift=0cm, yshift=-0cm, local bounding box=gbox]
\begin{axis}
[
	title  = \textbf{Acc},
	width=6.5cm,
	height=\gbh,
	box plot width=1.5em,
	xmin=0.0,
	xmax=1.0,
	boxplot/draw direction=x,
	tick label style=
	{
		/pgf/number format/fixed,
	  /pgf/number format/fixed zerofill,
	  /pgf/number format/precision=1,
	  font=\small
	},
	axis x line* = bottom,
	axis y line* = left,
	ytick={1,2,3},
	xtick={0,0.2,...,1.0},
	y tick label style={anchor=west,xshift= 2mm, yshift=3mm},
	yticklabels={RS+OSS, RS+SPIDER, Linear+SPIDER},
	legend style={at={(0.5,-0.15)}, anchor=north,legend columns=2},
]

\addplot+[boxplot prepared={box extend=0.3, lower whisker=0.8037, lower quartile=0.8781, median=0.9251, upper quartile=0.9587, upper whisker=0.9887}, ] coordinates{(0,0.7396) (0,0.7320) (0,0.7250) };
\addplot+[boxplot prepared={box extend=0.3, lower whisker=0.8166, lower quartile=0.9017, median=0.9412, upper quartile=0.9612, upper whisker=0.9883}, ] coordinates{(0,0.7500) (0,0.7353) (0,0.7180) };
\addplot+[boxplot prepared={box extend=0.3, lower whisker=0.8166, lower quartile=0.9006, median=0.9404, upper quartile=0.9616, upper whisker=0.9883}, ] coordinates{(0,0.7500) (0,0.7353) (0,0.7190) };
\end{axis}
\end{scope}

\begin{scope}[xshift=5.5cm, yshift=-0cm, local bounding box=gbox]
\begin{axis}
[
	title  = \textbf{TPR},
	width=6.5cm,
	height=\gbh,
	box plot width=1.5em,
	xmin=0.0,
	xmax=1.0,
	boxplot/draw direction=x,
	tick label style=
	{
		/pgf/number format/fixed,
	  /pgf/number format/fixed zerofill,
	  /pgf/number format/precision=1,
	  font=\small
	},
	axis x line* = bottom,
	axis y line* = left,
	ytick={1,2,3},
	xtick={0,0.2,...,1.0},
	y tick label style={anchor=west,xshift= 2mm, yshift=3mm},
	yticklabels={Linear+RUS, Gini+ADASYN, SBMLR+ADASYN},
	legend style={at={(0.5,-0.15)}, anchor=north,legend columns=2},
]

\addplot+[boxplot prepared={box extend=0.3, lower whisker=0, lower quartile=0.3536, median=0.6935, upper quartile=0.8687, upper whisker=0.9797}, ] coordinates{};
\addplot+[boxplot prepared={box extend=0.3, lower whisker=0.219, lower quartile=0.5694, median=0.801, upper quartile=0.8586, upper whisker=0.98}, ] coordinates{(0,0.0745) (0,0.0000) };
\addplot+[boxplot prepared={box extend=0.3, lower whisker=0.4898, lower quartile=0.7019, median=0.7955, upper quartile=0.8585, upper whisker=0.9818}, ] coordinates{(0,0.3821) (0,0.3800) (0,0.3273) };
\end{axis}
\end{scope}

\begin{scope}[xshift=11cm, yshift=-0cm, local bounding box=gbox]
\begin{axis}
[
	title  = \textbf{IBA},
	width=6.5cm,
	height=\gbh,
	box plot width=1.5em,
	xmin=0.0,
	xmax=1.0,
	boxplot/draw direction=x,
	tick label style=
	{
		/pgf/number format/fixed,
	  /pgf/number format/fixed zerofill,
	  /pgf/number format/precision=1,
	  font=\small
	},
	axis x line* = bottom,
	axis y line* = left,
	ytick={1,2,3},
	xtick={0,0.2,...,1.0},
	y tick label style={anchor=west,xshift= 2mm, yshift=3mm},
	yticklabels={RS+SMOTE, Linear+RUS, Linear+SMOTE},
	legend style={at={(0.5,-0.15)}, anchor=north,legend columns=2},
]

\addplot+[boxplot prepared={box extend=0.3, lower whisker=0.2556, lower quartile=0.5291, median=0.6838, upper quartile=0.7437, upper whisker=0.9788}, ] coordinates{};
\addplot+[boxplot prepared={box extend=0.3, lower whisker=0.2505, lower quartile=0.5529, median=0.6596, upper quartile=0.7602, upper whisker=0.9659}, ] coordinates{};
\addplot+[boxplot prepared={box extend=0.3, lower whisker=0.1631, lower quartile=0.5184, median=0.6688, upper quartile=0.76, upper whisker=0.9793}, ] coordinates{};
\end{axis}
\end{scope}

\begin{scope}[xshift=16.5cm, yshift=0cm, local bounding box=gbox]
\begin{axis}
[
	title  = \textbf{G-Mean},
	width=6.5cm,
	height=\gbh,
	box plot width=1.5em,
	xmin=0.0,
	xmax=1.0,
	boxplot/draw direction=x,
	tick label style=
	{
		/pgf/number format/fixed,
	  /pgf/number format/fixed zerofill,
	  /pgf/number format/precision=1,
	  font=\small
	},
	axis x line* = bottom,
	axis y line* = left,
	ytick={1,2,3},
	xtick={0,0.2,...,1.0},
	y tick label style={anchor=west,xshift= 2mm, yshift=3mm},
	yticklabels={Linear+RUS, RS+SMOTE, Linear+SMOTE},
	legend style={at={(0.5,-0.15)}, anchor=north,legend columns=2},
]

\addplot+[boxplot prepared={box extend=0.3, lower whisker=0.6327, lower quartile=0.7815, median=0.8405, upper quartile=0.8824, upper whisker=0.9838}, ] coordinates{(0,0.6249) (0,0.6245) (0,0.6039) (0,0.5248) (0,0.5028) (0,0.4772) };
\addplot+[boxplot prepared={box extend=0.3, lower whisker=0.5902, lower quartile=0.768, median=0.8478, upper quartile=0.888, upper whisker=0.9878}, ] coordinates{(0,0.5690) (0,0.4724) (0,0.3304) (0,0.1843) };
\addplot+[boxplot prepared={box extend=0.3, lower whisker=0.6136, lower quartile=0.7639, median=0.8537, upper quartile=0.8862, upper whisker=0.9883}, ] coordinates{(0,0.4923) (0,0.3401) (0,0.1838) };
\end{axis}
\end{scope}

\begin{scope}[xshift=22cm, yshift=0cm, local bounding box=gbox]
\begin{axis}
[
	title  = \textbf{F1},
	width=6.5cm,
	height=\gbh,
	box plot width=1.5em,
	xmin=0.0,
	xmax=1.0,
	boxplot/draw direction=x,
	tick label style=
	{
		/pgf/number format/fixed,
	  /pgf/number format/fixed zerofill,
	  /pgf/number format/precision=1,
	  font=\small
	},
	axis x line* = bottom,
	axis y line* = left,
	ytick={1,2,3},
	xtick={0,0.2,...,1.0},
	y tick label style={anchor=west,xshift= 2mm, yshift=3mm},
	yticklabels={Linear+SPIDER, RS+SMOTE, Linear+SMOTE},
	legend style={at={(0.5,-0.15)}, anchor=north,legend columns=2},
]

\addplot+[boxplot prepared={box extend=0.3, lower whisker=0.1167, lower quartile=0.5185, median=0.7324, upper quartile=0.8067, upper whisker=0.9596}, ] coordinates{(0,0.0667) (0,0.0000) (0,0.0000) (0,0.0000) };
\addplot+[boxplot prepared={box extend=0.3, lower whisker=0.4099, lower quartile=0.63, median=0.7348, upper quartile=0.8073, upper whisker=0.952}, ] coordinates{(0,0.3259) (0,0.1449) (0,0.0649) };
\addplot+[boxplot prepared={box extend=0.3, lower whisker=0.4003, lower quartile=0.6435, median=0.7448, upper quartile=0.8144, upper whisker=0.9554}, ] coordinates{(0,0.3487) (0,0.1511) (0,0.0587) };
\end{axis}
\end{scope}

\end{tikzpicture}
}
\gvsa
\caption{Box plots for the cross-datasets averaged value of each metric, for the Top-3 specific combinations using C4.5 and  FS+DS pipeline}
\label{gf:7}
\end{figure*}

\begin{figure*}[h!]
\label{fig:boxplot_c45_dsfs}
\resizebox{1.0\textwidth}{!}
{
\centering
\begin{tikzpicture}
\pgfsetplotmarksize{0.3ex}
\begin{scope}[xshift=0cm, yshift=-0cm, local bounding box=gbox]
\begin{axis}
[
	title  = \textbf{Acc},
	width=6.5cm,
	height=\gbh,
	box plot width=1.5em,
	xmin=0.0,
	xmax=1.0,
	boxplot/draw direction=x,
	tick label style=
	{
		/pgf/number format/fixed,
	  /pgf/number format/fixed zerofill,
	  /pgf/number format/precision=1,
	  font=\small
	},
	axis x line* = bottom,
	axis y line* = left,
	ytick={1,2,3},
	xtick={0,0.2,...,1.0},
	y tick label style={anchor=west,xshift= 2mm, yshift=3mm},
	yticklabels={RS+OSS, Linear+OSS, Linear+SPIDER},
	legend style={at={(0.5,-0.15)}, anchor=north,legend columns=2},
]

\addplot+[boxplot prepared={box extend=0.3, lower whisker=0.8365, lower quartile=0.8972, median=0.9277, upper quartile=0.9468, upper whisker=0.9892}, ] coordinates{(0,0.8140) (0,0.7318) (0,0.7310) (0,0.7288) };
\addplot+[boxplot prepared={box extend=0.3, lower whisker=0.814, lower quartile=0.8924, median=0.9293, upper quartile=0.9516, upper whisker=0.9892}, ] coordinates{(0,0.7380) (0,0.7331) (0,0.7288) };
\addplot+[boxplot prepared={box extend=0.3, lower whisker=0.8168, lower quartile=0.9016, median=0.9398, upper quartile=0.9597, upper whisker=0.9874}, ] coordinates{(0,0.8132) (0,0.7565) (0,0.7320) (0,0.7300) };
\end{axis}
\end{scope}

\begin{scope}[xshift=5.5cm, yshift=-0cm, local bounding box=gbox]
\begin{axis}
[
	title  = \textbf{TPR},
	width=6.5cm,
	height=\gbh,
	box plot width=1.5em,
	xmin=0.0,
	xmax=1.0,
	boxplot/draw direction=x,
	tick label style=
	{
		/pgf/number format/fixed,
	  /pgf/number format/fixed zerofill,
	  /pgf/number format/precision=1,
	  font=\small
	},
	axis x line* = bottom,
	axis y line* = left,
	ytick={1,2,3},
	xtick={0,0.2,...,1.0},
	y tick label style={anchor=west,xshift= 2mm, yshift=3mm},
	yticklabels={RS+SMOTE, RS+RUS, Info+RUS},
	legend style={at={(0.5,-0.15)}, anchor=north,legend columns=2},
]

\addplot+[boxplot prepared={box extend=0.3, lower whisker=0.4533, lower quartile=0.6802, median=0.7954, upper quartile=0.8822, upper whisker=0.9878}, ] coordinates{(0,0.3400) (0,0.1691) (0,0.1000) };
\addplot+[boxplot prepared={box extend=0.3, lower whisker=0.475, lower quartile=0.7154, median=0.8231, upper quartile=0.8814, upper whisker=1}, ] coordinates{(0,0.4656) (0,0.4571) (0,0.4066) (0,0.3600) (0,0.3321) (0,0.2200) };
\addplot+[boxplot prepared={box extend=0.3, lower whisker=0.22, lower quartile=0.6054, median=0.7941, upper quartile=0.8868, upper whisker=1}, ] coordinates{};
\end{axis}
\end{scope}

\begin{scope}[xshift=11cm, yshift=-0cm, local bounding box=gbox]
\begin{axis}
[
	title  = \textbf{IBA},
	width=6.5cm,
	height=\gbh,
	box plot width=1.5em,
	xmin=0.0,
	xmax=1.0,
	boxplot/draw direction=x,
	tick label style=
	{
		/pgf/number format/fixed,
	  /pgf/number format/fixed zerofill,
	  /pgf/number format/precision=1,
	  font=\small
	},
	axis x line* = bottom,
	axis y line* = left,
	ytick={1,2,3},
	xtick={0,0.2,...,1.0},
	y tick label style={anchor=west,xshift= 2mm, yshift=3mm},
	yticklabels={RS+RUS, Linear+SMOTE, RS+SMOTE},
	legend style={at={(0.5,-0.15)}, anchor=north,legend columns=2},
]

\addplot+[boxplot prepared={box extend=0.3, lower whisker=0.2606, lower quartile=0.5788, median=0.6968, upper quartile=0.793, upper whisker=0.974}, ] coordinates{};
\addplot+[boxplot prepared={box extend=0.3, lower whisker=0.2238, lower quartile=0.5487, median=0.6923, upper quartile=0.7844, upper whisker=0.9771}, ] coordinates{};
\addplot+[boxplot prepared={box extend=0.3, lower whisker=0.2233, lower quartile=0.5529, median=0.6955, upper quartile=0.7844, upper whisker=0.9771}, ] coordinates{};
\end{axis}
\end{scope}

\begin{scope}[xshift=16.5cm, yshift=0cm, local bounding box=gbox]
\begin{axis}
[
	title  = \textbf{G-Mean},
	width=6.5cm,
	height=\gbh,
	box plot width=1.5em,
	xmin=0.0,
	xmax=1.0,
	boxplot/draw direction=x,
	tick label style=
	{
		/pgf/number format/fixed,
	  /pgf/number format/fixed zerofill,
	  /pgf/number format/precision=1,
	  font=\small
	},
	axis x line* = bottom,
	axis y line* = left,
	ytick={1,2,3},
	xtick={0,0.2,...,1.0},
	y tick label style={anchor=west,xshift= 2mm, yshift=3mm},
	yticklabels={Linear+RUS, Linear+SMOTE, RS+SMOTE},
	legend style={at={(0.5,-0.15)}, anchor=north,legend columns=2},
]

\addplot+[boxplot prepared={box extend=0.3, lower whisker=0.6348, lower quartile=0.7872, median=0.8524, upper quartile=0.8986, upper whisker=0.9848}, ] coordinates{(0,0.6078) (0,0.5603) (0,0.5390) (0,0.5370) (0,0.4756) (0,0.4712) (0,0.4595) (0,0.4134) };
\addplot+[boxplot prepared={box extend=0.3, lower whisker=0.6067, lower quartile=0.774, median=0.8513, upper quartile=0.9033, upper whisker=0.9891}, ] coordinates{(0,0.5690) (0,0.5285) (0,0.4212) (0,0.2944) };
\addplot+[boxplot prepared={box extend=0.3, lower whisker=0.6088, lower quartile=0.7733, median=0.8538, upper quartile=0.9041, upper whisker=0.9891}, ] coordinates{(0,0.5275) (0,0.3976) (0,0.3102) };
\end{axis}
\end{scope}

\begin{scope}[xshift=22cm, yshift=0cm, local bounding box=gbox]
\begin{axis}
[
	title  = \textbf{F1},
	width=6.5cm,
	height=\gbh,
	box plot width=1.5em,
	xmin=0.0,
	xmax=1.0,
	boxplot/draw direction=x,
	tick label style=
	{
		/pgf/number format/fixed,
	  /pgf/number format/fixed zerofill,
	  /pgf/number format/precision=1,
	  font=\small
	},
	axis x line* = bottom,
	axis y line* = left,
	ytick={1,2,3},
	xtick={0,0.2,...,1.0},
	y tick label style={anchor=west,xshift= 2mm, yshift=3mm},
	yticklabels={Linear+RUS, RS+SMOTE, Linear+SMOTE},
	legend style={at={(0.5,-0.15)}, anchor=north,legend columns=2},
]

\addplot+[boxplot prepared={box extend=0.3, lower whisker=0.2554, lower quartile=0.5645, median=0.708, upper quartile=0.7883, upper whisker=0.9558}, ] coordinates{(0,0.2177) (0,0.1742) };
\addplot+[boxplot prepared={box extend=0.3, lower whisker=0.4467, lower quartile=0.6322, median=0.7487, upper quartile=0.8138, upper whisker=0.9659}, ] coordinates{(0,0.3150) (0,0.1806) (0,0.1349) };
\addplot+[boxplot prepared={box extend=0.3, lower whisker=0.4133, lower quartile=0.6358, median=0.7498, upper quartile=0.8155, upper whisker=0.9659}, ] coordinates{(0,0.3228) (0,0.1846) (0,0.1263) };
\end{axis}
\end{scope}

\end{tikzpicture}
}
\gvsa
\caption{Box plots for the cross-datasets averaged value of each metric, for the Top-3 specific combinations using C4.5 and  DS+FS pipeline}
\label{gf:8}
\end{figure*}
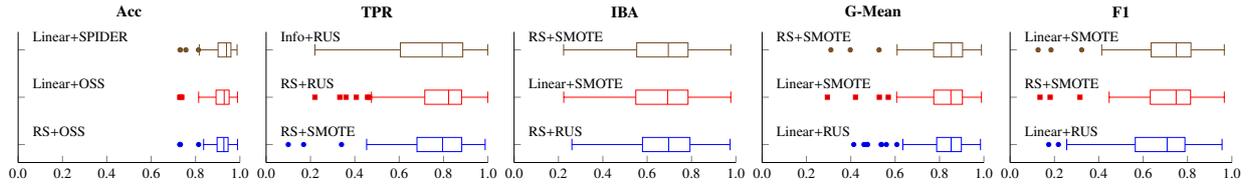

\begin{figure*}[h!]
\label{fig:boxplot_svm_fsds}
\resizebox{1.0\textwidth}{!}
{
\centering
\begin{tikzpicture}
\pgfsetplotmarksize{0.3ex}
\begin{scope}[xshift=0cm, yshift=-0cm, local bounding box=gbox]
\begin{axis}
[
	title  = \textbf{Acc},
	width=6.5cm,
	height=\gbh,
	box plot width=1.5em,
	xmin=0.0,
	xmax=1.0,
	boxplot/draw direction=x,
	tick label style=
	{
		/pgf/number format/fixed,
	  /pgf/number format/fixed zerofill,
	  /pgf/number format/precision=1,
	  font=\small
	},
	axis x line* = bottom,
	axis y line* = left,
	ytick={1,2,3},
	xtick={0,0.2,...,1.0},
	y tick label style={anchor=west,xshift= 2mm, yshift=3mm},
	yticklabels={Info+OSS, Info+SPIDER, FWD+SPIDER},
	legend style={at={(0.5,-0.15)}, anchor=north,legend columns=2},
]

\addplot+[boxplot prepared={box extend=0.3, lower whisker=0.8682, lower quartile=0.9157, median=0.9415, upper quartile=0.9649, upper whisker=0.9953}, ] coordinates{(0,0.7849) (0,0.7380) (0,0.7188) (0,0.7057) (0,0.6543) };
\addplot+[boxplot prepared={box extend=0.3, lower whisker=0.8691, lower quartile=0.9133, median=0.9481, upper quartile=0.9657, upper whisker=0.9957}, ] coordinates{(0,0.7710) (0,0.7350) (0,0.7321) (0,0.7222) (0,0.7057) };
\addplot+[boxplot prepared={box extend=0.3, lower whisker=0.7863, lower quartile=0.893, median=0.9453, upper quartile=0.9659, upper whisker=0.9909}, ] coordinates{(0,0.7761) (0,0.7450) (0,0.7287) (0,0.7283) (0,0.7200) };
\end{axis}
\end{scope}

\begin{scope}[xshift=5.5cm, yshift=-0cm, local bounding box=gbox]
\begin{axis}
[
	title  = \textbf{TPR},
	width=6.5cm,
	height=\gbh,
	box plot width=1.5em,
	xmin=0.0,
	xmax=1.0,
	boxplot/draw direction=x,
	tick label style=
	{
		/pgf/number format/fixed,
	  /pgf/number format/fixed zerofill,
	  /pgf/number format/precision=1,
	  font=\small
	},
	axis x line* = bottom,
	axis y line* = left,
	ytick={1,2,3},
	xtick={0,0.2,...,1.0},
	y tick label style={anchor=west,xshift= 2mm, yshift=3mm},
	yticklabels={Info+RUS, Fish+RUS, FWD+RUS},
	legend style={at={(0.5,-0.15)}, anchor=north,legend columns=2},
]

\addplot+[boxplot prepared={box extend=0.3, lower whisker=0.5298, lower quartile=0.7075, median=0.8277, upper quartile=0.9, upper whisker=1}, ] coordinates{(0,0.3941) (0,0.2236) (0,0.2200) };
\addplot+[boxplot prepared={box extend=0.3, lower whisker=0.5298, lower quartile=0.6945, median=0.8312, upper quartile=0.8942, upper whisker=0.9848}, ] coordinates{(0,0.3713) (0,0.2873) (0,0.2511) (0,0.2200) };
\addplot+[boxplot prepared={box extend=0.3, lower whisker=0.3125, lower quartile=0.6458, median=0.819, upper quartile=0.8989, upper whisker=1}, ] coordinates{(0,0.2393) (0,0.2150) (0,0.1600) };
\end{axis}
\end{scope}

\begin{scope}[xshift=11cm, yshift=-0cm, local bounding box=gbox]
\begin{axis}
[
	title  = \textbf{IBA},
	width=6.5cm,
	height=\gbh,
	box plot width=1.5em,
	xmin=0.0,
	xmax=1.0,
	boxplot/draw direction=x,
	tick label style=
	{
		/pgf/number format/fixed,
	  /pgf/number format/fixed zerofill,
	  /pgf/number format/precision=1,
	  font=\small
	},
	axis x line* = bottom,
	axis y line* = left,
	ytick={1,2,3},
	xtick={0,0.2,...,1.0},
	y tick label style={anchor=west,xshift= 2mm, yshift=3mm},
	yticklabels={FWD+RUS, Ttest+RUS, ChiS+RUS},
	legend style={at={(0.5,-0.15)}, anchor=north,legend columns=2},
]

\addplot+[boxplot prepared={box extend=0.3, lower whisker=0.0927, lower quartile=0.4897, median=0.7172, upper quartile=0.8101, upper whisker=0.9727}, ] coordinates{};
\addplot+[boxplot prepared={box extend=0.3, lower whisker=0.377, lower quartile=0.5641, median=0.6788, upper quartile=0.7885, upper whisker=0.9798}, ] coordinates{};
\addplot+[boxplot prepared={box extend=0.3, lower whisker=0.0927, lower quartile=0.4897, median=0.7172, upper quartile=0.8101, upper whisker=0.9727}, ] coordinates{};
\end{axis}
\end{scope}

\begin{scope}[xshift=16.5cm, yshift=0cm, local bounding box=gbox]
\begin{axis}
[
	title  = \textbf{G-Mean},
	width=6.5cm,
	height=\gbh,
	box plot width=1.5em,
	xmin=0.0,
	xmax=1.0,
	boxplot/draw direction=x,
	tick label style=
	{
		/pgf/number format/fixed,
	  /pgf/number format/fixed zerofill,
	  /pgf/number format/precision=1,
	  font=\small
	},
	axis x line* = bottom,
	axis y line* = left,
	ytick={1,2,3},
	xtick={0,0.2,...,1.0},
	y tick label style={anchor=west,xshift= 2mm, yshift=3mm},
	yticklabels={FWD+RUS, ChiS+RUS, Info+RUS},
	legend style={at={(0.5,-0.15)}, anchor=north,legend columns=2},
]

\addplot+[boxplot prepared={box extend=0.3, lower whisker=0.5486, lower quartile=0.7121, median=0.8664, upper quartile=0.9077, upper whisker=0.9773}, ] coordinates{(0,0.3717) (0,0.3607) (0,0.3506) (0,0.3283) (0,0.2655) };
\addplot+[boxplot prepared={box extend=0.3, lower whisker=0.6064, lower quartile=0.7741, median=0.8794, upper quartile=0.9037, upper whisker=0.9944}, ] coordinates{(0,0.5292) (0,0.5170) (0,0.4595) };
\addplot+[boxplot prepared={box extend=0.3, lower whisker=0.6304, lower quartile=0.7865, median=0.8695, upper quartile=0.9094, upper whisker=0.9888}, ] coordinates{(0,0.5673) (0,0.4595) (0,0.4009) };
\end{axis}
\end{scope}

\begin{scope}[xshift=22cm, yshift=0cm, local bounding box=gbox]
\begin{axis}
[
	title  = \textbf{F1},
	width=6.5cm,
	height=\gbh,
	box plot width=1.5em,
	xmin=0.0,
	xmax=1.0,
	boxplot/draw direction=x,
	tick label style=
	{
		/pgf/number format/fixed,
	  /pgf/number format/fixed zerofill,
	  /pgf/number format/precision=1,
	  font=\small
	},
	axis x line* = bottom,
	axis y line* = left,
	ytick={1,2,3},
	xtick={0,0.2,...,1.0},
	y tick label style={anchor=west,xshift= 2mm, yshift=3mm},
	yticklabels={Info+SPIDER, Info+RUS, Info+OSS},
	legend style={at={(0.5,-0.15)}, anchor=north,legend columns=2},
]

\addplot+[boxplot prepared={box extend=0.3, lower whisker=0.265, lower quartile=0.589, median=0.7657, upper quartile=0.8266, upper whisker=0.9849}, ] coordinates{(0,0.0000) (0,0.0000) };
\addplot+[boxplot prepared={box extend=0.3, lower whisker=0.336, lower quartile=0.58, median=0.7667, upper quartile=0.8295, upper whisker=0.9846}, ] coordinates{(0,0.0000) (0,0.0000) };
\addplot+[boxplot prepared={box extend=0.3, lower whisker=0.336, lower quartile=0.5965, median=0.7358, upper quartile=0.8342, upper whisker=0.9573}, ] coordinates{(0,0.2022) (0,0.1393) };
\end{axis}
\end{scope}

\end{tikzpicture}
}
\gvsa
\caption{Box plots for the cross-datasets averaged value of each metric, for the Top-3 specific combinations using SVM and  FS+DS pipeline}
\label{gf:11}
\end{figure*}
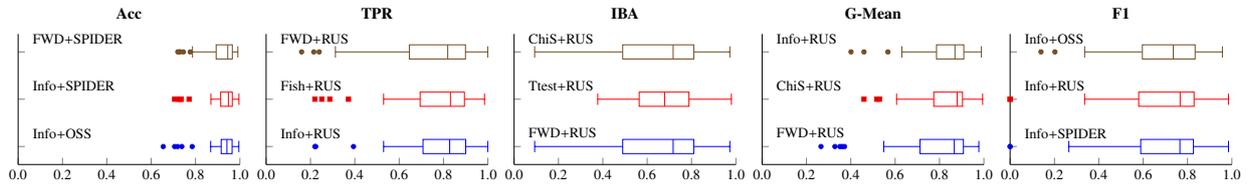

\begin{figure*}[h!]
\label{fig:boxplot_svm_dsfs}
\resizebox{1.0\textwidth}{!}
{
\centering
\begin{tikzpicture}
\pgfsetplotmarksize{0.3ex}
\begin{scope}[xshift=0cm, yshift=-0cm, local bounding box=gbox]
\begin{axis}
[
	title  = \textbf{Acc},
	width=6.5cm,
	height=\gbh,
	box plot width=1.5em,
	xmin=0.0,
	xmax=1.0,
	boxplot/draw direction=x,
	tick label style=
	{
		/pgf/number format/fixed,
	  /pgf/number format/fixed zerofill,
	  /pgf/number format/precision=1,
	  font=\small
	},
	axis x line* = bottom,
	axis y line* = left,
	ytick={1,2,3},
	xtick={0,0.2,...,1.0},
	y tick label style={anchor=west,xshift= 2mm, yshift=3mm},
	yticklabels={FWD+OSS, Info+SPIDER, Info+OSS},
	legend style={at={(0.5,-0.15)}, anchor=north,legend columns=2},
]

\addplot+[boxplot prepared={box extend=0.3, lower whisker=0.8513, lower quartile=0.9, median=0.9287, upper quartile=0.9591, upper whisker=0.99}, ] coordinates{(0,0.7993) (0,0.7320) (0,0.7252) (0,0.7187) (0,0.5577) };
\addplot+[boxplot prepared={box extend=0.3, lower whisker=0.863, lower quartile=0.9082, median=0.947, upper quartile=0.9644, upper whisker=1}, ] coordinates{(0,0.8184) (0,0.7784) (0,0.7759) (0,0.7460) (0,0.7287) (0,0.6953) };
\addplot+[boxplot prepared={box extend=0.3, lower whisker=0.87, lower quartile=0.9091, median=0.9353, upper quartile=0.9646, upper whisker=0.9935}, ] coordinates{(0,0.7947) (0,0.7784) (0,0.7520) (0,0.7220) (0,0.7057) };
\end{axis}
\end{scope}

\begin{scope}[xshift=5.5cm, yshift=-0cm, local bounding box=gbox]
\begin{axis}
[
	title  = \textbf{TPR},
	width=6.5cm,
	height=\gbh,
	box plot width=1.5em,
	xmin=0.0,
	xmax=1.0,
	boxplot/draw direction=x,
	tick label style=
	{
		/pgf/number format/fixed,
	  /pgf/number format/fixed zerofill,
	  /pgf/number format/precision=1,
	  font=\small
	},
	axis x line* = bottom,
	axis y line* = left,
	ytick={1,2,3},
	xtick={0,0.2,...,1.0},
	y tick label style={anchor=west,xshift= 2mm, yshift=3mm},
	yticklabels={Info+RUS, RS+RUS, Linear+RUS},
	legend style={at={(0.5,-0.15)}, anchor=north,legend columns=2},
]

\addplot+[boxplot prepared={box extend=0.3, lower whisker=0.3133, lower quartile=0.6383, median=0.7969, upper quartile=0.8629, upper whisker=0.9818}, ] coordinates{(0,0.2695) (0,0.2400) };
\addplot+[boxplot prepared={box extend=0.3, lower whisker=0.5674, lower quartile=0.7308, median=0.8336, upper quartile=0.9004, upper whisker=0.9958}, ] coordinates{(0,0.4607) (0,0.4518) (0,0.4500) (0,0.4321) (0,0.4000) (0,0.3964) (0,0.3316) (0,0.2200) };
\addplot+[boxplot prepared={box extend=0.3, lower whisker=0.866, lower quartile=0.9331, median=0.9887, upper quartile=1, upper whisker=1}, ] coordinates{(0,0.8062) (0,0.7880) (0,0.7536) (0,0.6211) };
\end{axis}
\end{scope}

\begin{scope}[xshift=11cm, yshift=-0cm, local bounding box=gbox]
\begin{axis}
[
	title  = \textbf{IBA},
	width=6.5cm,
	height=\gbh,
	box plot width=1.5em,
	xmin=0.0,
	xmax=1.0,
	boxplot/draw direction=x,
	tick label style=
	{
		/pgf/number format/fixed,
	  /pgf/number format/fixed zerofill,
	  /pgf/number format/precision=1,
	  font=\small
	},
	axis x line* = bottom,
	axis y line* = left,
	ytick={1,2,3},
	xtick={0,0.2,...,1.0},
	y tick label style={anchor=west,xshift= 2mm, yshift=3mm},
	yticklabels={ChiS+RUS, RS+RUS, Linear+RUS},
	legend style={at={(0.5,-0.15)}, anchor=north,legend columns=2},
]

\addplot+[boxplot prepared={box extend=0.3, lower whisker=0.1484, lower quartile=0.4914, median=0.6443, upper quartile=0.7487, upper whisker=0.9772}, ] coordinates{};
\addplot+[boxplot prepared={box extend=0.3, lower whisker=0.2663, lower quartile=0.6003, median=0.7177, upper quartile=0.833, upper whisker=0.9522}, ] coordinates{};
\addplot+[boxplot prepared={box extend=0.3, lower whisker=0.2663, lower quartile=0.5879, median=0.7168, upper quartile=0.8383, upper whisker=0.9522}, ] coordinates{};
\end{axis}
\end{scope}

\begin{scope}[xshift=16.5cm, yshift=-0cm, local bounding box=gbox]
\begin{axis}
[
	title  = \textbf{G-Mean},
	width=6.5cm,
	height=\gbh,
	box plot width=1.5em,
	xmin=0.0,
	xmax=1.0,
	boxplot/draw direction=x,
	tick label style=
	{
		/pgf/number format/fixed,
	  /pgf/number format/fixed zerofill,
	  /pgf/number format/precision=1,
	  font=\small
	},
	axis x line* = bottom,
	axis y line* = left,
	ytick={1,2,3},
	xtick={0,0.2,...,1.0},
	y tick label style={anchor=west,xshift= 2mm, yshift=3mm},
	yticklabels={Linear+RUS, Linear+SMOTE, RS+RUS},
	legend style={at={(0.5,-0.15)}, anchor=north,legend columns=2},
]

\addplot+[boxplot prepared={box extend=0.3, lower whisker=0.6704, lower quartile=0.8096, median=0.8586, upper quartile=0.9095, upper whisker=0.9798}, ] coordinates{(0,0.6435) (0,0.6226) (0,0.6209) (0,0.6125) (0,0.5903) (0,0.5365) (0,0.5292) (0,0.4595) };
\addplot+[boxplot prepared={box extend=0.3, lower whisker=0.1441, lower quartile=0.5506, median=0.7886, upper quartile=0.8879, upper whisker=0.9916}, ] coordinates{(0,0.0000) (0,0.0000) (0,0.0000) (0,0.0000) };
\addplot+[boxplot prepared={box extend=0.3, lower whisker=0.6791, lower quartile=0.8096, median=0.8592, upper quartile=0.9086, upper whisker=0.9798}, ] coordinates{(0,0.6435) (0,0.6136) (0,0.6127) (0,0.6125) (0,0.5903) (0,0.5365) (0,0.5292) (0,0.4595) };
\end{axis}
\end{scope}

\begin{scope}[xshift=22cm, yshift=-0cm, local bounding box=gbox]
\begin{axis}
[
	title  = \textbf{F1},
	width=6.5cm,
	height=\gbh,
	box plot width=1.5em,
	xmin=0.0,
	xmax=1.0,
	boxplot/draw direction=x,
	tick label style=
	{
		/pgf/number format/fixed,
	  /pgf/number format/fixed zerofill,
	  /pgf/number format/precision=1,
	  font=\small
	},
	axis x line* = bottom,
	axis y line* = left,
	ytick={1,2,3},
	xtick={0,0.2,...,1.0},
	y tick label style={anchor=west,xshift= 2mm, yshift=3mm},
	yticklabels={Info+SPIDER, ChiS+OSS, Info+OSS},
	legend style={at={(0.5,-0.15)}, anchor=north,legend columns=2},
]

\addplot+[boxplot prepared={box extend=0.3, lower whisker=0.2503, lower quartile=0.5638, median=0.7595, upper quartile=0.8266, upper whisker=1}, ] coordinates{(0,0.0000) (0,0.0000) (0,0.0000) };
\addplot+[boxplot prepared={box extend=0.3, lower whisker=0.2933, lower quartile=0.584, median=0.7633, upper quartile=0.808, upper whisker=0.9805}, ] coordinates{(0,0.0000) (0,0.0000) (0,0.0000) };
\addplot+[boxplot prepared={box extend=0.3, lower whisker=0.3276, lower quartile=0.5955, median=0.7556, upper quartile=0.8237, upper whisker=0.9775}, ] coordinates{(0,0.0000) (0,0.0000) (0,0.0000) };
\end{axis}
\end{scope}

\end{tikzpicture}
}
\gvsa
\caption{Box plots for the cross-datasets averaged value of each metric, for the Top-3 specific combinations using SVM and  DS+FS pipeline}
\label{gf:12}
\end{figure*}
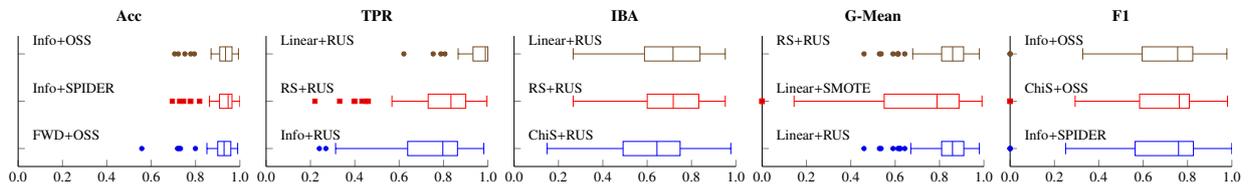

Besides the Rank-Sum values, we also report the absolute cross-dataset averaged accuracy value for each of the top-3 combinations. That is, for each specific combination, we compute the averaged evaluation value of this method across all the 52 datasets.  Shown in Figures \ref{gf:7} and \ref{gf:8}, we use box plots to demonstrate the cross-dataset average accuracy value of the top-3 performers when using the C4.5 classifier. This way, one can intuitively check the general performance of each specific combination under different evaluation metrics. Similarly, in Figures \ref{gf:11} and \ref{gf:12}, we also present the cross-dataset average accuracy values for SVM. It is clear that, in terms of the $G\text{-}Mean$ metric and with SVM, the top-3 performers under the DS+FS pipeline slightly outperform their counterparts under the FS+DS pipeline. 

With the absolute cross-dataset average evaluation values, it is also possible to roughly compare the overall performance of different combinations across different classifiers. For instance, shown in Figures \ref{gf:8} and \ref{gf:12}, in terms of IBA and $G\text{-}Mean$, the top-3 combinations under C4.5 and the DS+FS pipeline, and the top-3 combinations under SVM and the same pipeline, yielded remarkably better performance than the others.

\begin{figure*}[h!]
\label{fig:boxplot_mlp_fsds}
\resizebox{1.0\textwidth}{!}
{
\centering
\begin{tikzpicture}
\pgfsetplotmarksize{0.3ex}
\begin{scope}[xshift=0cm, yshift=-0cm, local bounding box=gbox]
\begin{axis}
[
	title  = \textbf{Acc},
	width=6.5cm,
	height=\gbh,
	box plot width=1.5em,
	xmin=0.0,
	xmax=1.0,
	boxplot/draw direction=x,
	tick label style=
	{
		/pgf/number format/fixed,
	  /pgf/number format/fixed zerofill,
	  /pgf/number format/precision=1,
	  font=\small
	},
	axis x line* = bottom,
	axis y line* = left,
	ytick={1,2,3},
	xtick={0,0.2,...,1.0},
	y tick label style={anchor=west,xshift= 2mm, yshift=3mm},
	yticklabels={, , Info+RUS},
	legend style={at={(0.5,-0.15)}, anchor=north,legend columns=2},
]

\addplot+[boxplot prepared={box extend=0.3, lower whisker=0, lower quartile=0, median=0, upper quartile=0, upper whisker=0}, ] coordinates{};
\addplot+[boxplot prepared={box extend=0.3, lower whisker=0, lower quartile=0, median=0, upper quartile=0, upper whisker=0}, ] coordinates{} node[right,inner sep=1pt,font=\scriptsize,text=black] at (yticklabel cs: 0.5,-5.0) {\parbox{6cm}{Info+CNN , Info+OSS , Info+SMOTE \\ \& Info+SPIDER are identical to Info+RUS}};
\addplot+[boxplot prepared={box extend=0.3, lower whisker=0.7449, lower quartile=0.8715, median=0.9127, upper quartile=0.9565, upper whisker=0.9939}, ] coordinates{(0,0.7230) (0,0.7153) (0,0.6838) (0,0.6521) };
\end{axis}
\end{scope}

\begin{scope}[xshift=5.5cm, yshift=-0cm, local bounding box=gbox]
\begin{axis}
[
	title  = \textbf{TPR},
	width=6.5cm,
	height=\gbh,
	box plot width=1.5em,
	xmin=0.0,
	xmax=1.0,
	boxplot/draw direction=x,
	tick label style=
	{
		/pgf/number format/fixed,
	  /pgf/number format/fixed zerofill,
	  /pgf/number format/precision=1,
	  font=\small
	},
	axis x line* = bottom,
	axis y line* = left,
	ytick={1,2,3},
	xtick={0,0.2,...,1.0},
	y tick label style={anchor=west,xshift= 2mm, yshift=3mm},
	yticklabels={, FWD+RUS, Linear+ADASYN},
	legend style={at={(0.5,-0.15)}, anchor=north,legend columns=2},
]

\addplot+[boxplot prepared={box extend=0.3, lower whisker=0, lower quartile=0, median=0, upper quartile=0, upper whisker=0}, ] coordinates{} node[right,inner sep=1pt,font=\scriptsize,text=black] at (yticklabel cs: 0.15,-5.0) {\parbox{6cm}{FWD+CNN , FWD+OSS , FWD+SMOTE \\ \& FWD+SPIDER are identical to FWD+RUS}};
\addplot+[boxplot prepared={box extend=0.3, lower whisker=0.2358, lower quartile=0.608, median=0.7852, upper quartile=0.8885, upper whisker=0.997}, ] coordinates{(0,0.1600) (0,0.1000) };
\addplot+[boxplot prepared={box extend=0.3, lower whisker=0.3211, lower quartile=0.605, median=0.7794, upper quartile=0.873, upper whisker=0.9878}, ] coordinates{(0,0.0571) };
\end{axis}
\end{scope}

\begin{scope}[xshift=11cm, yshift=-0cm, local bounding box=gbox]
\begin{axis}
[
	title  = \textbf{IBA},
	width=6.5cm,
	height=\gbh,
	box plot width=1.5em,
	xmin=0.0,
	xmax=1.0,
	boxplot/draw direction=x,
	tick label style=
	{
		/pgf/number format/fixed,
	  /pgf/number format/fixed zerofill,
	  /pgf/number format/precision=1,
	  font=\small
	},
	axis x line* = bottom,
	axis y line* = left,
	ytick={1,2,3},
	xtick={0,0.2,...,1.0},
	y tick label style={anchor=west,xshift= 2mm, yshift=3mm},
	yticklabels={, , FWD+RUS},
	legend style={at={(0.5,-0.15)}, anchor=north,legend columns=2},
]

\addplot+[boxplot prepared={box extend=0.3, lower whisker=0, lower quartile=0, median=0, upper quartile=0, upper whisker=0}, ] coordinates{};
\addplot+[boxplot prepared={box extend=0.3, lower whisker=0, lower quartile=0, median=0, upper quartile=0, upper whisker=0}, ] coordinates{} node[right,inner sep=1pt,font=\scriptsize,text=black] at (yticklabel cs: 0.5,-5.0) {\parbox{6cm}{FWD+CNN , FWD+PSS , FWD+SMOTE \\ FWD+SPIDER\& FWD+ADASYN \\ are identical to FWD+RUS}};
\addplot+[boxplot prepared={box extend=0.3, lower whisker=0.0538, lower quartile=0.4664, median=0.6737, upper quartile=0.8041, upper whisker=0.9919}, ] coordinates{};
\end{axis}
\end{scope}

\begin{scope}[xshift=16.50cm, yshift=0cm, local bounding box=gbox]
\begin{axis}
[
	title  = \textbf{G-Mean},
	width=6.5cm,
	height=\gbh,
	box plot width=1.5em,
	xmin=0.0,
	xmax=1.0,
	boxplot/draw direction=x,
	tick label style=
	{
		/pgf/number format/fixed,
	  /pgf/number format/fixed zerofill,
	  /pgf/number format/precision=1,
	  font=\small
	},
	axis x line* = bottom,
	axis y line* = left,
	ytick={1,2,3},
	xtick={0,0.2,...,1.0},
	y tick label style={anchor=west,xshift= 2mm, yshift=3mm},
	yticklabels={, FWD+RUS, Linear+ADASYN},
	legend style={at={(0.5,-0.15)}, anchor=north,legend columns=2},
]

\addplot+[boxplot prepared={box extend=0.3, lower whisker=0, lower quartile=0, median=0, upper quartile=0, upper whisker=0}, ] coordinates{} node[right,inner sep=1pt,font=\scriptsize,text=black] at (yticklabel cs: 0.15,-5.0) {\parbox{6cm}{FWD+CNN , FWD+OSS , FWD+SMOTE \\ \& FWD+SPIDER are identical to FWD+RUS}};
\addplot+[boxplot prepared={box extend=0.3, lower whisker=0.4604, lower quartile=0.719, median=0.8524, upper quartile=0.8964, upper whisker=0.9949}, ] coordinates{(0,0.4140) (0,0.3927) (0,0.3854) (0,0.3614) (0,0.2775) (0,0.1080) };
\addplot+[boxplot prepared={box extend=0.3, lower whisker=0.4895, lower quartile=0.7201, median=0.8447, upper quartile=0.8935, upper whisker=0.9824}, ] coordinates{(0,0.0816) };
\end{axis}
\end{scope}

\begin{scope}[xshift=22cm, yshift=0cm, local bounding box=gbox]
\begin{axis}
[
	title  = \textbf{F1},
	width=6.5cm,
	height=\gbh,
	box plot width=1.5em,
	xmin=0.0,
	xmax=1.0,
	boxplot/draw direction=x,
	tick label style=
	{
		/pgf/number format/fixed,
	  /pgf/number format/fixed zerofill,
	  /pgf/number format/precision=1,
	  font=\small
	},
	axis x line* = bottom,
	axis y line* = left,
	ytick={1,2,3},
	xtick={0,0.2,...,1.0},
	y tick label style={anchor=west,xshift= 2mm, yshift=3mm},
	yticklabels={, , Info+RUS},
	legend style={at={(0.5,-0.15)}, anchor=north,legend columns=2},
]

\addplot+[boxplot prepared={box extend=0.3, lower whisker=0, lower quartile=0, median=0, upper quartile=0, upper whisker=0}, ] coordinates{};
\addplot+[boxplot prepared={box extend=0.3, lower whisker=0, lower quartile=0, median=0, upper quartile=0, upper whisker=0}, ] coordinates{} node[right,inner sep=1pt,font=\scriptsize,text=black] at (yticklabel cs: 0.5,-5.0) {\parbox{6cm}{Info+CNN , Info+OSS , Info+SMOTE \\ \& Info+SPIDER are identical to Info+RUS}};
\addplot+[boxplot prepared={box extend=0.3, lower whisker=0.06, lower quartile=0.5331, median=0.6869, upper quartile=0.8565, upper whisker=0.98}, ] coordinates{};
\end{axis}
\end{scope}

\end{tikzpicture}
}
\gvsa
\caption{Box plots for the cross-datasets averaged value of each metric, for the Top-3 specific combinations using MLP and FS+DS pipeline}
\label{gf:9}
\end{figure*}
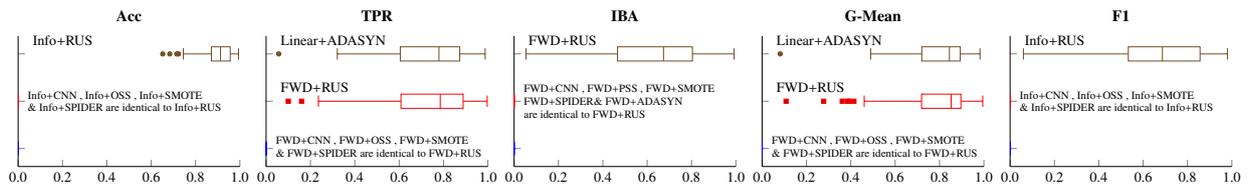

\textbf{MLP}. For MLP under the FS+DS pipeline, the patterns are quite different from C4.5 and SVM. From Figure \ref{gf:9}, we observe that, in terms of $G\text{-}Mean$, the best performer is Linear+ADASYN, while in terms of IBA, the best method is FWD+RUS. But for the other combinations, it seems that the influence of data re-sampling on the performance is insignificant, and what matters is the feature selection method. For instance, in terms of IBA, the best feature selection method is constantly FWD; in terms of $G\text{-}Mean$, except the top-1 combination (Linear+ADASYN), FWD is also the best feature selection method for the rest  top-performing methods/combinations. Still, in terms of F1, Info is constantly the best feature selection method for all the top performers. Therefore, for MLP under the FS+DS pipeline, one can resort to a two-stage method for efficiently deriving the best combination. In the first stage,

\onecolumn


\captionsetup[figure]{aboveskip=0pt, belowskip=0pt}

\gvsb

\gvsb

\gvsb

\gvsb

\gvsb

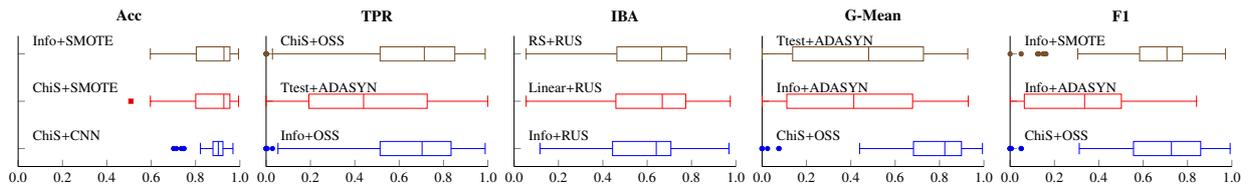
\begin{figure*}[hb!]
\label{fig:boxplot_mlp_dsfs}
\resizebox{1.0\textwidth}{!}
{
\centering
\begin{tikzpicture}
\pgfsetplotmarksize{0.3ex}
\begin{scope}[xshift=0cm, yshift=-0cm, local bounding box=gbox]
\begin{axis}
[
	title  = \textbf{Acc},
	width=6.5cm,
	height=\gbh,
	box plot width=1.5em,
	xmin=0.0,
	xmax=1.0,
	boxplot/draw direction=x,
	tick label style=
	{
		/pgf/number format/fixed,
	  /pgf/number format/fixed zerofill,
	  /pgf/number format/precision=1,
	  font=\small
	},
	axis x line* = bottom,
	axis y line* = left,
	ytick={1,2,3},
	xtick={0,0.2,...,1.0},
	y tick label style={anchor=west,xshift= 2mm, yshift=3mm},
	yticklabels={ChiS+CNN, ChiS+SMOTE, Info+SMOTE},
	legend style={at={(0.5,-0.15)}, anchor=north,legend columns=2},
]

\addplot+[boxplot prepared={box extend=0.3, lower whisker=0.8221, lower quartile=0.8786, median=0.9027, upper quartile=0.9236, upper whisker=0.9688}, ] coordinates{(0,0.7483) (0,0.7435) (0,0.7353) (0,0.7140) (0,0.7008) };
\addplot+[boxplot prepared={box extend=0.3, lower whisker=0.5964, lower quartile=0.8008, median=0.9277, upper quartile=0.9549, upper whisker=0.994}, ] coordinates{(0,0.5085) };
\addplot+[boxplot prepared={box extend=0.3, lower whisker=0.5964, lower quartile=0.8033, median=0.9282, upper quartile=0.9551, upper whisker=0.9941}, ] coordinates{};
\end{axis}
\end{scope}

\begin{scope}[xshift=5.5cm, yshift=-0cm, local bounding box=gbox]
\begin{axis}
[
	title  = \textbf{TPR},
	width=6.5cm,
	height=\gbh,
	box plot width=1.5em,
	xmin=0.0,
	xmax=1.0,
	boxplot/draw direction=x,
	tick label style=
	{
		/pgf/number format/fixed,
	  /pgf/number format/fixed zerofill,
	  /pgf/number format/precision=1,
	  font=\small
	},
	axis x line* = bottom,
	axis y line* = left,
	ytick={1,2,3},
	xtick={0,0.2,...,1.0},
	y tick label style={anchor=west,xshift= 2mm, yshift=3mm},
	yticklabels={Info+OSS, Ttest+ADASYN, ChiS+OSS},
	legend style={at={(0.5,-0.15)}, anchor=north,legend columns=2},
]

\addplot+[boxplot prepared={box extend=0.3, lower whisker=0.0536, lower quartile=0.5139, median=0.7034, upper quartile=0.8341, upper whisker=0.9879}, ] coordinates{(0,0.0286) (0,0.0049) (0,0.0000) (0,0.0000) (0,0.0000) (0,0.0000) };
\addplot+[boxplot prepared={box extend=0.3, lower whisker=0, lower quartile=0.193, median=0.4397, upper quartile=0.7264, upper whisker=1}, ] coordinates{};
\addplot+[boxplot prepared={box extend=0.3, lower whisker=0.0286, lower quartile=0.5139, median=0.7134, upper quartile=0.851, upper whisker=0.9879}, ] coordinates{(0,0.0030) (0,0.0000) (0,0.0000) (0,0.0000) (0,0.0000) };
\end{axis}
\end{scope}

\begin{scope}[xshift=11cm, yshift=-0cm, local bounding box=gbox]
\begin{axis}
[
	title  = \textbf{IBA},
	width=6.5cm,
	height=\gbh,
	box plot width=1.5em,
	xmin=0.0,
	xmax=1.0,
	boxplot/draw direction=x,
	tick label style=
	{
		/pgf/number format/fixed,
	  /pgf/number format/fixed zerofill,
	  /pgf/number format/precision=1,
	  font=\small
	},
	axis x line* = bottom,
	axis y line* = left,
	ytick={1,2,3},
	xtick={0,0.2,...,1.0},
	y tick label style={anchor=west,xshift= 2mm, yshift=3mm},
	yticklabels={Info+RUS, Linear+RUS, RS+RUS},
	legend style={at={(0.5,-0.15)}, anchor=north,legend columns=2},
]

\addplot+[boxplot prepared={box extend=0.3, lower whisker=0.1166, lower quartile=0.444, median=0.641, upper quartile=0.7072, upper whisker=0.9688}, ] coordinates{};
\addplot+[boxplot prepared={box extend=0.3, lower whisker=0.0544, lower quartile=0.4588, median=0.6681, upper quartile=0.774, upper whisker=0.9749}, ] coordinates{};
\addplot+[boxplot prepared={box extend=0.3, lower whisker=0.0544, lower quartile=0.4634, median=0.6648, upper quartile=0.7779, upper whisker=0.9749}, ] coordinates{};
\end{axis}
\end{scope}

\begin{scope}[xshift=16.5cm, yshift=0cm, local bounding box=gbox]
\begin{axis}
[
	title  = \textbf{G-Mean},
	width=6.5cm,
	height=\gbh,
	box plot width=1.5em,
	xmin=0.0,
	xmax=1.0,
	boxplot/draw direction=x,
	tick label style=
	{
		/pgf/number format/fixed,
	  /pgf/number format/fixed zerofill,
	  /pgf/number format/precision=1,
	  font=\small
	},
	axis x line* = bottom,
	axis y line* = left,
	ytick={1,2,3},
	xtick={0,0.2,...,1.0},
	y tick label style={anchor=west,xshift= 2mm, yshift=3mm},
	yticklabels={ChiS+OSS, Info+ADASYN, Ttest+ADASYN},
	legend style={at={(0.5,-0.15)}, anchor=north,legend columns=2},
]

\addplot+[boxplot prepared={box extend=0.3, lower whisker=0.4395, lower quartile=0.6826, median=0.8238, upper quartile=0.8991, upper whisker=0.9937}, ] coordinates{(0,0.0756) (0,0.0756) (0,0.0245) (0,0.0000) (0,0.0000) (0,0.0000) (0,0.0000) };
\addplot+[boxplot prepared={box extend=0.3, lower whisker=0, lower quartile=0.1114, median=0.4135, upper quartile=0.6787, upper whisker=0.9295}, ] coordinates{};
\addplot+[boxplot prepared={box extend=0.3, lower whisker=0, lower quartile=0.1377, median=0.4807, upper quartile=0.7276, upper whisker=0.9273}, ] coordinates{};
\end{axis}
\end{scope}

\begin{scope}[xshift=22cm, yshift=0cm, local bounding box=gbox]
\begin{axis}
[
	title  = \textbf{F1},
	width=6.5cm,
	height=\gbh,
	box plot width=1.5em,
	xmin=0.0,
	xmax=1.0,
	boxplot/draw direction=x,
	tick label style=
	{
		/pgf/number format/fixed,
	  /pgf/number format/fixed zerofill,
	  /pgf/number format/precision=1,
	  font=\small
	},
	axis x line* = bottom,
	axis y line* = left,
	ytick={1,2,3},
	xtick={0,0.2,...,1.0},
	y tick label style={anchor=west,xshift= 2mm, yshift=3mm},
	yticklabels={ChiS+OSS, Info+ADASYN, Info+SMOTE},
	legend style={at={(0.5,-0.15)}, anchor=north,legend columns=2},
]

\addplot+[boxplot prepared={box extend=0.3, lower whisker=0.3118, lower quartile=0.5565, median=0.7275, upper quartile=0.8594, upper whisker=0.9923}, ] coordinates{(0,0.0500) (0,0.0500) (0,0.0057) (0,0.0000) (0,0.0000) (0,0.0000) (0,0.0000) };
\addplot+[boxplot prepared={box extend=0.3, lower whisker=0, lower quartile=0.0641, median=0.3358, upper quartile=0.5018, upper whisker=0.8408}, ] coordinates{};
\addplot+[boxplot prepared={box extend=0.3, lower whisker=0.3052, lower quartile=0.5841, median=0.7072, upper quartile=0.7775, upper whisker=0.9711}, ] coordinates{(0,0.1617) (0,0.1495) (0,0.1295) (0,0.1241) (0,0.0500) (0,0.0000) };
\end{axis}
\end{scope}

\end{tikzpicture}
}
\gvsa
\caption{Box plots for the cross-datasets averaged value of each metric, for the Top-3 specific combinations using MLP and DS+FS pipeline}
\label{gf:10}
\end{figure*}


\twocolumn

\noindent  one can find which feature selection methods yield the best performance for a given evaluation metric. In the second stage, one only needs to combine the selected feature selection method(s) with different re-sampling methods. This way, the computation cost can be substantially reduced.  But the above patterns do not hold for MLP with the DS+FS pipeline. From Figure \ref{gf:10}, we see that Chis+OSS is generally the  winner in terms of $G\text{-}Mean$, TPR and F1, since it obtains the best cross-dataset averaged value for each of these three metrics. But IBA under the same DS+FS pipeline, the best combination is  RS+RUS, and RUS is the best data re-sampling method for all the top-3 combinations. 

\textbf{Discussions}. Comparing results from the statistical Iman-Davenport Tests in subsection \ref{imantest} and the heuristic Top-K measures based on Rank-SUM in this subsection, we can check whether the findings or conclusions from these two different measures are generally consistent or not. When comparing Table~\ref{tab:tabella_risultati_C45} and Figures \ref{gf:1} and \ref{gf:2}, we see that for IBA under the C4.5 classifier, the specific combinations of SMOTE+RS and SMOTE+Linear under the DS+FS pipeline, and  Linear+RUS and Linear+RUS under the FS+DS pipeline, are consistently among the top-3 best performers for both the statistical Iman-Davenport Tests and the heuristic Top-K measures. For SVM, the top performers using these two different measures are partially consistent.  For instance, the specific combinations of ChiS+RUS and Ttest+RUS under the FS+DS pipeline are among the top-3 performers for both measures. Similar observations also hold for MLP. For instance, RUS+RS and RUS+Linear under the DS+FS pipeline, and FWD+RUS and FWD+CNN under the FS+DS pipeline, are among the top performers for both measures. 

Overall, Iman-Davenport Test is a stronger statistical measure, but it might yield too many candidates as top performers; it also needs to check both the Rank value and the $\#p$ value.  The ideal case is to have a small Rank value but a large $\#p$ value; however, it will be extremely hard to pick the specific combinations when they present large Rank value but low $\#p$ value. The  Top-K measure based on Rank-SUM is heuristic yet  straightforward, it recommends the top-3 combinations/methods for each metric, based on the number of datasets that a method ranks among the top-3 best performers. Therefore, the joint use of these two measures may provide more reliable results. For instance, one can only pick the specific combinations that are the top performers for both measures.


\section{Concluding Remarks}  \label{section:considANDrecomm}
This work carried out an extensive empirical study to investigate the performance of the feature selection before/after data re-sampling pipelines for improving two-class imbalance learning.  From this study, we can draw the following conclusions:

\begin{enumerate}
\item To derive the best performing imbalance classification model, both the feature selection before and after data re-sampling pipelines should be considered; in particular, the feature selection after data  re-sampling pipeline deserves significantly more attention from the community, because it was commonly overlooked before.

\item When using the decision tree as base learner,  DS+FS outperforms FS+DS in more cases than not, in terms of $G\text{-}Mean$ and $F1$, and this pattern is more evident when using oversampling methods.

\item When the base classifier is SVM,  with undersampling methods, FS+DS  outperforms DS+FS in general. However, when using oversampling methods, there is no clear winner between the two pipelines. We argue that a preliminary identification of most salient descriptors helps the kernel improve the linear separation between classes; thus, the learner derived is less sensitive to oversampling and undersampling.  

\item Besides the base learners, imbalance ratio of the datasets and samples-to-feature ratio also have a great influence on the performance of the two pipelines. On datasets with high imbalance ratios, FS+DS outperforms DS+FS in general in terms of the $G\text{-}Mean$, $F1$, and $IBA$. But for datasets with a large value of the SFR but a small value of IR, there is no predominance between the two pipelines. Similar considerations also hold for datasets with medium SFR and IR values.

\end{enumerate}

In general, we have seen that the choice of the two pipelines depends on the specific dataset, the  combinations of feature selection and data re-sampling algorithms, as well as the adopted classification algorithms.  We have also provided heuristic yet practical guidance for selecting the Top-K specific combinations for  each of the evaluation metrics and the two pipelines. 

Overall, this study provides new empirical findings on how to derive the best imbalance classification model under different configurations of pipelines, and different combinations of feature selection and re-sampling methods. We suggest the readers to consider both pipelines when looking for the best imbalance classification model. 


\section*{Declaration of Competing Interest}
We declare that there is no conflict of interest.

\section*{Declaration of Contributions}
All the authors discussed the results and contributed to the final manuscript. Their detailed contributions are specified below. 

Chongsheng Zhang: Conceptualization, Methodology, Validation,  Formal analysis, Investigation,  Writing-Review \& Editing.
 
Paolo Soda:  Conceptualization, Methodology, Validation,  Software, Formal analysis, Writing-Original Draft, Visualization.

Jingjun Bi: Software.

Gaojuan Fan: Resources, Writing-Review \& Editing, Supervision, Project administration.

George Almpanidis: Formal analysis, Writing-Review \& Editing, Visualization.

Salvador Garc\'{\i}a: Writing-Review \& Editing.

\section*{Acknowledgments}
This work has been partially supported by the research project TIN2017-89517-P.



\bibliographystyle{elsarticle/elsarticle-num}

\bibliography{FSDS_refs}


\clearpage
\onecolumn

\appendix
\section{}
\setcounter{table}{0}
\setcounter{figure}{0}

\begin{table}[H]
\centering
\caption{Summary of the used datasets. IR and  SFR denote the imbalance ratio  and the samples-to-features ratio, respectively.}
\label{tab:summaryDataset}
\resizebox{0.8\textwidth}{!}{
\begin{tabular}{@{}lccccc@{}}
\toprule

\textbf{Dataset}                    & \textbf{\#  instances  } & \textbf{\#  features } & \textbf{Prior minority class (\%) }  & \textbf{IR} & \textbf{SFR} \\ 

\midrule

BioCells & 489 & 64 & 20.45 & 3.89 & 7.64 \\
cleveland & 173 & 13 & 7.51 & 12.31 & 13.31  \\
ecoli-0-1-4-7\_vs\_2-3-5-6 & 336 & 7 & 8.63 & 10.59 & 48.00 \\
ecoli-0-1-4-7\_vs\_5-6 & 332 & 6 & 7.53 & 12.28 & 55.33 \\
ecoli-0-2-6-7\_vs\_3-5 & 224 & 7 & 9.82 & 9.18 & 32.00 \\
ecoli-0-3-4-6\_vs\_5 & 205 & 7 & 9.76 & 9.25 & 29.29 \\
ecoli-0-3-4-7\_vs\_5-6 & 257 & 7 & 9.73 & 9.28 & 36.71 \\
ecoli-0-6-7\_vs\_5 & 220 & 6 & 9.09 & 10.00 & 36.67 \\
ecoli1 & 336 & 7 & 22.92 & 3.36 & 48.00 \\
ecoli2 & 336 & 7 & 15.48 & 5.46 & 48.00 \\
ecoli3 & 336 & 7 & 10.42 & 8.60 & 48.00 \\
ecoli4 & 336 & 7 & 5.95 & 15.80 & 48.00 \\
german & 1000 & 38 & 30.00 & 2.33 & 26.32 \\
glass-0-1-2-3\_vs\_4-5-6 & 214 & 9 & 23.83 & 3.20 & 23.78 \\
glass-0-1-6\_vs\_2 & 336 & 9 & 11.01 & 8.08 & 37.33 \\
glass-0-1-6\_vs\_5 & 336 & 9 & 11.01 & 8.08 & 37.33 \\
glass0 & 214 & 9 & 32.71 & 2.06 & 23.78 \\
glass2 & 336 & 9 & 11.01 & 8.08 & 37.33 \\
glass4 & 336 & 9 & 9.82 & 9.18 & 37.33 \\
glass5 & 336 & 9 & 8.63 & 10.59 & 37.33 \\
glass6 & 214 & 9 & 13.55 & 6.38 & 23.78 \\
haberman & 306 & 3 & 26.47 & 2.78 & 102.00 \\
Insurance\_Company & 9822 & 85 & 5.97 & 15.76 & 115.55 \\ 
led7digit & 443 & 7 & 8.35 & 10.97 & 63.29 \\
newthyroid & 215 & 5 & 16.28 & 5.14 & 43.00 \\
optdigit\_class0-1vs\_2-9 & 5620 & 64 & 10.16 & 8.84 & 87.81 \\
optdigit\_class0-2vs\_3-9 & 5620 & 64 & 20.07 & 3.98 & 87.81 \\
optdigit\_class0-3vs\_4-9 & 5620 & 64 & 30.25 & 2.31 & 87.81 \\
pageblocks & 472 & 10 & 5.93 & 15.86 & 47.20 \\
pima & 768 & 8 & 34.90 & 1.87 & 96.00 \\
segment & 2308 & 19 & 14.26 & 6.02 & 121.47 \\
semeion\_class1-2vs\_3-9 & 1593 & 256 & 20.15 & 3.96 & 6.22 \\
semeion\_class1-3vs\_4-9 & 1593 & 256 & 30.13 & 2.32 & 6.22 \\
semeion\_class1vs\_2-9 & 1593 & 256 & 10.17 & 8.83 & 6.22 \\
vehicle & 846 & 18 & 25.77 & 2.88 & 47.00 \\
vowel0 & 1829 & 13 & 11.65 & 7.59 & 140.69 \\
waveform\_no\_noise\_class0vs\_1-2 & 5000 & 21 & 66.86 & 0.50 & 238.10 \\
waveform\_no\_noise\_class2vs\_0-1 & 5000 & 21 & 33.92 & 1.95 & 238.10 \\
waveform\_yes\_noise\_class2vs\_0-1 & 5000 & 40 & 33.10 & 2.02 & 125.00 \\
winequality-red-4 & 1599 & 11 & 3.31 & 29.17 & 145.36 \\
yeast-0-2-5-6\_vs\_3-7-8-9 & 1004 & 8 & 9.86 & 9.14 & 125.50 \\
yeast-0-2-5-7-9\_vs\_3-6-8 & 1004 & 8 & 9.86 & 9.14 & 125.50 \\
yeast-0-3-5-9\_vs\_7-8 & 506 & 8 & 9.88 & 9.12 & 63.25 \\
yeast-0-5-6-7-9\_vs\_4 & 1829 & 13 & 11.81 & 7.47 & 140.69 \\
yeast-1-2-8-9\_vs\_7 & 1829 & 13 & 8.58 & 10.65 & 140.69 \\
yeast-1-4-5-8\_vs\_7 & 1829 & 13 & 10.22 & 8.78 & 140.69 \\
yeast-1\_vs\_7 & 1829 & 13 & 11.86 & 7.43 & 140.69 \\
yeast-2\_vs\_4 & 1829 & 13 & 13.01 & 6.68 & 140.69 \\
yeast-2\_vs\_8 & 1829 & 13 & 13.07 & 6.65 & 140.69 \\
yeast4 & 1829 & 13 & 9.51 & 9.51 & 140.69 \\
yeast5 & 1829 & 13 & 9.13 & 9.95 & 140.69 \\
yeast6 & 1829 & 13 & 8.64 & 10.58 & 140.69 \\

\bottomrule   
\end{tabular}}
\end{table}




\begin{table}
\centering
\caption{Wilcoxon test for the comparison between FS+DS ($R^{+}$) and DS+FS ($R^{-}$), when the SVM classifier is  used.}
\label{tab:tabella_risultati_SVM}
\resizebox{1\textwidth}{!}
{
\begin{tabular}{llllllllllllll}
\toprule 
\toprule
\multirow{2}{*}{FS} & \multirow{2}{*}{DS} & \multicolumn{3}{l}{Accuracy (Acc)} & \multicolumn{3}{l}{G-Mean (g)} & \multicolumn{3}{l}{F1} & \multicolumn{3}{l}{IBA} \\ \cline{3-14} 
                    &                     & R+      & R-       & $p$-value     & R+      & R-     & $p$-value   & R+   & R-  & $p$-value & R+    & R-  & $p$-value \\ 
                   \hline
 
Info & RUS & 1185 & 141 & 9.9e-07 & 1156 & 222 & 2.1e-05 & 1217 & 161 & 1.5e-06 & 644 & 734 & 6.8e-01 \\
ChiS & RUS & 1148 & 178 & 5.5e-06 & 1083 & 243 & 8.3e-05 & 1172 & 154 & 1.8e-06 & 518 & 757 & 2.5e-01 \\
Fish & RUS & 1137.5 & 188.5 & 8.7e-06 & 1136 & 190 & 9.3e-06 & 1146 & 180 & 6.0e-06 & 679 & 647 & 8.8e-01 \\
Gini & RUS & 1141 & 185 & 7.4e-06 & 1114 & 264 & 1.1e-04 & 1175 & 203 & 9.6e-06 & 552 & 826 & 2.1e-01 \\
SBMLR & RUS & 1136.5 & 189.5 & 9.1e-06 & 930 & 345 & 4.7e-03 & 994 & 281 & 5.8e-04 & 626 & 649 & 9.1e-01 \\
Ttest & RUS & 1007 & 218 & 8.7e-05 & 1201 & 125 & 4.6e-07 & 1173 & 153 & 1.7e-06 & 791 & 484 & 1.4e-01 \\
RS & RUS & 635 & 640 & 9.8e-01 & 663 & 663 & 1.0e+00 & 700 & 626 & 7.3e-01 & 668 & 607 & 7.7e-01 \\
Linear & RUS & 633.5 & 641.5 & 9.7e-01 & 530 & 796 & 2.1e-01 & 621 & 705 & 6.9e-01 & 440 & 886 & 3.7e-02 \\
FWD & RUS & 1215 & 163 & 1.7e-06 & 1288 & 90 & 4.9e-08 & 1315 & 63 & 1.2e-08 & 817 & 561 & 2.4e-01 \\
Info & CNN & 1053 & 222 & 6.0e-05 & 1036 & 239 & 1.2e-04 & 1126 & 149 & 2.4e-06 & 432 & 843 & 4.7e-02 \\
ChiS & CNN & 1032 & 243 & 1.4e-04 & 1020 & 255 & 2.2e-04 & 1118 & 157 & 3.5e-06 & 445 & 830 & 6.3e-02 \\
Fish & CNN & 893.5 & 331.5 & 5.2e-03 & 912 & 363 & 8.1e-03 & 982 & 293 & 8.8e-04 & 488 & 737 & 2.2e-01 \\
Gini & CNN & 965 & 310 & 1.6e-03 & 882 & 343 & 7.3e-03 & 958 & 267 & 5.9e-04 & 525 & 750 & 2.8e-01 \\
SBMLR & CNN & 443 & 685 & 2.0e-01 & 678 & 312 & 3.3e-02 & 787 & 294 & 7.1e-03 & 631 & 450 & 3.2e-01 \\
Ttest & CNN & 925.5 & 202.5 & 1.3e-04 & 925 & 401 & 1.4e-02 & 1002 & 324 & 1.5e-03 & 419 & 856 & 3.5e-02 \\
RS & CNN & 916 & 165 & 4.1e-05 & 958 & 123 & 5.1e-06 & 964 & 117 & 3.7e-06 & 457 & 624 & 3.6e-01 \\
Linear & CNN & 946 & 182 & 5.3e-05 & 916 & 165 & 4.1e-05 & 951 & 130 & 7.3e-06 & 426 & 655 & 2.1e-01 \\
FWD & CNN & 975.5 & 350.5 & 3.4e-03 & 963 & 363 & 4.9e-03 & 1034 & 292 & 5.1e-04 & 385 & 941 & 9.2e-03 \\
Info & OSS & 596 & 580 & 9.3e-01 & 541 & 635 & 6.3e-01 & 610 & 566 & 8.2e-01 & 575 & 601 & 8.9e-01 \\
ChiS & OSS & 591.5 & 633.5 & 8.3e-01 & 486 & 642 & 4.1e-01 & 516 & 612 & 6.1e-01 & 575 & 506 & 7.1e-01 \\
Fish & OSS & 381.5 & 746.5 & 5.3e-02 & 368 & 808 & 2.4e-02 & 364 & 812 & 2.2e-02 & 655 & 521 & 4.9e-01 \\
Gini & OSS & 514.5 & 760.5 & 2.4e-01 & 485 & 740 & 2.0e-01 & 509.5 & 715.5 & 3.1e-01 & 695 & 530 & 4.1e-01 \\
SBMLR & OSS & 436.5 & 466.5 & 8.5e-01 & 263 & 683 & 1.1e-02 & 279 & 667 & 1.9e-02 & 456 & 490 & 8.4e-01 \\
Ttest & OSS & 332.5 & 795.5 & 1.4e-02 & 479 & 697 & 2.6e-01 & 379 & 797 & 3.2e-02 & 648 & 528 & 5.4e-01 \\
RS & OSS & 603.5 & 524.5 & 6.8e-01 & 608 & 568 & 8.4e-01 & 640 & 536 & 5.9e-01 & 563 & 613 & 8.0e-01 \\
Linear & OSS & 433 & 602 & 3.4e-01 & 559 & 617 & 7.7e-01 & 469 & 707 & 2.2e-01 & 620 & 508 & 5.5e-01 \\
FWD & OSS & 560 & 568 & 9.7e-01 & 531 & 645 & 5.6e-01 & 629 & 547 & 6.7e-01 & 594 & 582 & 9.5e-01 \\
Info & SMOTE & 800 & 328 & 1.3e-02 & 755 & 421 & 8.7e-02 & 742 & 434 & 1.1e-01 & 543 & 585 & 8.2e-01 \\
ChiS & SMOTE & 840.5 & 335.5 & 9.6e-03 & 652 & 524 & 5.1e-01 & 677 & 451 & 2.3e-01 & 409 & 719 & 1.0e-01 \\
Fish & SMOTE & 802 & 374 & 2.8e-02 & 813 & 363 & 2.1e-02 & 789 & 387 & 3.9e-02 & 554 & 574 & 9.2e-01 \\
Gini & SMOTE & 829.5 & 346.5 & 1.3e-02 & 813 & 363 & 2.1e-02 & 789 & 387 & 3.9e-02 & 515 & 613 & 6.0e-01 \\
SBMLR & SMOTE & 684 & 351 & 6.0e-02 & 521 & 514 & 9.7e-01 & 560 & 475 & 6.3e-01 & 453 & 582 & 4.7e-01 \\
Ttest & SMOTE & 779.5 & 445.5 & 9.7e-02 & 735 & 490 & 2.2e-01 & 665 & 560 & 6.0e-01 & 567 & 658 & 6.5e-01 \\
RS & SMOTE & 427 & 749 & 9.9e-02 & 654 & 522 & 5.0e-01 & 655 & 521 & 4.9e-01 & 706 & 422 & 1.3e-01 \\
Linear & SMOTE & 322 & 759 & 1.7e-02 & 620 & 556 & 7.4e-01 & 580 & 596 & 9.3e-01 & 647 & 434 & 2.4e-01 \\
FWD & SMOTE & 869.5 & 306.5 & 3.9e-03 & 938 & 238 & 3.3e-04 & 939 & 237 & 3.2e-04 & 563 & 565 & 9.9e-01 \\
Info & SPIDER & 222 & 306 & 4.3e-01 & 279 & 351 & 5.6e-01 & 280 & 350 & 5.7e-01 & 302 & 293 & 9.4e-01 \\
ChiS & SPIDER & 227.5 & 402.5 & 1.5e-01 & 288 & 378 & 4.8e-01 & 247 & 419 & 1.8e-01 & 327 & 339 & 9.2e-01 \\
Fish & SPIDER & 216.5 & 524.5 & 2.6e-02 & 271 & 470 & 1.5e-01 & 213 & 528 & 2.2e-02 & 441 & 300 & 3.1e-01 \\
Gini & SPIDER & 353 & 467 & 4.4e-01 & 427.5 & 475.5 & 7.6e-01 & 363 & 540 & 2.7e-01 & 397 & 464 & 6.6e-01 \\
SBMLR & SPIDER & 353.5 & 466.5 & 4.5e-01 & 328 & 533 & 1.8e-01 & 349 & 512 & 2.9e-01 & 412 & 449 & 8.1e-01 \\
Ttest & SPIDER & 257 & 409 & 2.3e-01 & 252 & 378 & 3.0e-01 & 246 & 384 & 2.6e-01 & 312 & 318 & 9.6e-01 \\
RS & SPIDER & 307.5 & 322.5 & 9.0e-01 & 288 & 342 & 6.6e-01 & 291 & 304 & 9.1e-01 & 193 & 402 & 7.4e-02 \\
Linear & SPIDER & 241 & 287 & 6.7e-01 & 240 & 321 & 4.7e-01 & 244 & 284 & 7.1e-01 & 138 & 358 & 3.1e-02 \\
FWD & SPIDER & 365.5 & 300.5 & 6.1e-01 & 406.5 & 296.5 & 4.1e-01 & 340.5 & 362.5 & 8.7e-01 & 375 & 328 & 7.2e-01 \\
Info & ADASYN & 866 & 512 & 1.1e-01 & 1310 & 68 & 1.6e-08 & 1199 & 179 & 3.4e-06 & 789 & 589 & 3.6e-01 \\
ChiS & ADASYN & 937 & 441 & 2.4e-02 & 1327 & 51 & 6.2e-09 & 1255 & 123 & 2.5e-07 & 712 & 666 & 8.3e-01 \\
Fish & ADASYN & 1052 & 326 & 9.5e-04 & 1290 & 88 & 4.4e-08 & 1287 & 91 & 5.2e-08 & 549 & 829 & 2.0e-01 \\
Gini & ADASYN & 1036 & 290 & 4.7e-04 & 1341 & 37 & 2.9e-09 & 1323 & 55 & 7.8e-09 & 443 & 935 & 2.5e-02 \\
SBMLR & ADASYN & 346 & 980 & 3.0e-03 & 1081 & 245 & 8.9e-05 & 1095 & 283 & 2.2e-04 & 949 & 377 & 7.3e-03 \\
Ttest & ADASYN & 1107 & 219 & 3.2e-05 & 1222 & 53 & 1.7e-08 & 1231 & 44 & 1.0e-08 & 450 & 876 & 4.6e-02 \\
RS & ADASYN & 399 & 927 & 1.3e-02 & 1071 & 255 & 1.3e-04 & 791 & 535 & 2.3e-01 & 963 & 363 & 4.9e-03 \\
Linear & ADASYN & 419 & 856 & 3.5e-02 & 1067 & 259 & 1.5e-04 & 801 & 525 & 2.0e-01 & 1005 & 321 & 1.3e-03 \\
FWD & ADASYN & 822 & 556 & 2.3e-01 & 1041 & 285 & 4.0e-04 & 1014 & 312 & 1.0e-03 & 559 & 767 & 3.3e-01 \\
\bottomrule
\bottomrule
\end{tabular}
}
\end{table}

\begin{table}
\centering
\caption{Wilcoxon test for the comparison between FS+DS ($R^{+}$) and DS+FS ($R^{-}$), when the MLP classifier is used.}
\label{tab:tabella_risultati_MLP}
\resizebox{1\textwidth}{!}
{
\begin{tabular}{llllllllllllll}
\toprule 
\toprule
\multirow{2}{*}{FS} & \multirow{2}{*}{DS} & \multicolumn{3}{l}{Accuracy (Acc)} & \multicolumn{3}{l}{G-Mean (g)} & \multicolumn{3}{l}{F1} & \multicolumn{3}{l}{IBA} \\ \cline{3-14} 
                    &                     & R+      & R-       & $p$-value     & R+      & R-     & $p$-value   & R+   & R-  & $p$-value & R+    & R-  & $p$-value \\ 
                   \hline

Info & RUS & 856 & 225 & 5.7e-04 & 767 & 361 & 3.2e-02 & 828 & 300 & 5.2e-03 & 469 & 659 & 3.1e-01 \\
ChiS & RUS & 819 & 216 & 6.7e-04 & 786 & 342 & 1.9e-02 & 857 & 271 & 1.9e-03 & 454 & 674 & 2.4e-01 \\
Fish & RUS & 816 & 312 & 7.7e-03 & 936 & 240 & 3.6e-04 & 927 & 249 & 5.1e-04 & 562 & 614 & 7.9e-01 \\
Gini & RUS & 989 & 236 & 1.8e-04 & 994 & 231 & 1.5e-04 & 921 & 255 & 6.4e-04 & 492 & 733 & 2.3e-01 \\
SBMLR & RUS & 1289 & 37 & 4.4e-09 & 1008 & 267 & 3.5e-04 & 1120 & 155 & 3.2e-06 & 494 & 781 & 1.7e-01 \\
Ttest & RUS & 832 & 249 & 1.4e-03 & 1074 & 102 & 6.2e-07 & 1032 & 144 & 5.3e-06 & 662 & 466 & 3.0e-01 \\
RS & RUS & 445 & 830 & 6.3e-02 & 671 & 604 & 7.5e-01 & 551 & 724 & 4.0e-01 & 552 & 723 & 4.1e-01 \\
Linear & RUS & 410 & 766 & 6.8e-02 & 631 & 545 & 6.6e-01 & 497 & 679 & 3.5e-01 & 672 & 504 & 3.9e-01 \\
FWD & RUS & 1213 & 113 & 2.5e-07 & 1215 & 163 & 1.7e-06 & 1274 & 104 & 1.0e-07 & 587 & 791 & 3.5e-01 \\
Info & CNN & 843 & 535 & 1.6e-01 & 1377 & 1 & 3.7e-10 & 1366 & 12 & 7.0e-10 & 932 & 446 & 2.7e-02 \\
ChiS & CNN & 843 & 535 & 1.6e-01 & 1372 & 6 & 5.0e-10 & 1355 & 23 & 1.3e-09 & 895 & 483 & 6.1e-02 \\
Fish & CNN & 793 & 585 & 3.4e-01 & 1348 & 30 & 2.0e-09 & 1302 & 76 & 2.4e-08 & 989 & 389 & 6.3e-03 \\
Gini & CNN & 880 & 498 & 8.2e-02 & 1326 & 0 & 5.1e-10 & 1303 & 23 & 2.0e-09 & 903 & 423 & 2.4e-02 \\
SBMLR & CNN & 842 & 433 & 4.8e-02 & 1064 & 112 & 1.0e-06 & 1042 & 134 & 3.2e-06 & 555 & 621 & 7.4e-01 \\
Ttest & CNN & 945 & 381 & 8.2e-03 & 1215 & 10 & 2.1e-09 & 1178 & 47 & 1.9e-08 & 929 & 296 & 1.6e-03 \\
RS & CNN & 917 & 461 & 3.8e-02 & 1378 & 0 & 3.5e-10 & 1378 & 0 & 3.5e-10 & 784 & 594 & 3.9e-01 \\
Linear & CNN & 892 & 486 & 6.5e-02 & 1378 & 0 & 3.5e-10 & 1378 & 0 & 3.5e-10 & 862 & 516 & 1.2e-01 \\
FWD & CNN & 964 & 414 & 1.2e-02 & 1353 & 25 & 1.5e-09 & 1347 & 31 & 2.1e-09 & 725 & 653 & 7.4e-01 \\
Info & OSS & 167 & 1058 & 9.4e-06 & 985 & 240 & 2.1e-04 & 626 & 599 & 8.9e-01 & 930 & 295 & 1.6e-03 \\
ChiS & OSS & 129.5 & 1095.5 & 1.6e-06 & 994 & 281 & 5.8e-04 & 728 & 547 & 3.8e-01 & 938 & 337 & 3.7e-03 \\
Fish & OSS & 164.5 & 1110.5 & 5.0e-06 & 1106 & 220 & 3.3e-05 & 734 & 592 & 5.1e-01 & 1004 & 322 & 1.4e-03 \\
Gini & OSS & 127 & 1148 & 8.3e-07 & 1038 & 187 & 2.3e-05 & 627 & 598 & 8.9e-01 & 943 & 282 & 1.0e-03 \\
SBMLR & OSS & 341.5 & 786.5 & 1.9e-02 & 749 & 379 & 5.0e-02 & 697 & 431 & 1.6e-01 & 720 & 408 & 9.9e-02 \\
Ttest & OSS & 158 & 970 & 1.7e-05 & 841 & 335 & 9.5e-03 & 399 & 777 & 5.3e-02 & 921 & 255 & 6.4e-04 \\
RS & OSS & 230 & 1045 & 8.4e-05 & 1219 & 107 & 1.9e-07 & 769 & 557 & 3.2e-01 & 1001 & 325 & 1.5e-03 \\
Linear & OSS & 216 & 960 & 1.4e-04 & 1116 & 109 & 5.5e-07 & 722 & 503 & 2.8e-01 & 982 & 243 & 2.4e-04 \\
FWD & OSS & 255 & 1123 & 7.7e-05 & 1012 & 366 & 3.3e-03 & 647 & 731 & 7.0e-01 & 969 & 409 & 1.1e-02 \\
Info & SMOTE & 602 & 574 & 8.9e-01 & 830 & 445 & 6.3e-02 & 680 & 595 & 6.8e-01 & 483 & 742 & 2.0e-01 \\
ChiS & SMOTE & 671 & 604 & 7.5e-01 & 880 & 395 & 1.9e-02 & 702 & 573 & 5.3e-01 & 442 & 833 & 5.9e-02 \\
Fish & SMOTE & 656.5 & 618.5 & 8.5e-01 & 984 & 291 & 8.2e-04 & 745 & 530 & 3.0e-01 & 589 & 686 & 6.4e-01 \\
Gini & SMOTE & 599 & 577 & 9.1e-01 & 952 & 273 & 7.3e-04 & 656 & 569 & 6.7e-01 & 500 & 725 & 2.6e-01 \\
SBMLR & SMOTE & 786 & 390 & 4.2e-02 & 736 & 489 & 2.2e-01 & 686 & 539 & 4.6e-01 & 445 & 780 & 9.6e-02 \\
Ttest & SMOTE & 677 & 548 & 5.2e-01 & 835 & 390 & 2.7e-02 & 649 & 576 & 7.2e-01 & 689 & 536 & 4.5e-01 \\
RS & SMOTE & 313 & 1013 & 1.0e-03 & 617 & 709 & 6.7e-01 & 372 & 954 & 6.4e-03 & 690 & 636 & 8.0e-01 \\
Linear & SMOTE & 326 & 949 & 2.6e-03 & 678 & 597 & 7.0e-01 & 389 & 886 & 1.6e-02 & 746 & 529 & 2.9e-01 \\
FWD & SMOTE & 831.5 & 546.5 & 1.9e-01 & 1034 & 344 & 1.7e-03 & 928 & 450 & 3.0e-02 & 549 & 829 & 2.0e-01 \\
Info & SPIDER & 73 & 1055 & 2.0e-07 & 1085 & 91 & 3.4e-07 & 574 & 602 & 8.9e-01 & 912 & 264 & 8.9e-04 \\
ChiS & SPIDER & 94 & 1131 & 2.5e-07 & 1135 & 90 & 2.0e-07 & 662 & 563 & 6.2e-01 & 887 & 338 & 6.3e-03 \\
Fish & SPIDER & 81 & 1047 & 3.2e-07 & 1088 & 88 & 2.9e-07 & 625 & 551 & 7.0e-01 & 919 & 257 & 6.9e-04 \\
Gini & SPIDER & 59 & 1166 & 3.7e-08 & 1126 & 99 & 3.3e-07 & 568 & 657 & 6.6e-01 & 965 & 260 & 4.5e-04 \\
SBMLR & SPIDER & 142 & 893 & 2.3e-05 & 818 & 263 & 2.4e-03 & 670 & 411 & 1.6e-01 & 674 & 407 & 1.4e-01 \\
Ttest & SPIDER & 68 & 1060 & 1.5e-07 & 998 & 130 & 4.4e-06 & 457 & 671 & 2.6e-01 & 909 & 219 & 2.6e-04 \\
RS & SPIDER & 167 & 1058 & 9.4e-06 & 1192 & 33 & 8.2e-09 & 756 & 469 & 1.5e-01 & 972 & 253 & 3.5e-04 \\
Linear & SPIDER & 155 & 926 & 2.5e-05 & 1103 & 25 & 1.2e-08 & 741 & 387 & 6.1e-02 & 939 & 189 & 7.2e-05 \\
FWD & SPIDER & 77 & 1249 & 4.0e-08 & 1212 & 166 & 1.9e-06 & 687 & 691 & 9.9e-01 & 1047 & 331 & 1.1e-03 \\
Info & ADASYN & 1097 & 281 & 2.0e-04 & 1345 & 33 & 2.3e-09 & 1351 & 27 & 1.7e-09 & 584 & 794 & 3.4e-01 \\
ChiS & ADASYN & 1130 & 248 & 5.9e-05 & 1338 & 40 & 3.4e-09 & 1348 & 30 & 2.0e-09 & 486 & 892 & 6.5e-02 \\
Fish & ADASYN & 1124 & 254 & 7.4e-05 & 1369 & 9 & 5.9e-10 & 1375 & 3 & 4.2e-10 & 532 & 846 & 1.5e-01 \\
Gini & ADASYN & 1145 & 233 & 3.3e-05 & 1365 & 13 & 7.4e-10 & 1370 & 8 & 5.6e-10 & 415 & 963 & 1.3e-02 \\
SBMLR & ADASYN & 1174 & 152 & 1.7e-06 & 942 & 384 & 8.9e-03 & 1044 & 282 & 3.6e-04 & 440 & 886 & 3.7e-02 \\
Ttest & ADASYN & 1227 & 151 & 9.6e-07 & 1321 & 57 & 8.6e-09 & 1348 & 30 & 2.0e-09 & 649 & 729 & 7.2e-01 \\
RS & ADASYN & 837 & 541 & 1.8e-01 & 1324 & 54 & 7.3e-09 & 1311 & 67 & 1.5e-08 & 675 & 703 & 9.0e-01 \\
Linear & ADASYN & 837 & 541 & 1.8e-01 & 1326 & 52 & 6.6e-09 & 1297 & 81 & 3.1e-08 & 749 & 629 & 5.8e-01 \\
FWD & ADASYN & 1031 & 295 & 5.6e-04 & 1309 & 69 & 1.6e-08 & 1309 & 69 & 1.6e-08 & 632 & 746 & 6.0e-01 \\
\bottomrule
\bottomrule
\end{tabular}
}
\end{table}

\begin{table*}
    \centering
		\caption{Results of the Iman-Davenport test and the  Shaffer's post-hoc test described in subsection~\ref{subsec:StatisticalTests} when the SVM classifier is used as the base learner. The tests are applied to classification results computed in terms of IBA. The notations for ``method'',  ``Rank''  and ``$\#p$''  are the same as Table \ref{tab:ShafferAccuracyIR_IBA_c45}. And the explanation and meaning of  the empty columns are the same as Table \ref{tab:ShafferAccuracyIR_IBA_mlp}.}
		\label{tab:ShafferAccuracyIR_IBA_lib} 
\small
     \begin{tabular}{lcllllll}
  \hline 
  \hline
	\multicolumn{2}{l}{\multirow{2}{*}{\diagbox[width=3.1cm,height=0.8cm]{\raisebox{0.2ex}{FS}}{\raisebox{0.2ex}{DS}}}} & \multicolumn{3}{c}{Oversampling} & \multicolumn{3}{c}{Undersampling}\\
	\cline{3-8}
	\multicolumn{2}{l}{}
	                    & Method      & Rank  & $\#p$   & Method      & Rank  & $\#p$ \\
	\hline
        \multirow{12}{*}{\rotatebox[origin=c]{90}{0-33th IR Percentile}}   & \multirow{5}{*}{Filter}   & Gini+ADASYN       & 8.7647  & 2  & Info+RUS        & 4.9118 & 10 \TBstrut\\
                                                   &                       &                   &         &    & ChiS+RUS        & 5.5    & 10 \\
                                                   &                       &                   &         &    & Fish+RUS        & 5.8529 & 10 \\
                                                   &                       &                   &         &    & Gini+RUS        & 6.0588 & 8  \\
                                                   &                       &                   &         &    & Ttest+RUS       & 8.4412 & 3  \\
				\cline{2-8}
										 	  		                       & \multirow{7}{*}{Wrapper}	 &                   &         &    & FWD+RUS         & 3.4118 & 6  \TBstrut\\
													                         &                       &                   &         &    & RUS+Linear       & 3.7941 & 5  \\
													                         &                       &                   &         &    & RUS+RS  & 4.0882 & 4  \\
													                         &                       &                   &         &    & RS+RUS  & 4.2059 & 4  \\
													                         &                       &                   &         &    & Linear+RUS       & 4.9706 & 3  \\
													                         &                       &                   &         &    & FWD+OSS         & 7.6176 & 1  \\
        \midrule
        \multirow{10}{*}{\rotatebox[origin=c]{90}{34th-66th IR Percentile}}     & \multirow{5}{*}{Filter}   & Info+SMOTE        & 8.8824  & 1  & Info+RUS        & 6.7941 & 5 \TBstrut\\
																									 & 										   &                   &         &    & ChiS+RUS        & 7.6471 & 2 \\
																								   &&&&&&&\\
																								   &&&&&&&\\
																								   &&&&&&&\TBstrut\\																									 
				\cline{2-8}
																								   & \multirow{5}{*}{Wrapper}  &                   &         &    & RS+RUS  & 4.9118 & 2 \TBstrut\\
																								   &&&&&&&\\
																								   &&&&&&&\\
																								   &&&&&&&\\
																								   &&&&&&&\TBstrut\\
        \midrule
        \midrule
				\multirow{10}{*}{\rotatebox[origin=c]{90}{67th-100th IR Percentile}}    & \multirow{5}{*}{Filter}   & Fish+SMOTE        & 10.5833 & 1  & Fish+RUS        & 4.4444 & 11 \TBstrut\\
				                                           &                       & Fish+ADASYN       & 10.8056 & 1  & Ttest+RUS       & 4.6111 & 11 \\
				                                           &                       &                   &         &    & Info+RUS        & 5.4167 & 11 \\
				                                           &                       &                   &         &    & Gini+RUS        & 5.5556 & 11 \\
				                                           &                       &                   &         &    & ChiS+RUS        & 6.0833 & 10 \\
			  \cline{2-8}
																								   & \multirow{5}{*}{Wrapper}  &                   &         &    & FWD+RUS         & 3.2778 & 6  \TBstrut\\
																									 &                       &                   &         &    & RS+RUS  & 4.0556 & 6  \\
																									 &                       &                   &         &    & RUS+Linear       & 4.7778 & 4  \\
																									 &                       &                   &         &    & RUS+RS  & 4.8056 & 4  \\
																									 &                       &                   &         &    & Linear+RUS       & 5.1944 & 3  \\
	         												                 &                       &                   &         &    & OSS+FWD         & 8.0294 & 1  \\
 			  \bottomrule
 			  \bottomrule
    \end{tabular}
\end{table*}

\begin{table*}
    \centering
		\caption{Results of the Iman-Davenport test and the  Shaffer's post-hoc test described in subsection~\ref{subsec:StatisticalTests} when the MLP classifier is used as the base learner. The tests are applied to classification results computed in terms of IBA. The notations for ``Method'',  ``Rank''  and ``$\#p$''  are the same as Table \ref{tab:ShafferAccuracyIR_IBA_c45}.  Note that, when the columns in each section of the tables are empty, this means that we do not observe any outperforming method (e.g.,  the wrapper and oversampling combination for datasets having IR percentile between 0 and 33 using the MLP), that is, their statistical difference is insignificant at all. In such cases, our results suggest that researchers to exhaustively compare each possible combination in the two pipelines  to determine which one is the best.(\textit{Continues to Table} \ref{tab:ShafferAccuracyIR_IBA_mlp2})}
\label{tab:ShafferAccuracyIR_IBA_mlp}
\small
     \begin{tabular}{lcllllll}
  \hline 
  \hline
	\multicolumn{2}{l}{\multirow{2}{*}{\diagbox[width=3.1cm,height=0.8cm]{\raisebox{0.2ex}{FS}}{\raisebox{0.2ex}{DS}}}} & \multicolumn{3}{c}{Oversampling} & \multicolumn{3}{c}{Undersampling}\\
	\cline{3-8}
	\multicolumn{2}{l}{}
	                    & Method      & Rank  & $\#p$   & Method      & Rank  & $\#p$ \\
	\hline
  \multirow{28}{*}{\rotatebox[origin=c]{90}{0-33th IR Percentile}}     & \multirow{17}{*}{Filter}   & Fish+SMOTE        & 9.1176  & 5  & Fish+RUS         & 8.1765  & 5 \TBstrut\\
																						       &                        & Fish+SPIDER       & 9.1176  & 5  & Fish+CNN         & 8.1765  & 5 \\
																						       &                        & Fish+ADASYN       & 9.1176  & 5  & Fish+OSS         & 8.1765  & 5 \\
																						       &                        & ChiS+SMOTE        & 10.2647 & 4  & ChiS+RUS         & 9.7941  & 4 \\
																						       &                        & ChiS+SPIDER       & 10.2647 & 4  & ChiS+CNN         & 9.7941  & 4 \\
																						       &                        & ChiS+ADASYN       & 10.2647 & 4  & ChiS+OSS         & 9.7941  & 4 \\
																						       &                        & Gini+SMOTE        & 10.7647 & 3  & Gini+RUS         & 10.0588 & 4 \\
																						       &                        & Gini+SPIDER       & 10.7647 & 3  & Gini+CNN         & 10.0588 & 4 \\
																						       &                        & Gini+ADASYN       & 10.7647 & 3  & Gini+OSS         & 10.0588 & 4 \\
																						       &                        & Info+SMOTE        & 11.1765 & 3  & Info+RUS         & 10.4412 & 4 \\
																						       &                        & Info+SPIDER       & 11.1765 & 3  & Info+CNN         & 10.4412 & 4 \\
																						       &                        & Info+ADASYN       & 11.1765 & 3  & Info+OSS         & 10.4412 & 4 \\
																						       &                        & SMOTE+ChiS        & 11.6471 & 3  & Ttest+RUS        & 11.4412 & 2 \\
																						       &                        & Ttest+SMOTE       & 11.8235 & 1  & Ttest+CNN        & 11.4412 & 2 \\
																						       &                        & Ttest+SPIDER      & 11.8235 & 1  & Ttest+OSS        & 11.4412 & 2 \\
																						       &                        & Ttest+ADASYN      & 11.8235 & 1  &                  &         &   \\
																						       &                        & SMOTE+Info        & 12.1765 & 1  &                  &         &   \\
				\cline{2-8}
													                         & \multirow{11}{*}{Wrapper}	& FWD+SMOTE         & 5.5294  & 2  & FWD+RUS          & 5.0588  & 4 \TBstrut\\
													                         &                        & FWD+SPIDER        & 5.5294  & 2  & FWD+CNN          & 5.0588  & 4 \\
													                         &                        & FWD+ADASYN        & 5.5294  & 2  & FWD+OSS          & 5.0588  & 4 \\
													                         &                        &                   &         &    & Linear+RUS        & 5.9118  & 3 \\
													                         &                        &                   &         &    & Linear+CNN        & 5.9118  & 3 \\
													                         &                        &                   &         &    & Linear+OSS        & 5.9118  & 3 \\
													                         &                        &                   &         &    & RUS+RS   & 6.6765  & 3 \\
													                         &                        &                   &         &    & RUS+Linear        & 7.0294  & 3 \\
													                         &                        &                   &         &    & RS+RUS   & 8.2059  & 1 \\
													                         &                        &                   &         &    & RS+CNN   & 8.2059  & 1 \\
													                         &                        &                   &         &    & RS+OSS   & 8.2059  & 1 \\	                         
			  \bottomrule
			  \bottomrule	  
    \end{tabular}
\end{table*}

\begin{table*}
    \centering
		\caption{Results of the Iman-Davenport test and the  Shaffer's post-hoc test described in subsection~\ref{subsec:StatisticalTests} when the MLP classifier is used as the base learner. (\textit{Continuation of Table \ref{tab:ShafferAccuracyIR_IBA_mlp}, showing  the 34-66th and 67th-100th IR Percentiles)}}
\label{tab:ShafferAccuracyIR_IBA_mlp2}
\small
     \begin{tabular}{lcllllll}
  \hline 
  \hline
	\multicolumn{2}{l}{\multirow{2}{*}{\diagbox[width=3.1cm,height=0.8cm]{\raisebox{0.2ex}{FS}}{\raisebox{0.2ex}{DS}}}} & \multicolumn{3}{c}{Oversampling} & \multicolumn{3}{c}{Undersampling}\\
	\cline{3-8}
	\multicolumn{2}{l}{}
	                    & Method      & Rank  & $\#p$   & Method      & Rank  & $\#p$ \\
	\hline
	        \multirow{22}{*}{\rotatebox[origin=c]{90}{34th-66th IR Percentile}}     & \multirow{11}{*}{Filter}   & ChiS+SMOTE        & 10.4706 & 4  & Ttest+RUS        & 10.6471 & 3 \TBstrut\\
	                                                   &                        & ChiS+SPIDER       & 10.4706 & 4  & Ttest+CNN        & 10.6471 & 3 \\
	                                                   &                        & ChiS+ADASYN       & 10.4706 & 4  & Ttest+OSS        & 10.6471 & 3 \\
	                                                   &                        & SMOTE+Info        & 10.6471 & 4  & ChiS+RUS         & 10.9412 & 2 \\
	                                                   &                        & Ttest+SMOTE       & 10.6471 & 4  & ChiS+CNN         & 10.9412 & 2 \\
	                                                   &                        & Ttest+SPIDER      & 10.6471 & 4  & ChiS+OSS         & 10.9412 & 2 \\
	                                                   &                        & Ttest+ADASYN      & 10.6471 & 4  & Gini+RUS         & 11.5588 & 2 \\
	                                                   &                        &                   &         &    & Gini+CNN         & 11.5588 & 2 \\
	                                                   &                        &                   &         &    & Gini+OSS         & 11.5588 & 2 \\
	                                                   &                        &                   &         &    & RUS+ChiS         & 11.9706 & 2 \\
	                                                   &                        &                   &         &    & RUS+Info         & 12.2059 & 1 \\
					\cline{2-8}
																										 & \multirow{11}{*}{Wrapper}  &                   &         &    & RUS+RS   & 5.2059  & 4 \TBstrut\\
																										 &                        &                   &         &    & RUS+Linear        & 5.2059  & 4 \\
																										 &                        &                   &         &    & RS+RUS   & 5.9412  & 3 \\
																										 &                        &                   &         &    & RS+CNN   & 5.9412  & 3 \\
																										 &                        &                   &         &    & RS+OSS   & 5.9412  & 3 \\
																										 &                        &                   &         &    & Linear+RUS        & 7.6765  & 2 \\
																										 &                        &                   &         &    & Linear+CNN        & 7.6765  & 2 \\
																										 &                        &                   &         &    & Linear+OSS        & 7.6765  & 2 \\
																										 &                        &                   &         &    & FWD+RUS          & 7.8235  & 2 \\
																										 &                        &                   &         &    & FWD+CNN          & 7.8235  & 2 \\
																										 &                        &                   &         &    & FWD+OSS          & 7.8235  & 2 \\
				\midrule
				\midrule
				\multirow{21}{*}{\rotatebox[origin=c]{90}{67th-100th IR Percentile}}    & \multirow{12}{*}{Filter}   & Fish+SMOTE        & 8.75    & 4  & Fish+RUS         & 8.8611  & 5 \TBstrut\\
				                                           &                        & Fish+SPIDER       & 8.75    & 4  & Fish+CNN         & 8.8611  & 5 \\
				                                           &                        & Fish+ADASYN       & 8.75    & 4  & Fish+OSS         & 8.8611  & 5 \\
				                                           &                        & ChiS+SMOTE        & 9.5278  & 3  & ChiS+RUS         & 9.1389  & 5 \\
				                                           &                        & ChiS+SPIDER       & 9.5278  & 3  & ChiS+CNN         & 9.1389  & 5 \\
				                                           &                        & ChiS+ADASYN       & 9.5278  & 3  & ChiS+OSS         & 9.1389  & 5 \\
				                                           &                        & Gini+SMOTE        & 10.3611 & 1  & Gini+RUS         & 9.6667  & 5 \\
				                                           &                        & Gini+SPIDER       & 10.3611 & 1  & Gini+CNN         & 9.6667  & 5 \\
				                                           &                        & Gini+ADASYN       & 10.3611 & 1  & Gini+OSS         & 9.6667  & 5 \\
				                                           &                        &                   &         &    & Info+RUS         & 9.9444  & 5 \\
				                                           &                        &                   &         &    & Info+CNN         & 9.9444  & 5 \\
				                                           &                        &                   &         &    & Info+OSS         & 9.9444  & 5 \\
			  \cline{2-8}
																									 & \multirow{9}{*}{Wrapper}   & SMOTE+RS  & 5.8611  & 3  & Linear+RUS        & 5.75    & 3 \TBstrut\\
																									 &                        & Linear+SMOTE       & 6.3889  & 3  & Linear+CNN        & 5.75    & 3 \\
																									 &                        & Linear+SPIDER      & 6.3889  & 3  & Linear+OSS        & 5.75    & 3 \\
																									 &                        & Linear+ADASYN      & 6.3889  & 3  & FWD+RUS          & 6       & 3 \\
																									 &                        & SMOTE+Linear       & 6.4167  & 3  & FWD+CNN          & 6       & 3 \\
																									 &                        & FWD+SMOTE         & 6.4444  & 3  & FWD+OSS          & 6       & 3 \\
																									 &                        & FWD+SPIDER        & 6.4444  & 3  & RS+RUS   & 7.3056  & 2 \\
																									 &                        & FWD+ADASYN        & 6.4444  & 3  & RS+CNN   & 7.3056  & 2 \\
																									 &                        &                   &         &    & RS+OSS   & 7.3056  & 2 \\	
			  \bottomrule
			  \bottomrule	  
    \end{tabular}
\end{table*}

\twocolumn

\end{document}